\documentclass{article}
\usepackage{iclr2027_conference,times}

\usepackage{amsmath,amsfonts,bm}

\def\eqref#1{equation~\ref{#1}}

\def\1{\bm{1}}

\DeclareMathAlphabet{\mathsfit}{\encodingdefault}{\sfdefault}{m}{sl}
\SetMathAlphabet{\mathsfit}{bold}{\encodingdefault}{\sfdefault}{bx}{n}

\usepackage{graphicx}
\usepackage[table]{xcolor}
\usepackage{multirow}
\usepackage{booktabs}
\usepackage{subcaption}
\usepackage{bbm}
\usepackage{booktabs}       
\usepackage{amsfonts}       
\usepackage{nicefrac}       
\usepackage{microtype}      
\usepackage{xcolor}         
\usepackage{amssymb}
\usepackage{multirow}
\usepackage{algorithm}
\usepackage{algpseudocode}
\usepackage{tabularray}
\usepackage{tabularx}
\newcolumntype{C}{>{\centering\arraybackslash}X}
\usepackage{amsmath}
\usepackage{amsthm}
\usepackage{wrapfig}

\usepackage[utf8]{inputenc} 
\usepackage[T1]{fontenc}    
\usepackage{hyperref}       
\usepackage{url}            
\usepackage{booktabs}       
\usepackage{amsfonts}       
\usepackage{nicefrac}       
\usepackage{microtype}      
\usepackage{xcolor}         
\usepackage{amssymb}
\usepackage{multirow}
\usepackage{algorithm}
\usepackage{algpseudocode}
\usepackage{tabularray}
\usepackage{tabularx}
\newcolumntype{C}{>{\centering\arraybackslash}X}

\usepackage{amsmath}
\usepackage{amsthm}
\usepackage{wrapfig}
\title{CHAIN: Calibrated LLM Forecasting via Causal-Temporal Hypergraph Inference}

\author{%
  {\bfseries Wenjin Liu\textsuperscript{\normalfont 1,2}, Chenxi Wang\textsuperscript{\normalfont 2}, Yue Lu\textsuperscript{\normalfont 2}, Zhe Cui\textsuperscript{\normalfont 1}\thanks{Corresponding authors.\quad \textsuperscript{2}\url{https://github.com/QwenQKing/Chain}}\hspace{0.5em}, Haoran Luo\textsuperscript{\normalfont 1}\footnotemark[1]}\\
  \textsuperscript{1}Nanyang Technological University\qquad \textsuperscript{2}Hithink research
}

\iclrfinalcopy

\begin{document}

\maketitle
\lhead{Preprint}

\begin{abstract}
Large language models have achieved significant progress in event forecasting, yet their probability outputs exhibit systematic calibration bias that varies heterogeneously across different domains and question types, undermining the trustworthiness of probabilistic outputs for decision-making under uncertainty. However, existing calibration methods typically correct probability outputs after prediction is complete, without modeling the structural sources of bias within the prediction process itself. To address this challenge, we decompose probabilistic prediction over causal-temporal hypergraphs into three stages, evidence weighting, evidence aggregation, and source fusion, and propose \textbf{\textsc{Chain}}, which designs stage-specific mechanisms to mitigate bias at each stage: \textit{(i)} modulating the temporal decay function by causal topological distance, \textit{(ii)} aggregating approximately independent causal chains via Noisy-OR after direction-aware deduplication, and \textit{(iii)} driving adaptive fusion by causal coverage and directional balance. Experimental results on cross-domain forecasting benchmarks show \textbf{\textsc{Chain}} outperforms existing methods in expected calibration error, Brier score, and accuracy.
Our project is available\href{https://github.com/QwenQKing/Chain}{\textsuperscript{1}}.
\end{abstract}

\section{Introduction}

\begin{wrapfigure}{r}{0.5\textwidth} 
  \vspace{-9pt}                       
  \centering
  \includegraphics[width=0.98\linewidth]{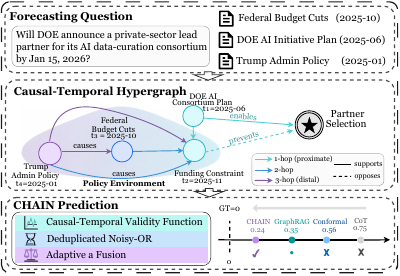}
  \caption{An illustration of the proposed \textbf{\textsc{Chain}}.}
  \label{chain_case}
  \vspace{-3pt}                       
\end{wrapfigure}

Large language models (LLMs) have achieved significant progress in event forecasting~\citep{zhang2026following,paleka2025pitfalls}, advancing from direct prompting~\citep{halawi2024approaching,yang2026okg,wildman2025bench} and chain-of-thought (CoT)~\citep{wei2022chain} reasoning to retrieval-augmented prediction~\citep{wang2024news,guan2024openep} and consistency-aware evaluation~\citep{farquhar2024detecting,nakkiran2025trained}, and further to structured reasoning over temporal knowledge graphs~\citep{chen2024temporal}, with prediction accuracy approaching human crowd levels~\citep{baan2022stop,peng2025graph}. However, probability calibration remains a critical bottleneck for reliability, as model probability outputs exhibit persistent and widespread overconfidence~\citep{xiao2025graphrag}, and calibration bias is heterogeneous across domains and question types~\citep{niu2025event,luo2024open}, undermining the trustworthiness of probabilistic outputs for real-world decision-making under uncertainty~\citep{zeng2025futurex,tao2025prophet}.

\begin{figure}[t]
\centering
\includegraphics[width=1\linewidth]{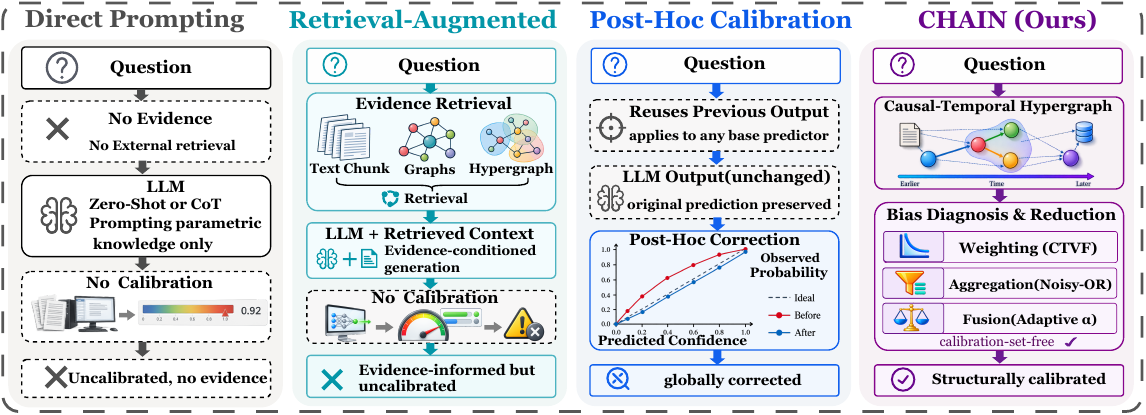}
\vspace{-3.6mm}
\caption{Inference-time approaches to calibrated event forecasting: existing methods and \textbf{\textsc{Chain}}.}
\label{baselines}
\end{figure}

To calibrate predictions, post-hoc methods correct probability outputs after prediction~\citep{geng2024survey}, progressing from global parameter adjustment~\citep{guo2017calibration,platt1999probabilistic} to sample-adaptive~\citep{xie2024calibrating}, semantic-level~\citep{lamb2025semantic}, and alignment-aware~\citep{xiao2025restoring} calibration. Non-parametric methods~\citep{zadrozny2002transforming,zadrozny2001obtaining} and conformal prediction~\citep{cherian2024large,manggala2024qa} provide distribution-free alternatives; prompt-based methods~\citep{zhang2024calibrating} elicit verbalized confidence. These methods correct outputs post hoc~\citep{geng2024survey} without modeling the structural sources of bias in the prediction.

However, these methods still face three key challenges in calibration: \textit{(i)} Evidence at different causal distances is treated equally, with no distinction between \textbf{proximate causes} and \textbf{indirect background}, causing distal evidence to influence estimates as much as proximal ones. \textit{(ii)} Multiple retrieval paths referencing the same fact participate in estimation without deduplication, causing the model to treat \textbf{redundant paths} as \textbf{independent evidence}, leading to systematically inflated confidence. \textit{(iii)} The reliability of external evidence and internal estimates varies across questions, yet existing methods fuse them in a \textbf{fixed manner}, \textbf{lacking adaptive adjustment} according to evidence quality.

To address these challenges, we decompose probabilistic prediction over causal-temporal hypergraphs into three stages of evidence weighting, aggregation, and source fusion, and propose \textbf{\textsc{Chain}} (see Figure \ref{chain_case}), which employs stage-specific mechanisms to mitigate bias at each stage. First, we propose the Causal-Temporal Validity Function (CTVF), modulating temporal decay by causal topological distance, capturing recency and causal relevance across retrieved evidence. Furthermore, we introduce direction-aware deduplication with Noisy-OR aggregation over approximately independent causal paths, reducing redundant confidence accumulation. In addition, we design adaptive fusion driven by causal coverage and directional balance, weighted by evidence quality to adjust source contributions.

We conduct experiments on 8 cross-domain forecasting benchmarks with multiple LLMs. Experimental results show \textbf{\textsc{Chain}} outperforms the baselines (see Figure \ref{baselines}) in both calibration and accuracy, with ablations showing that each removal degrades final calibration or prediction performance, validating the effectiveness of structural diagnosis over causal-temporal hypergraphs for probability calibration.

\section{Related Work}

\textbf{Event Forecasting.} Event forecasting assigns probabilities to events~\citep{zou2022forecasting,yuan2025forecast}. Direct prompting~\citep{yuan2025forecast,lee2023temporal}, CoT~\citep{wei2022chain}, and retrieval methods (AutoCast++~\citep{yan2023autocast}; The Power of Simplicity~\citep{zhang2025power}) cover evidence-driven forecasting. Large-scale training~\citep{lee2025advancing}, consistency checks~\citep{paleka2024consistency}, and ensembles~\citep{schoenegger2024wisdom} improve forecasts, while calibration varies across domains~\citep{karkar2025future,lu2025evaluating}. GenTKG~\citep{liao2024gentkg}, INFER~\citep{li2025infer}, TPAR~\citep{chen2024unified}, and zrLLM~\citep{ding2024zrllm} perform graph extrapolation; context-aware modeling~\citep{manggala2024qa} uses history. RAG~\citep{gutierrez2024hipporag,luo2025hypergraphrag} and multi-cognition aggregation~\citep{wang2025beyond} provide knowledge and multi-perspective fusion.

\textbf{Probability Calibration.} Calibration aligns predictions with observed frequencies~\citep{geng2024survey}. Post-hoc methods include temperature scaling~\citep{guo2017calibration}, Platt scaling~\citep{platt1999probabilistic}, ATS~\citep{xie2024calibrating}, and internal consistency calibration~\citep{xie2024internal}. Isotonic regression~\citep{zadrozny2002transforming}, histogram binning~\citep{zadrozny2001obtaining}, and QA-calibration~\citep{manggala2024qa} are non-parametric; ConU~\citep{wang2024conu} and enhanced conformal prediction~\citep{cherian2024large} provide distribution-free coverage guarantees. Just Ask~\citep{tian2023just} and Fidelity~\citep{zhang2024calibrating} elicit confidence, while CFT~\citep{xiao2025restoring} repairs alignment-induced degradation. Ca2KG~\citep{ren2026trust} measures evidence-reliability variation, and Double-Calibration~\citep{lu2026double} calibrates evidence and reasoning confidence.

\begin{figure*}[t]
\centering
\includegraphics[width=0.98\linewidth]{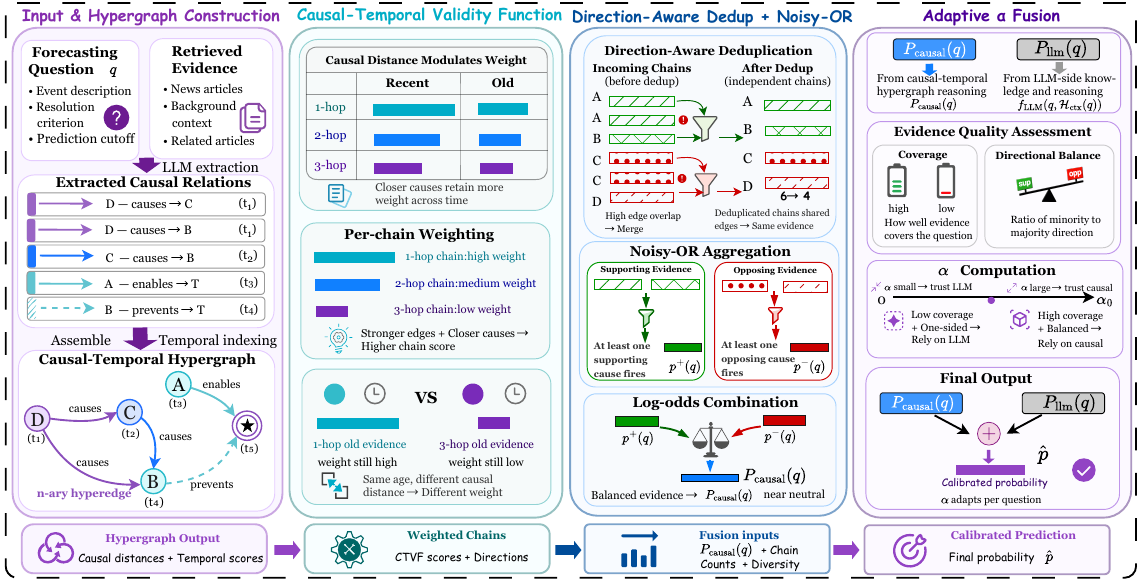}
\caption{The overall framework of \textbf{\textsc{Chain}}, with
CTVF, Noisy-OR aggregation, and adaptive fusion.}
\label{fig:framework}
\end{figure*}

\section{Preliminaries}

\noindent\textbf{Event Forecasting and Calibration.} At prediction cutoff $t_c$, 
the forecaster $f$ outputs for question $q$:
\begin{equation}
p = f(q, t_c) \in [0, 1], \quad y \in \{0, 1\}.
\end{equation}
Calibration quality is measured by the Expected Calibration Error (ECE), 
which is formally defined as 
$\mathrm{ECE} = \sum_{b=1}^{B} (n_b / N) \lvert \bar{p}_b - \bar{y}_b \rvert$, 
where $B$ is the number of equal-width bins, $n_b$, $\bar{p}_b$, $\bar{y}_b$ 
are the size, mean prediction, and empirical frequency of bin $b$ 
respectively, and $N$ is the total sample count.

\noindent\textbf{Causal-Temporal Hypergraph.} Evidence is organized as a 
causal-temporal hypergraph:
\begin{equation}
\mathcal{G} = (V, E_H, \mathcal{C}),
\end{equation}
where $V$ is the entity set; $(V_e,t,s)\in E_H$ stores $V_e\subseteq V$, the
source-chunk timestamp $t$, and optional stored similarity $s\in[0,1]$ when available;
and $(u,v,\mathrm{type},w)\in\mathcal C$ is a directed edge, with
$\mathrm{type}\in\{\mathrm{causes},\mathrm{enables},\mathrm{prevents}\}$ and
$w\in[0,1]$. For each chunk $x$, $\phi_{\mathrm{LLM}}$ returns propositions and
typed edges $\mathcal{C}_x$; the builder attaches $t,s$ to form $\mathcal{P}_x$ before graph assembly:
\begin{equation}
E_H = \bigcup_{x} \mathcal{P}_x, \quad 
\mathcal{C} = \bigcup_{x} \mathcal{C}_x, \quad 
V = \bigcup_{(V_e, t, s) \in E_H} V_e.
\end{equation}
At inference time, a cutoff-conditioned constructor derives an active graph
$\mathcal G_q$ from the complete hypergraph, and both estimators use only this view:
\begin{equation}
\mathcal{H}_{\mathrm{ctx}}(q)=\mathrm{PromptContext}(\mathcal{G}_q,q,t_c),\qquad
P_{\mathrm{llm}}=f_{\mathrm{LLM}}(q,\mathcal{H}_{\mathrm{ctx}}(q))\in[0,1].
\end{equation}
Here $\mathcal H_{\mathrm{ctx}}(q)$ contains narratives, trajectories, contradictions,
causal-context chains, source events, and prior outcomes; outcome mapping projects the estimate onto the binary axis.

\noindent\textbf{Causal Topological Distance.}
Using $\mathrm{Enc}(\cdot)$, keyword-filtered dense reranking maps signed target
descriptions from the reasoning-direction call to validated targets
$\mathcal{T}(q)\subseteq V_q$, with $a_q(v)\in\{-1,+1\}$ marking the negative/positive axis side.
For any node $n$ in $\mathcal{G}_q$, define its shortest distance as:
\begin{equation}
d(n, q) = \min_{v \in \mathcal{T}(q)} d_{\mathcal{C}_q}(v, n),
\end{equation}
where $d_{\mathcal{C}_q}$ is the unweighted Breadth-First Search (BFS) distance
on the reverse of $\mathcal{C}_q$ (all types treated uniformly); a hyperedge
$e$ inherits $d(e, q) = \min_{v \in V_e} d(v, q)$; unreachable nodes have 
$d = \infty$.

\section{Method: \textbf{\textsc{Chain}}}

In this section, we introduce \textbf{\textsc{Chain}} (see 
Figure~\ref{fig:framework}), which addresses calibration bias at three 
pipeline stages through the Causal-Temporal Validity Function for evidence 
weighting, direction-aware Noisy-OR for chain aggregation, and 
causal-coverage-balanced adaptive fusion for source combination.

\subsection{Causal-Temporal Validity Function}

The CTVF assigns each piece of evidence in $\mathcal{G}_q$ a validity
$V_{\mathrm{CTVF}} \in (0, 1]$ reflecting temporal recency and causal 
proximity to $\mathcal{T}(q)$, propagated along causal paths into 
chain-level confidences $\Phi(\pi, q)$.

\noindent\textbf{Causal-Proximity Exponent.}
Using the causal topological distance $d(n, q)$ defined in the 
preliminaries, a proximity exponent $\rho(n, q) \in [0, 1]$ is obtained 
for every node $n$ in the active graph with $n \in V_q \cup E_{H,q}$ by normalizing
$d(n, q)$ and clipping it against the reachability horizon 
$d_{\max} \in \mathbb{Z}^+$:
\begin{equation}
\rho(n, q) = \min\!\left(\frac{d(n, q)}{d_{\max}},\; 1\right).
\end{equation}
Thus $\rho=0$ at the target, $\rho\in(0,1)$ within the horizon, and $\rho=1$
at or beyond it or for unreachable nodes; larger $\rho$ induces faster temporal decay for distant evidence.


\noindent\textbf{Temporal Recency Component.}
For a hyperedge $e = (V_e, t, s) \in E_{H,q}$ with a parseable timestamp $t < t_c$,
let $\Delta t(e)=\max\{1,(t_c-t)_{\mathrm{days}}\}$. The temporal recency component 
$V_{\mathrm{rec}}(e, q) \in (0, 1]$ is modeled as an exponential function of 
the logarithmic span, whose decay rate is scaled by $\rho(e, q)$:
\begin{equation}
V_{\mathrm{rec}}(e, q) = 
\exp\!\Bigl(-\tfrac{1}{2}\,\rho(e, q) \cdot \ln \Delta t(e)\Bigr).
\end{equation}
At $\rho(e, q) = 0$ the exponent vanishes and $V_{\mathrm{rec}} = 1$, so 
causally direct evidence retains full recency; at $\rho(e, q) = 1$, 
$V_{\mathrm{rec}} = 1 / \sqrt{\Delta t(e)}$, yielding maximal temporal 
decay for distant evidence.

\noindent\textbf{Recurrence-Salience Component.}
Let $N(e)$ count distinct admitted record IDs for $e$'s canonicalized proposition key, and let
$M_r(e)=s\in[0,1]$ be an optional stored similarity. The centered logistic coefficient
$\mu(N)$ amplifies co-supported evidence and suppresses low-signal singletons:
\begin{equation}
\mu(N) = \frac{1}{1 + \exp(-\sqrt{N} / \eta)} - \tfrac{1}{2}, 
\qquad 
V_{\mathrm{sal}}(e) = \mu(N(e)) \cdot \sqrt{N(e)} \cdot M_r(e),
\end{equation}
where $\eta>0$ controls saturation; $\sqrt{N(e)}$ yields sub-linear recurrence
amplification, and $M_r(e)$ optionally down-weights weak semantic matches under weak similarity.

\noindent\textbf{CTVF Validity.}
For valid pre-cutoff timestamps, the two components are summed and capped at $1$:
\begin{equation}
V_{\mathrm{CTVF}}(e, q) = 
\min\!\bigl(V_{\mathrm{rec}}(e, q) + V_{\mathrm{sal}}(e),\; 1\bigr).
\end{equation}
Known $t\geq t_c$ is excluded; missing or unparseable $t$ uses $v_{\mathrm{fb}}$ only with a verified pre-cutoff availability bound.
Entity scores average admitted incident-hyperedge values, with fallback $0.5$ for aggregation.

\noindent\textbf{Edge-Level Weight.}
For each directed causal edge $c = (u, v, \mathrm{type}, w) \in \mathcal{C}_q$,
the CTVF-weighted edge scalar $\omega(c) \in [0, 1]$ is the mean of the 
endpoint validities raised to an edge-level exponent $\rho_c$:
\begin{equation}
\omega(c) = 
\left(\tfrac{1}{2}\bigl(V_{\mathrm{CTVF}}(u, q) + V_{\mathrm{CTVF}}(v, q)\bigr)\right)^{\!\rho_c}, 
\qquad 
\rho_c = \min\!\bigl(\tfrac{d(u, q) + d(v, q)}{2\, d_{\max}},\;1\bigr) \in [0, 1].
\end{equation}
Edges located near the target receive a small $\rho_c$ and keep their 
endpoint mean close to $1$; meanwhile, distal edges have $\rho_c \to 1$ and inherit 
the uncorrected raw mean validity $\omega(c)$ of their endpoints.

\noindent\textbf{Target-Termination Gate.}
Writing $c_\ell=(v_{\ell-1},v_\ell,\mathrm{type}_\ell,w_\ell)$, a node-simple
directed chain $\pi=(c_1,\ldots,c_L)$, $1\leq L\leq d_{\max}$, is retained
only when it terminates at an oriented target:
\begin{equation}
\iota_q(\pi)=\mathbbm{1}\!\left[v_L\in\mathcal T(q)\right].
\end{equation}

\noindent\textbf{Chain-Level Confidence.}
For every score-qualified candidate chain, the confidence
$\Phi(\pi, q) \in [0, 1]$ is the target-gated form of the ungated
causal-chain score $\widetilde{\Phi}(\pi,q)$:
\begin{equation}
\widetilde{\Phi}(\pi,q)=\prod_{\ell=1}^{L}w_\ell\omega(c_\ell),
\qquad \Phi(\pi,q)=\iota_q(\pi)\widetilde{\Phi}(\pi,q).
\end{equation}
Forward breadth-first search expands at most $F_{\max}$ outgoing edges per node and
prunes prefixes whose ungated score is below $\tau_\Phi$. Only score-qualified chains
terminating at oriented targets enter the top-$B_\pi$ pool $\Pi_q$; otherwise the branch
abstains from contributing evidence to the final forecast.

\par\noindent\textbf{Proposition 1.} \textit{Under normalized weighting, CTVF lowers
weighting error versus uniform weighting by favoring lower-error proximal evidence
without increasing the weighted error within either subset.}\vspace{-3mm}
\begin{proof}
We provide experimental results in Section~\ref{sec:exp_ablation} and 
theoretical proofs in Appendix~\ref{app:proof1}.
\end{proof}

\subsection{Evidence Aggregation via Direction-aware Noisy-OR}

Direction-Aware Noisy-OR reduces redundant paths from acting as independent evidence through polarity-wise deduplication and neutral-anchored log-odds aggregation of directional evidence.

\noindent\textbf{Causal Polarity Encoding.}
Each retained chain $\pi=(c_1,\ldots,c_L)$ receives a binary question-conditioned
polarity $\delta_q(\pi)\in\{+1,-1\}$. Let $P(\pi)$ count its prevents edges;
target orientation and relation parity determine the polarity partitions:
\begin{equation}
\delta_q(\pi) = a_q(v_L)(-1)^{P(\pi)}, 
\qquad 
\Pi_q^{+} = \{\pi \in \Pi_q : \delta_q(\pi) = +1\}, 
\quad 
\Pi_q^{-} = \{\pi \in \Pi_q : \delta_q(\pi) = -1\}.
\end{equation}
The endpoint orientation fixes the target's positive/negative axis side and final polarity, while each
$\mathrm{prevents}$ edge reverses the direction propagated along the chain during aggregation and fusion.

\noindent\textbf{Same-Polarity Deduplication.}
Two same-polarity chains $\pi_i, \pi_j$ are treated as non-independent 
when their edge sets overlap excessively. Let $E(\pi)$ denote the edge 
set of chain $\pi$; the Jaccard overlap is:
\begin{equation}
J(\pi_i, \pi_j) = \frac{\lvert E(\pi_i) \cap E(\pi_j) \rvert}
{\lvert E(\pi_i) \cup E(\pi_j) \rvert}.
\end{equation}
Within each polarity set $\Pi_q^{\pm}$, the chains are sorted in descending 
order of $\Phi(\pi, q)$ and greedily added to $\widehat{\Pi}_q^{\pm}$ for subsequent greedy
selection: a chain $\pi$ is admitted iff
\begin{equation}
\max_{\pi' \in \widehat{\Pi}_q^{\pm}} J(\pi, \pi') \leq \theta,
\end{equation}
Under the convention $\max\varnothing := 0$ and overlap threshold $\theta \in (0, 1)$,
the procedure greedily retains each chain whose overlap with all previously retained
chains is at most $\theta$, and discards the rest.

\noindent\textbf{Polarity-Wise Noisy-OR.}
The surviving chains in each channel are combined under a within-polarity approximate-independence assumption. The resulting channel scores are:
\begin{equation}
p^{+}(q) = 1 - \prod_{\pi \in \widehat{\Pi}_q^{+}}\!\bigl(1 - \Phi(\pi, q)\bigr), 
\qquad 
p^{-}(q) = 1 - \prod_{\pi \in \widehat{\Pi}_q^{-}}\!\bigl(1 - \Phi(\pi, q)\bigr),
\end{equation}
Where $p^{+}(q),p^{-}(q)\in[0,1]$ are the positive- and negative-channel aggregates,
they summarize the surviving evidence in the two directional channels for downstream fusion.

\noindent\textbf{Neutral-Anchored Log-Odds.}
For fixed $\varepsilon=10^{-6}$, define $g_\varepsilon(p)=
\mathrm{logit}(\min(1-\varepsilon,\max(\varepsilon,(1+p)/2)))$ to map zero directional evidence
to neutral log-odds and obtain the causal probability:
\begin{equation}
P_{\mathrm{causal}}(q) =
\sigma\!\Bigl(g_\varepsilon(p^{+}(q))-g_\varepsilon(p^{-}(q))\Bigr),
\end{equation}
where $\mathrm{logit}(p)=\ln(p/(1-p))$ and $\sigma(z)=1/(1+e^{-z})$.
The rule returns $1/2$ for empty or equally strong channels, increases
continuously with $p^{+}$, and decreases continuously with $p^{-}$, and
remains neutral when directional evidence is absent or balanced across both channels.

\par\noindent\textbf{Proposition 2.} \textit{Direction-aware deduplication can eliminate the same-polarity Noisy-OR probability inflation induced by overlap-identified redundant causal chains.}
\vspace{-3mm}
\begin{proof}
We provide experimental results in Section~\ref{sec:exp_ablation} and 
theoretical proofs in Appendix~\ref{app:proof2}.
\end{proof}

\subsection{Source Fusion via Causal-Coverage-Balanced Adaptive Fusion}

Because evidence reliability varies, fixed weighting of $P_{\mathrm{llm}}$ and $P_{\mathrm{causal}}$ may be suboptimal; adaptive fusion derives bounded $\alpha(q)$ from retained evidence and blends them into $\hat{p}$.


\noindent\textbf{Evidence Reliability.}
Let $k = |\widehat{\Pi}_q^{+}| + |\widehat{\Pi}_q^{-}|$ denote the number of approximately independent chains after deduplication. A logarithmic reliability 
score $\kappa(k) \in [0, 1]$ saturates as more such chains accumulate:
\begin{equation}
\kappa(k) = \min\!\left(\frac{\ln(1 + k)}{\ln(1 + k_{\mathrm{sat}})},\; 1\right),
\end{equation}
where $k_{\mathrm{sat}} \in \mathbb{Z}^+$ is a saturation 
anchor controlling the logarithmic curve. Hence $\kappa(k)$ grows quickly 
with the first few independent chains and gradually saturates once their 
count $k$ exceeds $k_{\mathrm{sat}}$.

\noindent\textbf{Directional Balance.}
A polarity-imbalance factor $\beta(q) \in [0, 1]$ explicitly quantifies
the relative evenness of the positive- and negative-polarity chain sets
$\widehat{\Pi}_q^{+}$ and $\widehat{\Pi}_q^{-}$ via:
\begin{equation}
\beta(q) = \frac{\min\!\bigl(|\widehat{\Pi}_q^{+}|,\, |\widehat{\Pi}_q^{-}|\bigr)}
{\max\!\bigl(|\widehat{\Pi}_q^{+}|,\, |\widehat{\Pi}_q^{-}|,\, 1\bigr)}.
\end{equation}
A value of $\beta(q) = 0$ indicates fully one-sided evidence, whereas 
$\beta(q) = 1$ indicates a perfectly balanced split between sides. To 
prevent one-sided evidence from being entirely discarded, $\beta(q)$ is 
mapped through an affine floor $\widetilde{\beta}(q) = \beta_0 + (1 - \beta_0)\,\beta(q) \in 
[\beta_0, 1]$ with floor parameter $\beta_0 \in (0, 1)$.

\noindent\textbf{Causal Coverage.}
The causal coverage $\Omega(q) \in [0, 1]$ aggregates the count $k$, the 
mean confidence $\bar{\Phi}(q) = \tfrac{1}{k}\sum_{\pi \in 
\widehat{\Pi}_q^{+} \cup \widehat{\Pi}_q^{-}} \Phi(\pi, q)$ 
($0$ when $k=0$), the reliability $\kappa(k)$, and the floored 
balance $\widetilde{\beta}(q)$:
\begin{equation}
\Omega(q) = \min\!(\frac{k \cdot \bar{\Phi}(q) \cdot \kappa(k) \cdot 
\widetilde{\beta}(q)}{Z},\; 1),
\end{equation}
where $Z > 0$ is a predefined normalization constant. High $\Omega(q)$ 
corresponds to abundant, strong, and balanced causal evidence, whereas 
low $\Omega(q)$ corresponds to sparse, weak, or one-sided evidence.


\noindent\textbf{Adaptive Fusion Weight.}
A coverage-driven sigmoid maps $\Omega(q)$ to
$\alpha(q) \in [0, \alpha_0]$:
\begin{equation}
\alpha(q) = \mathbbm{1}[k > 0] \cdot
\min\!\bigl(2\alpha_b\,\sigma\!\bigl(\zeta\,(\Omega(q) - \Omega_0)\bigr),\; \alpha_0\bigr),
\end{equation}
where $\mathbbm{1}[\cdot]$ is an indicator and $\sigma$ is the sigmoid;
$\alpha_b>0$ is the mapping scale, $0<\alpha_0\leq 1$ is the fusion
upper bound, $\Omega_0$ is the coverage activation threshold, and
$\zeta>0$ is the sigmoid-transition sharpness.

\noindent\textbf{Final Forecast.}
The final probability $\hat{p} \in [0, 1]$ is the convex combination 
weighted by $\alpha(q)$:
\begin{equation}
\hat{p} = \alpha(q) \cdot P_{\mathrm{causal}}(q) + 
\bigl(1 - \alpha(q)\bigr) \cdot P_{\mathrm{llm}}(q).
\end{equation}
The causal contribution is zero when no chain is retained and otherwise
increases with causal coverage up to $\alpha_0$, while the complementary
LLM contribution decreases accordingly.

\par\noindent\textbf{Proposition 3.} \textit{Adaptive fusion using causal coverage and directional balance down-weights the causal estimate when evidence is unreliable, and can yield lower squared error than fixed weighting.}
\vspace{-3mm}
\begin{proof}
We provide experimental results in Section~\ref{sec:exp_alpha} and 
theoretical proofs in Appendix~\ref{app:proof3}.
\end{proof}

\section{Experiments}
\label{sec:experiments}

In this section, we answer these following research questions (RQ): \textbf{RQ1:} whether \textbf{\textsc{Chain}} outperforms the baselines? \textbf{RQ2:} its generalization to out-of-distribution (OOD) datasets? \textbf{RQ3:} its components' contributions to overall performance? \textbf{RQ4:} whether structural calibration outperforms post-hoc methods? \textbf{RQ5:} whether the adaptive fusion weight $\alpha$ outperforms any fixed alternative?

\subsection{Experimental Setup}
\label{sec:exp_setup}

\textbf{Datasets.}
We use eight datasets: 
\textbf{AI-Futures}~\citep{aifutures2025}, 
\textbf{Metaculus}~\citep{chandak2026scalingopenendedreasoningpredict}, 
\textbf{Polymarket}~\citep{polymarket}, 
\textbf{Future-as-Label (FaL)}~\citep{turtel2026future}, 
\textbf{Clinical Trial Outcomes (CTO)}~\citep{gao2024automatically}, 
\textbf{FOReCAst}~\citep{yuan2025forecast}, 
\textbf{Golf-Forecasting}~\citep{golfforecasting}, and 
\textbf{KalshiBench}~\citep{nel2025largelanguagemodelsknow}.
More details are shown in Appendix~\ref{app:datasets}.

\textbf{Baselines.}
We compare \textbf{\textsc{Chain}} with 13 baselines: 
\textbf{LLM Direct}, 
\textbf{LLM CoT}, 
\textbf{AutoCast}, 
\textbf{NaiveRAG}, 
\textbf{GraphRAG}, 
\textbf{LightRAG}, 
\textbf{HippoRAG}, 
\textbf{HyperGraphRAG}, 
\textbf{TempScaling}, 
\textbf{PlattScaling}, 
\textbf{IsotonicRegression}, 
\textbf{Histogram Binning}, and 
\textbf{ConformalAdj}.
More details are illustrated in Appendix~\ref{app:baselines}.

\textbf{Evaluation Metrics.}
We use seven evaluation metrics, including \textbf{Expected Calibration Error (ECE)}~\citep{guo2017calibration}, \textbf{Adaptive Calibration Error (ACE)}~\citep{nixon2019measuring}, \textbf{Maximum Calibration Error (MCE)}~\citep{guo2017calibration}, \textbf{Reliability (Rel)}~\citep{murphy1973new}, \textbf{Negative Log-Likelihood (NLL)}, \textbf{Brier Score}~\citep{glenn1950verification}, and \textbf{Accuracy (Acc)}.
More details see Appendix~\ref{app:metrics}.

\textbf{Implementation.}
We construct the causal-temporal hypergraph via \textbf{GPT-4o-mini}~\citep{hurst2024gpt} and evaluate \textbf{\textsc{Chain}} three LLMs, including \textbf{GPT-4o-mini}, \textbf{Gemini-2.0-Flash}~\citep{kavukcuoglu2025gemini2}, and \textbf{DeepSeek-V3}~\citep{deepseekai2025deepseekv3technicalreport}. More details are illustrated in Appendix~\ref{app:implementation}.

\definecolor{grpDirect}{rgb}{0.9608,0.9725,0.9804}
\definecolor{grpRetrieval}{rgb}{0.9000,0.9725,0.9843}
\definecolor{grpPosthoc}{rgb}{0.9098,0.9569,0.9882}
\definecolor{grpChain}{rgb}{0.9608,0.9098,0.9686}

\begin{table*}[t]
\caption{
Main results of \textbf{\textsc{Chain}} and the baselines across four in-distribution forecasting datasets, which are reported on three LLMs with \textbf{best} in bold. All values are in \%.}
\centering
\fontsize{7.3pt}{8pt}\selectfont
\setlength{\tabcolsep}{0.68mm}{
\vspace{-3mm}
\begin{tabular}{lccc|ccc|ccc|ccc|ccc}
\toprule
\multirow{2.5}{*}{\textbf{Method}} & \multicolumn{3}{c}{\textbf{AI-Futures}} & \multicolumn{3}{c}{\textbf{Metaculus}} & \multicolumn{3}{c}{\textbf{Polymarket}} & \multicolumn{3}{c}{\textbf{Future-as-Label}} & \multicolumn{3}{c}{\textbf{Average}} \\
\cmidrule(lr){2-4} \cmidrule(lr){5-7} \cmidrule(lr){8-10} \cmidrule(lr){11-13} \cmidrule(lr){14-16}
& \textbf{ECE$\downarrow$} & \textbf{Brier$\downarrow$} & \textbf{Acc$\uparrow$} & \textbf{ECE$\downarrow$} & \textbf{Brier$\downarrow$} & \textbf{Acc$\uparrow$} & \textbf{ECE$\downarrow$} & \textbf{Brier$\downarrow$} & \textbf{Acc$\uparrow$} & \textbf{ECE$\downarrow$} & \textbf{Brier$\downarrow$} & \textbf{Acc$\uparrow$} & \textbf{ECE$\downarrow$} & \textbf{Brier$\downarrow$} & \textbf{Acc$\uparrow$} \\
\midrule
\multicolumn{16}{c}{\textbf{\textit{GPT-4o-mini}}} \\
\midrule
\rowcolor{grpDirect} LLM Direct & 15.00 & 25.69 & 59.38 & 22.27 & 28.79 & 52.34 & 20.04 & 23.19 & 70.31 & 21.29 & 29.97 & 53.91 & 19.65 & 26.91 & 58.98 \\
\rowcolor{grpDirect} LLM CoT & 12.89 & 20.27 & 71.09 & 14.22 & 23.58 & 64.06 & 17.42 & 24.88 & 64.06 & 22.62 & 29.74 & 51.56 & 16.79 & 24.62 & 62.70 \\
\rowcolor{grpDirect} AutoCast & 32.85 & 32.49 & 43.75 & 37.77 & 37.40 & 42.19 & 22.38 & 25.56 & 68.75 & 34.92 & 35.30 & 49.22 & 31.98 & 32.69 & 50.98 \\
\rowcolor{grpRetrieval} NaiveRAG & 26.99 & 28.79 & 43.75 & 28.71 & 30.72 & 46.88 & 14.18 & 23.96 & 68.75 & 36.64 & 36.62 & 40.62 & 26.63 & 30.02 & 50.00 \\
\rowcolor{grpRetrieval} GraphRAG & 33.05 & 34.39 & 39.84 & 25.35 & 28.81 & 52.34 & 18.67 & 23.74 & 68.75 & 33.83 & 36.26 & 41.41 & 27.72 & 30.80 & 50.59 \\
\rowcolor{grpRetrieval} LightRAG & 36.29 & 36.38 & 41.41 & 28.44 & 29.67 & 51.56 & 25.94 & 25.24 & 67.19 & 21.68 & 27.19 & 60.16 & 28.09 & 29.62 & 55.08 \\
\rowcolor{grpRetrieval} HippoRAG & 20.72 & 24.80 & 59.38 & 28.59 & 32.64 & 49.22 & 16.45 & 23.22 & 70.31 & 29.69 & 32.77 & 53.12 & 23.86 & 28.36 & 58.01 \\
\rowcolor{grpRetrieval} HyperGraphRAG & 32.70 & 34.71 & 45.31 & 31.64 & 32.46 & 46.09 & 22.70 & 25.69 & 67.19 & 25.55 & 28.84 & 57.03 & 28.14 & 30.43 & 53.91 \\
\rowcolor{grpPosthoc} TempScaling & 14.14 & 23.84 & 59.38 & 24.00 & 29.69 & 52.34 & 18.15 & 22.80 & 70.31 & 16.99 & 27.57 & 53.91 & 18.32 & 25.98 & 58.98 \\
\rowcolor{grpPosthoc} PlattScaling & 31.84 & 31.77 & 33.59 & 24.98 & 29.67 & 52.34 & 18.24 & 23.12 & 70.31 & 27.37 & 30.99 & 42.19 & 25.61 & 28.89 & 49.61 \\
\rowcolor{grpPosthoc} IsotonicReg & 33.06 & 33.11 & 39.06 & 22.45 & 29.40 & 52.34 & 18.33 & 24.91 & 70.31 & 33.41 & 36.40 & 42.19 & 26.81 & 30.96 & 50.98 \\
\rowcolor{grpPosthoc} HistogramBin & 32.67 & 34.60 & 39.06 & 26.31 & 30.84 & 50.78 & 17.23 & 22.93 & 71.09 & 33.68 & 37.83 & 43.75 & 27.47 & 31.55 & 51.17 \\
\rowcolor{grpPosthoc} ConformalAdj & 15.41 & 24.08 & 59.38 & 15.04 & 24.81 & 52.34 & 14.83 & 23.58 & 70.31 & 14.72 & 24.97 & 53.91 & 15.00 & 24.36 & 58.98 \\
\rowcolor{grpChain} \textbf{CHAIN (ours)} & \textbf{9.45} & \textbf{17.15} & \textbf{77.34} & \textbf{7.52} & \textbf{18.77} & \textbf{71.88} & \textbf{13.74} & \textbf{19.54} & \textbf{71.88} & \textbf{10.20} & \textbf{19.68} & \textbf{67.97} & \textbf{10.23} & \textbf{18.79} & \textbf{72.27} \\
\midrule
\multicolumn{16}{c}{\textbf{\textit{Gemini-2.0-Flash}}} \\
\midrule
\rowcolor{grpDirect} LLM Direct & 9.18 & 20.32 & 67.97 & 21.89 & 25.57 & 58.59 & 15.39 & 22.57 & 71.88 & 28.52 & 33.23 & 49.22 & 18.74 & 25.42 & 61.91 \\
\rowcolor{grpDirect} LLM CoT & 9.18 & 20.41 & 67.19 & 13.43 & 21.98 & 65.62 & 15.00 & 20.46 & 71.88 & 19.77 & 28.34 & 53.91 & 14.35 & 22.80 & 64.65 \\
\rowcolor{grpDirect} AutoCast & 13.63 & 23.30 & 60.94 & 13.91 & 24.78 & 58.59 & 21.48 & 25.10 & 35.94 & 15.12 & 26.80 & 57.03 & 16.04 & 25.00 & 53.12 \\
\rowcolor{grpRetrieval} NaiveRAG & 18.44 & 25.04 & 42.97 & 13.91 & 22.76 & 63.28 & 19.69 & 25.35 & 32.03 & 14.70 & 25.28 & 57.81 & 16.68 & 24.61 & 49.02 \\
\rowcolor{grpRetrieval} GraphRAG & 12.15 & 22.01 & 57.81 & 14.97 & 23.56 & 59.38 & 23.19 & 24.27 & 68.75 & 19.27 & 29.88 & 45.31 & 17.39 & 24.93 & 57.81 \\
\rowcolor{grpRetrieval} LightRAG & 14.96 & 23.03 & 57.81 & 14.49 & 21.48 & 66.41 & 29.34 & 26.86 & 46.09 & 15.82 & 24.76 & 58.59 & 18.65 & 24.03 & 57.23 \\
\rowcolor{grpRetrieval} HippoRAG & 10.39 & 19.49 & 67.19 & 16.02 & 23.94 & 60.94 & 26.80 & 26.00 & 39.84 & 17.30 & 25.51 & 59.38 & 17.63 & 23.73 & 56.84 \\
\rowcolor{grpRetrieval} HyperGraphRAG & 13.40 & 24.31 & 56.25 & 14.45 & 22.51 & 64.84 & 23.75 & 24.79 & 35.94 & 17.15 & 23.62 & 56.25 & 17.19 & 23.81 & 53.32 \\
\rowcolor{grpPosthoc} TempScaling & 8.13 & 20.21 & 67.97 & 15.62 & 23.28 & 58.59 & 16.55 & 22.78 & 71.88 & 18.67 & 28.68 & 49.22 & 14.74 & 23.74 & 61.91 \\
\rowcolor{grpPosthoc} PlattScaling & 26.96 & 27.19 & 43.75 & 20.71 & 26.29 & 57.81 & 14.90 & 21.80 & 71.88 & 28.13 & 33.68 & 44.53 & 22.67 & 27.24 & 54.49 \\
\rowcolor{grpPosthoc} IsotonicReg & 26.27 & 27.48 & 54.69 & 22.80 & 28.37 & 64.06 & 23.43 & 25.48 & 43.75 & 29.47 & 35.52 & 44.53 & 25.49 & 29.21 & 51.76 \\
\rowcolor{grpPosthoc} HistogramBin & 25.82 & 28.70 & 54.69 & 24.67 & 34.65 & 53.91 & 26.49 & 29.39 & 44.53 & 29.39 & 35.89 & 42.19 & 26.59 & 32.16 & 48.83 \\
\rowcolor{grpPosthoc} ConformalAdj & 13.29 & 22.44 & 67.97 & 13.20 & 23.61 & 58.59 & 17.11 & 23.31 & 71.88 & 11.74 & 25.53 & 49.22 & 13.83 & 23.72 & 61.91 \\
\rowcolor{grpChain} \textbf{CHAIN (ours)} & \textbf{7.10} & \textbf{17.19} & \textbf{73.44} & \textbf{7.05} & \textbf{17.73} & \textbf{74.22} & \textbf{14.87} & \textbf{18.96} & \textbf{75.00} & \textbf{9.72} & \textbf{19.96} & \textbf{68.75} & \textbf{9.68} & \textbf{18.46} & \textbf{72.85} \\
\midrule
\multicolumn{16}{c}{\textbf{\textit{DeepSeek-V3}}} \\
\midrule
\rowcolor{grpDirect} LLM Direct & 9.34 & 21.17 & 68.75 & 20.90 & 26.69 & 54.69 & 15.43 & 20.71 & 74.22 & 19.69 & 29.16 & 54.69 & 16.34 & 24.43 & 63.09 \\
\rowcolor{grpDirect} LLM CoT & 11.48 & 21.62 & 66.41 & 12.49 & 18.86 & 71.88 & 14.70 & 20.63 & 71.88 & 15.72 & 26.99 & 59.38 & 13.60 & 22.03 & 67.38 \\
\rowcolor{grpDirect} AutoCast & 15.43 & 24.37 & 53.91 & 19.73 & 27.06 & 56.25 & 20.43 & 24.73 & 43.75 & 17.58 & 28.47 & 51.56 & 18.29 & 26.16 & 51.37 \\
\rowcolor{grpRetrieval} NaiveRAG & 11.91 & 21.31 & 63.28 & 19.88 & 25.19 & 57.81 & 16.95 & 23.54 & 58.59 & 24.34 & 29.42 & 47.66 & 18.27 & 24.87 & 56.84 \\
\rowcolor{grpRetrieval} GraphRAG & 16.99 & 23.57 & 58.59 & 18.98 & 25.64 & 58.59 & 15.12 & 20.44 & 73.44 & 24.34 & 30.76 & 47.66 & 18.86 & 25.10 & 59.57 \\
\rowcolor{grpRetrieval} LightRAG & 17.97 & 24.14 & 58.59 & 17.03 & 22.89 & 64.06 & 15.90 & 23.28 & 66.41 & 19.61 & 26.66 & 57.03 & 17.63 & 24.24 & 61.52 \\
\rowcolor{grpRetrieval} HippoRAG & 10.39 & 19.24 & 69.53 & 16.76 & 23.86 & 64.06 & 15.66 & 20.86 & 71.09 & 11.76 & 20.88 & 67.97 & 13.64 & 21.21 & 68.16 \\
\rowcolor{grpRetrieval} HyperGraphRAG & 19.73 & 25.90 & 55.47 & 23.98 & 25.02 & 59.38 & 16.09 & 20.78 & 71.09 & 12.77 & 19.43 & 71.09 & 18.14 & 22.78 & 64.26 \\
\rowcolor{grpPosthoc} TempScaling & 8.85 & 20.79 & 68.75 & 16.05 & 24.44 & 54.69 & 16.42 & 21.58 & 74.22 & 13.41 & 26.23 & 54.69 & 13.68 & 23.26 & 63.09 \\
\rowcolor{grpPosthoc} PlattScaling & 25.71 & 26.76 & 64.84 & 24.71 & 28.75 & 53.12 & 13.82 & 20.86 & 75.00 & 28.54 & 31.79 & 44.53 & 23.20 & 27.04 & 59.38 \\
\rowcolor{grpPosthoc} IsotonicReg & 24.72 & 25.57 & 68.75 & 25.48 & 30.33 & 54.69 & 27.55 & 26.89 & 33.59 & 31.05 & 34.61 & 44.53 & 27.20 & 29.35 & 50.39 \\
\rowcolor{grpPosthoc} HistogramBin & 30.53 & 27.63 & 53.12 & 26.62 & 30.51 & 53.12 & 29.59 & 28.42 & 32.81 & 32.65 & 35.49 & 50.78 & 29.85 & 30.51 & 47.46 \\
\rowcolor{grpPosthoc} ConformalAdj & 13.53 & 22.80 & 68.75 & 13.27 & 24.12 & 54.69 & 18.96 & 23.01 & 74.22 & 11.72 & 24.81 & 54.69 & 14.37 & 23.68 & 63.09 \\
\rowcolor{grpChain} \textbf{CHAIN (ours)} & \textbf{8.54} & \textbf{17.34} & \textbf{76.56} & \textbf{9.72} & \textbf{18.05} & \textbf{73.44} & \textbf{13.74} & \textbf{17.41} & \textbf{79.69} & \textbf{10.09} & \textbf{18.82} & \textbf{72.66} & \textbf{10.52} & \textbf{17.91} & \textbf{75.59} \\
\bottomrule
\end{tabular}}
\label{T1}
\vspace{-0.1mm}
\end{table*}

\definecolor{grpDirect}{rgb}{0.9608,0.9725,0.9804}
\definecolor{grpRetrieval}{rgb}{0.9000,0.9725,0.9843}
\definecolor{grpPosthoc}{rgb}{0.9098,0.9569,0.9882}
\definecolor{grpChain}{rgb}{0.9608,0.9098,0.9686}
\begin{table*}[t]
\caption{
Performance comparison of \textbf{\textsc{Chain}} and the baselines on 4 OOD datasets across 2 LLMs.}
\centering
\fontsize{7.3pt}{8pt}\selectfont
\setlength{\tabcolsep}{0.68mm}{
\vspace{-3mm}
\begin{tabular}{lccc|ccc|ccc|ccc|ccc}
\toprule
\multirow{2.5}{*}{\textbf{Method}} & \multicolumn{3}{c}{\textbf{CTO}} & \multicolumn{3}{c}{\textbf{FOReCAst}} & \multicolumn{3}{c}{\textbf{Golf-Forecasting}} & \multicolumn{3}{c}{\textbf{KalshiBench}} & \multicolumn{3}{c}{\textbf{Average}} \\
\cmidrule(lr){2-4} \cmidrule(lr){5-7} \cmidrule(lr){8-10} \cmidrule(lr){11-13} \cmidrule(lr){14-16}
& \textbf{ECE$\downarrow$} & \textbf{Brier$\downarrow$} & \textbf{Acc$\uparrow$} & \textbf{ECE$\downarrow$} & \textbf{Brier$\downarrow$} & \textbf{Acc$\uparrow$} & \textbf{ECE$\downarrow$} & \textbf{Brier$\downarrow$} & \textbf{Acc$\uparrow$} & \textbf{ECE$\downarrow$} & \textbf{Brier$\downarrow$} & \textbf{Acc$\uparrow$} & \textbf{ECE$\downarrow$} & \textbf{Brier$\downarrow$} & \textbf{Acc$\uparrow$} \\
\midrule
\multicolumn{16}{c}{\textbf{\textit{GPT-4o-mini}}} \\
\midrule
\rowcolor{grpDirect} LLM Direct & 18.83 & 27.61 & 55.47 & 12.27 & 23.43 & 67.19 & 19.77 & 27.18 & 53.91 & 20.04 & 27.81 & 59.38 & 17.72 & 26.51 & 58.98 \\
\rowcolor{grpDirect} LLM CoT & 18.12 & 26.88 & 56.25 & 16.37 & 24.78 & 60.16 & 11.25 & 25.13 & 60.16 & 16.51 & 26.71 & 57.81 & 15.56 & 25.88 & 58.59 \\
\rowcolor{grpDirect} AutoCast & 42.50 & 43.07 & 42.19 & 26.09 & 28.93 & 59.38 & 45.35 & 43.72 & 47.66 & 44.30 & 44.26 & 41.41 & 39.56 & 39.99 & 47.66 \\
\rowcolor{grpRetrieval} NaiveRAG & 30.04 & 32.62 & 42.97 & 30.35 & 33.02 & 50.00 & 24.14 & 30.31 & 48.44 & 30.94 & 34.34 & 44.53 & 28.87 & 32.57 & 46.48 \\
\rowcolor{grpRetrieval} GraphRAG & 23.79 & 29.67 & 48.44 & 22.27 & 28.45 & 57.81 & 23.16 & 30.17 & 48.44 & 25.70 & 31.29 & 50.00 & 23.73 & 29.90 & 51.17 \\
\rowcolor{grpRetrieval} LightRAG & 31.60 & 33.38 & 42.97 & 13.79 & 23.92 & 68.75 & 30.00 & 34.43 & 45.31 & 27.62 & 29.51 & 55.47 & 25.75 & 30.31 & 53.12 \\
\rowcolor{grpRetrieval} HippoRAG & 27.27 & 30.53 & 44.53 & 18.63 & 25.82 & 64.06 & 31.64 & 34.07 & 46.09 & 33.20 & 34.35 & 50.78 & 27.69 & 31.19 & 51.37 \\
\rowcolor{grpRetrieval} HyperGraphRAG & 33.09 & 34.63 & 42.19 & 21.41 & 28.23 & 59.38 & 31.41 & 34.32 & 45.31 & 27.54 & 31.56 & 52.34 & 28.36 & 32.19 & 49.80 \\
\rowcolor{grpPosthoc} TempScaling & 9.57 & 24.86 & 55.47 & 10.52 & 22.38 & 67.19 & 13.62 & 24.82 & 53.91 & 13.45 & 24.76 & 59.38 & 11.79 & 24.21 & 58.98 \\
\rowcolor{grpPosthoc} PlattScaling & 31.70 & 34.35 & 42.19 & 31.50 & 32.99 & 46.88 & 34.90 & 36.12 & 44.53 & 37.15 & 36.89 & 39.84 & 33.81 & 35.09 & 43.36 \\
\rowcolor{grpPosthoc} IsotonicReg & 29.93 & 33.28 & 42.19 & 34.82 & 34.56 & 46.88 & 32.32 & 34.55 & 44.53 & 37.69 & 38.53 & 39.84 & 33.69 & 35.23 & 43.36 \\
\rowcolor{grpPosthoc} HistogramBin & 34.10 & 35.92 & 46.09 & 31.42 & 31.50 & 58.59 & 31.24 & 33.71 & 50.78 & 33.09 & 36.25 & 50.78 & 32.46 & 34.34 & 51.56 \\
\rowcolor{grpPosthoc} ConformalAdj & 4.45 & 24.53 & 55.47 & 14.43 & 23.06 & 67.19 & 8.79 & 24.59 & 53.91 & 10.73 & 24.19 & 59.38 & 9.60 & 24.09 & 58.98 \\
\rowcolor{grpChain} \textbf{CHAIN (ours)} & \textbf{4.36} & \textbf{22.63} & \textbf{64.84} & \textbf{9.95} & \textbf{19.83} & \textbf{69.53} & \textbf{8.19} & \textbf{23.64} & \textbf{62.50} & \textbf{9.85} & \textbf{23.39} & \textbf{61.72} & \textbf{8.09} & \textbf{22.37} & \textbf{64.65} \\
\midrule
\multicolumn{16}{c}{\textbf{\textit{Gemini-2.0-Flash}}} \\
\midrule
\rowcolor{grpDirect} LLM Direct & 12.66 & 23.16 & 63.28 & 9.38 & 20.00 & 71.88 & 13.67 & 26.92 & 57.81 & 21.28 & 26.87 & 60.16 & 14.25 & 24.23 & 63.28 \\
\rowcolor{grpDirect} LLM CoT & 11.88 & 23.95 & 63.28 & 11.05 & 21.98 & 67.19 & 23.01 & 30.51 & 50.00 & 19.98 & 25.99 & 60.16 & 16.48 & 25.61 & 60.16 \\
\rowcolor{grpDirect} AutoCast & 7.93 & 24.92 & 57.81 & 11.76 & 24.19 & 56.25 & 11.48 & 26.19 & 58.59 & 11.37 & 26.36 & 56.25 & 10.63 & 25.41 & 57.23 \\
\rowcolor{grpRetrieval} NaiveRAG & 9.62 & 24.56 & 55.47 & 20.10 & 26.82 & 43.75 & 5.82 & 24.05 & 55.47 & 11.73 & 25.29 & 58.59 & 11.82 & 25.18 & 53.32 \\
\rowcolor{grpRetrieval} GraphRAG & 7.70 & 24.44 & 54.69 & 9.88 & 21.63 & 69.53 & 7.30 & 24.59 & 60.16 & 18.24 & 26.26 & 53.12 & 10.78 & 24.23 & 59.38 \\
\rowcolor{grpRetrieval} LightRAG & 11.87 & 26.09 & 50.00 & 17.54 & 22.76 & 68.75 & 6.84 & 25.14 & 53.12 & 13.44 & 23.35 & 56.25 & 12.42 & 24.34 & 57.03 \\
\rowcolor{grpRetrieval} HippoRAG & 12.89 & 24.30 & 55.47 & 16.60 & 21.09 & 70.31 & 12.28 & 28.16 & 43.75 & 21.25 & 25.65 & 50.78 & 15.76 & 24.80 & 55.08 \\
\rowcolor{grpRetrieval} HyperGraphRAG & 7.07 & 23.62 & 59.38 & 17.30 & 24.49 & 61.72 & 7.73 & 24.45 & 58.59 & 12.03 & 24.20 & 56.25 & 11.04 & 24.19 & 58.98 \\
\rowcolor{grpPosthoc} TempScaling & 4.71 & 22.27 & 63.28 & 10.57 & 19.74 & 71.88 & 7.40 & 24.84 & 57.81 & 14.47 & 23.56 & 60.16 & 9.29 & 22.60 & 63.28 \\
\rowcolor{grpPosthoc} PlattScaling & 29.91 & 31.62 & 42.19 & 27.91 & 28.22 & 46.88 & 31.53 & 34.05 & 44.53 & 31.83 & 33.08 & 39.84 & 30.30 & 31.74 & 43.36 \\
\rowcolor{grpPosthoc} IsotonicReg & 29.65 & 32.14 & 44.53 & 23.38 & 27.25 & 56.25 & 31.80 & 34.88 & 45.31 & 33.60 & 34.62 & 48.44 & 29.61 & 32.22 & 48.63 \\
\rowcolor{grpPosthoc} HistogramBin & 45.71 & 42.75 & 41.41 & 39.90 & 37.49 & 43.75 & 35.27 & 38.80 & 50.78 & 39.51 & 41.14 & 47.66 & 40.10 & 40.04 & 45.90 \\
\rowcolor{grpPosthoc} ConformalAdj & 7.25 & 23.33 & 63.28 & 15.49 & 21.95 & 71.88 & 5.63 & 24.45 & 57.81 & 9.70 & 23.59 & 60.16 & 9.51 & 23.33 & 63.28 \\
\rowcolor{grpChain} \textbf{CHAIN (ours)} & \textbf{4.54} & \textbf{22.15} & \textbf{64.84} & \textbf{8.05} & \textbf{18.95} & \textbf{74.22} & \textbf{5.41} & \textbf{22.42} & \textbf{63.28} & \textbf{9.63} & \textbf{23.17} & \textbf{65.62} & \textbf{6.91} & \textbf{21.67} & \textbf{66.99} \\
\bottomrule
\end{tabular}}
\label{T2}
\vspace{-0.1mm}
\end{table*}

\subsection{Main Results (RQ1)}
\label{sec:exp_main}

Table~\ref{T1} shows the main results for three LLMs across four in-distribution datasets. \textit{(i)} \textbf{Joint calibration and accuracy.} \textbf{\textsc{Chain}} has the lowest ECE and Brier and the highest accuracy, improving calibration without sacrificing performance. \textit{(ii)} \textbf{Prompting and retrieval remain insufficient.} Direct prompting has higher calibration error, while retrieval gains vary, so added evidence alone does not ensure calibrated forecasts. \textit{(iii)} \textbf{Post-hoc correction leaves a gap.} Post-hoc methods retain higher calibration error and sometimes reduce accuracy, showing that output correction alone is insufficient. \textit{(iv)} \textbf{Backbone-agnostic gains.} \textbf{\textsc{Chain}} leads across all three backbones, with lower average ECE and Brier and higher accuracy, showing consistent gains across these evaluated models.

\begin{figure}[t]
\centering
\begin{tabular}{@{}c@{}c@{}c@{}}
\includegraphics[width=0.32\linewidth]{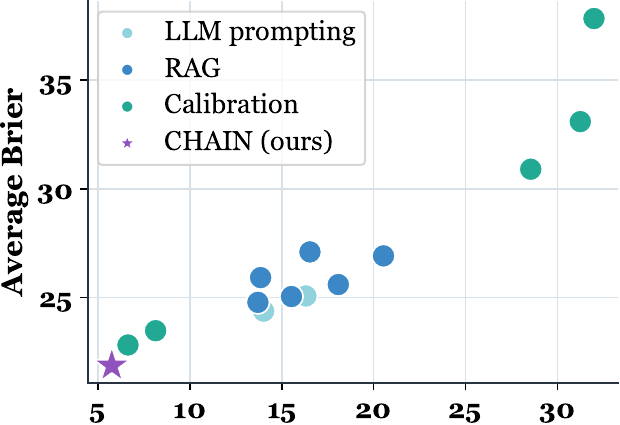} &
\includegraphics[width=0.32\linewidth]{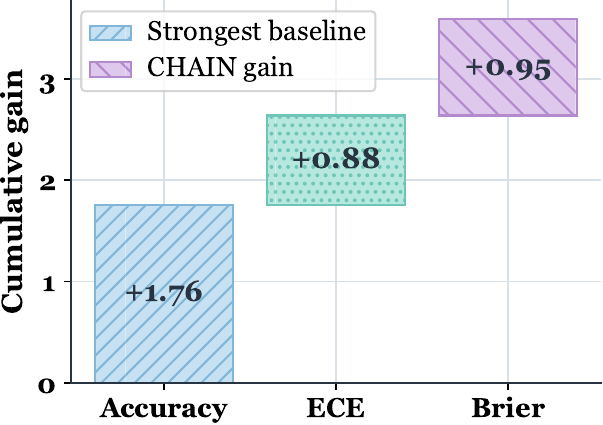} &
\includegraphics[width=0.32\linewidth]{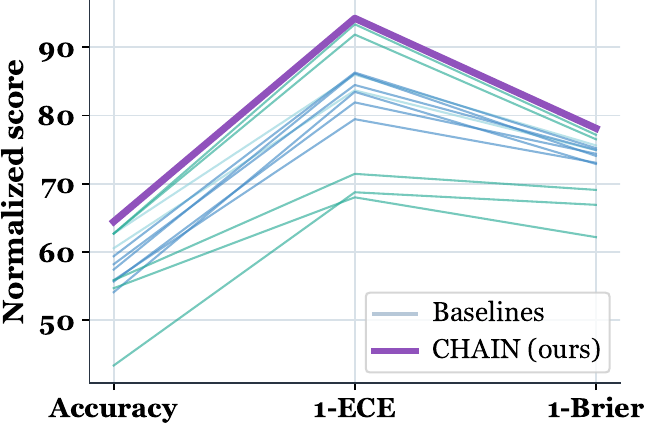} \\
\small (a) Pareto calibration trade-off &
\small (b) Cumulative gains &
\small (c) Cross-metric performance
\end{tabular}
\vspace{-1.6mm}
\caption{Calibration and predictive performance of \textbf{\textsc{Chain}} for four OOD datasets on DeepSeek-V3.}
\label{fig:ood_results}
\end{figure}

\subsection{Out-of-Distribution Generalization (RQ2)}
\label{sec:exp_ood}

Table~\ref{T2} and Figure~\ref{fig:ood_results} show the results for the same three LLMs across four out-of-distribution datasets. \textit{(i)} \textbf{Calibration transfers to OOD.} \textbf{\textsc{Chain}} attains the lowest ECE and Brier, maintaining its calibration advantage as forecasting tasks move beyond the in-distribution benchmarks. \textit{(ii)} \textbf{Accuracy remains highest under shift.} \textbf{\textsc{Chain}} leads in the OOD results, showing that lower calibration error does not come at the expense of prediction accuracy under distribution shift. \textit{(iii)} \textbf{Gains persist across backbones.} Calibration and accuracy advantages recur across models, indicating that the improvements are not tied to one model's forecasting capability or one source. \textit{(iv)} \textbf{Joint calibration--accuracy gains.} Lower average ECE and Brier accompany higher accuracy, indicating balanced improvements in both the quality of probability estimates and the correctness of event predictions.

\subsection{Ablation Study (RQ3)}
\label{sec:exp_ablation}
As shown in Table~\ref{T3}, \textbf{\textsc{Chain}}'s three components each contribute to calibration and prediction accuracy:
\textit{(i)} \textbf{CTVF and Noisy-OR improve calibration.} Removing either component raises ECE on every dataset, showing that temporal-causal validity weighting and direction-aware aggregation both improve the quality of predicted probabilities.
\textit{(ii)} \textbf{Noisy-OR safeguards OOD accuracy.} Removing Noisy-OR causes the largest accuracy drop on each OOD dataset, demonstrating the role of direction-aware aggregation in maintaining OOD prediction accuracy.
\textit{(iii)} \textbf{Adaptive $\alpha$ tightens residual calibration.} Replacing per-instance fusion with a fixed weight raises ECE on every dataset, showing that question-level reweighting improves calibration and predictive accuracy.

\definecolor{grpChain}{rgb}{0.9608,0.9098,0.9686}
\begin{table*}[t]
\caption{
Ablation study of \textbf{\textsc{Chain}}, where \textit{w/o} means without the
 corresponding component.
}
\centering
\fontsize{7.3pt}{8pt}\selectfont
\setlength{\tabcolsep}{1.66mm}

\begin{tabular}{lccc|ccc|ccc|ccc}
\toprule
\multirow{2.5}{*}{\textbf{Method}} & \multicolumn{3}{c}{\textbf{AI-Futures}} & \multicolumn{3}{c}{\textbf{Metaculus}} & \multicolumn{3}{c}{\textbf{Polymarket}} & \multicolumn{3}{c}{\textbf{Future-as-Label}} \\
\cmidrule(lr){2-4} \cmidrule(lr){5-7} \cmidrule(lr){8-10} \cmidrule(lr){11-13}
& \textbf{ECE$\downarrow$} & \textbf{Brier$\downarrow$} & \textbf{Acc$\uparrow$} & \textbf{ECE$\downarrow$} & \textbf{Brier$\downarrow$} & \textbf{Acc$\uparrow$} & \textbf{ECE$\downarrow$} & \textbf{Brier$\downarrow$} & \textbf{Acc$\uparrow$} & \textbf{ECE$\downarrow$} & \textbf{Brier$\downarrow$} & \textbf{Acc$\uparrow$} \\
\midrule
\textit{w/o CTVF} & 11.68 & 19.43 & 73.44 & 10.65 & 19.59 & 72.66 & 19.73 & 17.54 & 75.00 & 10.91 & 23.15 & 63.28 \\
\textit{w/o Noisy-OR} & 12.79 & 19.37 & 75.78 & 12.61 & 19.72 & 71.09 & 16.22 & 17.86 & 78.91 & 11.03 & 19.78 & 71.88 \\
\textit{w/o Adaptive $\alpha$} & 11.25 & 17.66 & 75.78 & 10.40 & 18.27 & 72.66 & 14.39 & 17.66 & 77.34 & 10.16 & 19.68 & 71.88 \\
\rowcolor{grpChain} \textbf{CHAIN (Full)} & \textbf{8.54} & \textbf{17.34} & \textbf{76.56} & \textbf{9.72} & \textbf{18.05} & \textbf{73.44} & \textbf{13.74} & \textbf{17.41} & \textbf{79.69} & \textbf{10.09} & \textbf{18.82} & \textbf{72.66} \\
\midrule
\multirow{2.5}{*}{\textbf{Method}} & \multicolumn{3}{c}{\textbf{CTO}} & \multicolumn{3}{c}{\textbf{FOReCAst}} & \multicolumn{3}{c}{\textbf{Golf-Forecasting}} & \multicolumn{3}{c}{\textbf{KalshiBench}} \\
\cmidrule(lr){2-4} \cmidrule(lr){5-7} \cmidrule(lr){8-10} \cmidrule(lr){11-13}
& \textbf{ECE$\downarrow$} & \textbf{Brier$\downarrow$} & \textbf{Acc$\uparrow$} & \textbf{ECE$\downarrow$} & \textbf{Brier$\downarrow$} & \textbf{Acc$\uparrow$} & \textbf{ECE$\downarrow$} & \textbf{Brier$\downarrow$} & \textbf{Acc$\uparrow$} & \textbf{ECE$\downarrow$} & \textbf{Brier$\downarrow$} & \textbf{Acc$\uparrow$} \\
\midrule
\textit{w/o CTVF} & 7.24 & 22.92 & 60.94 & 8.73 & 17.95 & 73.44 & 7.44 & 24.59 & 57.81 & 13.33 & 24.42 & 60.94 \\
\textit{w/o Noisy-OR} & 5.02 & 23.47 & 60.16 & 8.43 & 18.87 & 72.66 & 7.76 & 24.02 & 57.03 & 9.13 & 23.20 & 60.16 \\
\textit{w/o Adaptive $\alpha$} & 3.72 & 22.85 & 61.72 & 7.35 & 18.17 & 73.44 & 6.52 & 24.00 & 57.81 & 9.04 & 22.98 & 61.72 \\
\rowcolor{grpChain} \textbf{CHAIN (Full)} & \textbf{3.61} & \textbf{22.73} & \textbf{62.50} & \textbf{6.98} & \textbf{17.90} & \textbf{74.22} & \textbf{5.15} & \textbf{23.91} & \textbf{58.59} & \textbf{7.25} & \textbf{22.95} & \textbf{62.50} \\
\bottomrule
\end{tabular}
\label{T3}
\vspace{-0.1mm}
\end{table*}

\begin{figure}[t]
\centering
\includegraphics[width=0.986\linewidth]{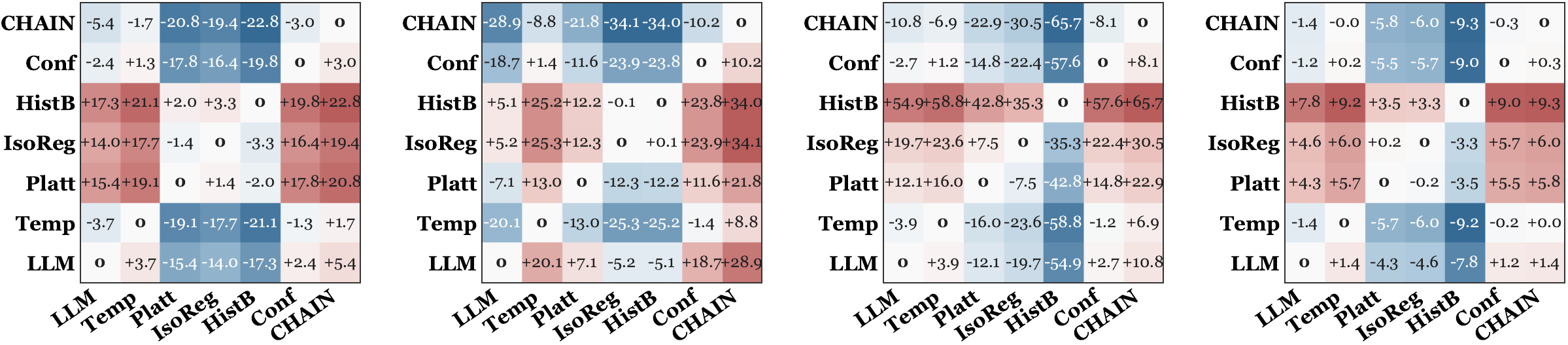}

\begin{subfigure}{0.24\linewidth}
  \caption{ACE.}\label{fig:rq4-aece}
\end{subfigure}%
\begin{subfigure}{0.24\linewidth}
  \caption{MCE.}\label{fig:rq4-mce}
\end{subfigure}%
\begin{subfigure}{0.24\linewidth}
  \caption{NLL ($\times 100$).}\label{fig:rq4-nll}
\end{subfigure}%
\begin{subfigure}{0.24\linewidth}
  \caption{Rel.}\label{fig:rq4-rel}
\end{subfigure}
\vspace{-1mm}
\caption{Comparison of \textbf{\textsc{Chain}} and post-hoc calibration methods on DeepSeek-V3 (lower is \textbf{better}).}
\label{fig:rq4-calibration}
\end{figure}

\subsection{Structural Calibration vs Post-Hoc Fitting (RQ4)}
\label{sec:exp_calibration}

\begin{wrapfigure}{r}{0.5\textwidth}
  \vspace{-9pt}
  \centering
  \includegraphics[width=1\linewidth]{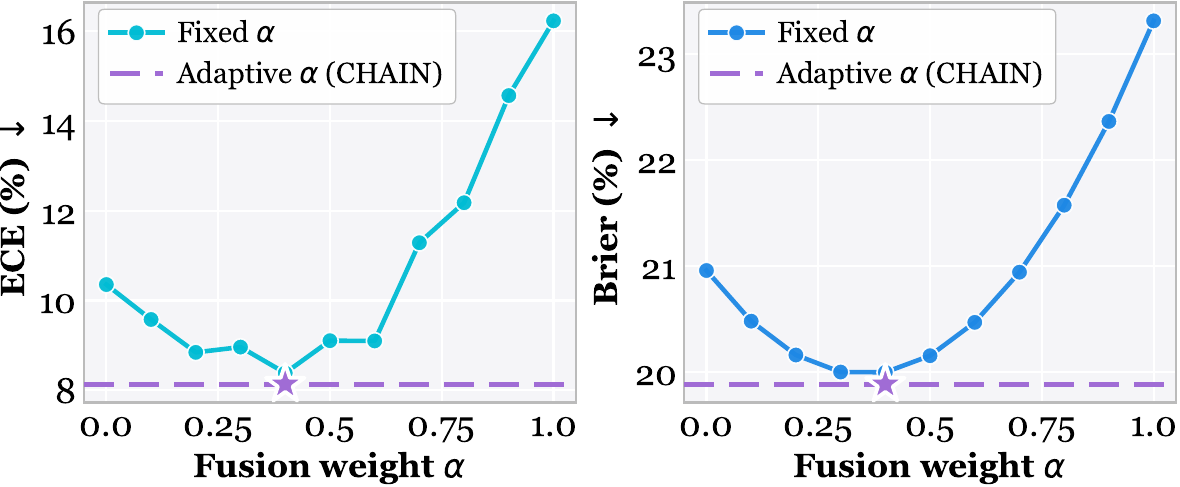}

  \begin{subfigure}{0.5\linewidth}
    \caption{ECE.}\label{fig:rq5-ece}
  \end{subfigure}%
  \begin{subfigure}{0.5\linewidth}
    \caption{Brier.}\label{fig:rq5-brier}
  \end{subfigure}
  \vspace{-3.6mm}
  \caption{DeepSeek-V3 macro-average ECE/Brier over eight ID/OOD datasets: adaptive vs. fixed $\alpha$.}
  \label{fig:rq5-alpha}
  \vspace{-3pt}
\end{wrapfigure}

Figure~\ref{fig:rq4-calibration} shows that, on the pooled eight-dataset evaluation, \textbf{\textsc{Chain}} outperforms all post-hoc baselines across all four metrics.
\textit{(i)} \textbf{Pairwise dominance.} \textbf{\textsc{Chain}} attains the lowest ACE, MCE, NLL, and Rel.
\textit{(ii)} \textbf{Worst-bin calibration error suppressed.} \textbf{\textsc{Chain}} reduces MCE relative to every post-hoc baseline, combining lower worst-bin error with improved average probability quality.
\textit{(iii)} \textbf{Post-hoc fitting can exacerbate miscalibration.} Platt, Isotonic, and Histogram Binning have higher ACE and Rel than LLM Direct, while \textbf{\textsc{Chain}} records the lowest values for both calibration metrics.

\subsection{Adaptive $\alpha$ vs Fixed (RQ5)}
\label{sec:exp_alpha}
As shown in Figure~\ref{fig:rq5-alpha}, the fixed-$\alpha$ sensitivity analysis reveals three patterns in the eight-dataset macro-average:
\textit{(i)} \textbf{Point-estimate comparison.} Adaptive $\alpha$ yields lower ECE and Brier than every evaluated fixed $\alpha$, revealing a consistent advantage over fixed fusion.
\textit{(ii)} \textbf{Fixed-weight sensitivity.} ECE and Brier vary across fixed weights, with excessively large fixed $\alpha$ degrading both metrics, revealing sensitivity to fusion choice.
\textit{(iii)} \textbf{Adaptive fusion.} These results support question-level weighting over globally fixed fusion when evidence quality differs across questions and datasets.

\section{Conclusion}
\label{sec:conclusion}

In this work, we propose \textbf{\textsc{Chain}}, a causal-temporal hypergraph framework for diagnosing and mitigating LLM miscalibration in event forecasting. Experimental results on eight datasets across multiple LLM backbones illustrate that \textbf{\textsc{Chain}} consistently improves calibration and prediction accuracy over the evaluated baselines, validating structural calibration within the prediction process.

\subsection*{AI use statement}
In this work, we used generative AI tools to reformat the experimental dataset, but not to clean the data. Apart from this task, we did not use generative AI for any other required research tasks. Generating synthetic datasets, translation, and qualitative or thematic analysis were not applicable to this work. We also used GPT-6 Astra to improve the manuscript's language, grammar, and readability and OpenAI's ChatGPT Images image-generation feature to assist with some figures. We reviewed and revised all AI-assisted outputs as necessary. We take responsibility for the final content of this work, including text, claims, or artifacts produced with the aid of generative AI.

\subsection*{Ethics statement}
This work uses public benchmark datasets without new human-subject recruitment or personal data collection. Forecasts are for research evaluation and not high-impact decisions.

\subsection*{Reproducibility statement}
The paper and appendix provide methods, prompts, proofs, datasets, evaluation settings, and extended results. Code and supporting materials are provided for reproduction.

\bibliography{custom}{}
\bibliographystyle{iclr2027_conference}

\newpage
\appendix

\section*{Appendix}

\section{Theoretical Proof}

\subsection{Proof of Proposition 1}
\label{app:proof1}

\noindent\textbf{Proposition 1.} \textit{Under normalized weighting, CTVF lowers
weighting error versus uniform weighting by favoring lower-error proximal evidence
without increasing the weighted error within either subset.}\vspace{-3mm}
\begin{proof}
To isolate the weighting effect of CTVF from subsequent chain
aggregation, let $\mathcal{E}=\{e_i\}_{i=1}^{M}$ denote the candidate evidence set,
where each $e_i$ carries an effective causal topological distance
$d_i\in\{0,1,\dots,d_{\max}\}$, obtained by capping its raw distance
at $d_{\max}$ and assigning unreachable evidence $d_i=d_{\max}$.
Let $V_i>0$ be the validity assigned by CTVF, $p_i\in[0,1]$ the
per-evidence event probability, $y\in\{0,1\}$ the binary ground truth,
and $\xi_i=|p_i-y|$ the corresponding error term.

Fix a threshold $d^{\ast}\in\{0,1,\dots,d_{\max}-1\}$ for which the
causally proximal set
$\mathcal{P}=\{i:d_i\leq d^{\ast}\}$ and the causally distal set
$\mathcal{D}=\{i:d_i>d^{\ast}\}$ are both nonempty. Let
$M_{\mathcal{P}}=|\mathcal{P}|>0$ and
$M_{\mathcal{D}}=|\mathcal{D}|>0$, so that
\begin{equation}
M=M_{\mathcal{P}}+M_{\mathcal{D}}.
\end{equation}

For any weighting scheme $\Lambda=\{\lambda_i\}_{i=1}^{M}$ satisfying
$\sum_{i=1}^{M}\lambda_i=1$ and $\lambda_i\geq 0$, define the weighting
error as
\begin{equation}
\mathcal{E}(\Lambda)=\sum_{i=1}^{M}\lambda_i\xi_i.
\end{equation}

Under uniform weighting,
\begin{equation}
\lambda_i^{\mathrm{uni}}=\frac{1}{M},
\end{equation}
and hence
\begin{equation}
\mathcal{E}(\Lambda^{\mathrm{uni}})
=
\frac{1}{M}\sum_{i=1}^{M}\xi_i
=
\frac{M_{\mathcal{P}}\bar{\xi}_{\mathcal{P}}
+M_{\mathcal{D}}\bar{\xi}_{\mathcal{D}}}{M},
\end{equation}
where
\begin{equation}
\bar{\xi}_{\mathcal{P}}
=
\frac{1}{M_{\mathcal{P}}}\sum_{i\in\mathcal{P}}\xi_i,
\qquad
\bar{\xi}_{\mathcal{D}}
=
\frac{1}{M_{\mathcal{D}}}\sum_{i\in\mathcal{D}}\xi_i
\end{equation}
denote the mean errors of the proximal and distal subsets,
respectively.

Under CTVF weighting,
\begin{equation}
\lambda_i^{\mathrm{CTVF}}=\frac{V_i}{C_V},
\qquad
C_V=\sum_{j=1}^{M}V_j.
\end{equation}
Define the mean validities of the two subsets as
\begin{equation}
\bar{V}_{\mathcal{P}}
=
\frac{1}{M_{\mathcal{P}}}\sum_{i\in\mathcal{P}}V_i,
\qquad
\bar{V}_{\mathcal{D}}
=
\frac{1}{M_{\mathcal{D}}}\sum_{i\in\mathcal{D}}V_i,
\end{equation}
and their total validities as
\begin{equation}
S_{\mathcal{P}}=M_{\mathcal{P}}\bar{V}_{\mathcal{P}},
\qquad
S_{\mathcal{D}}=M_{\mathcal{D}}\bar{V}_{\mathcal{D}}.
\end{equation}

Further define the within-subset CTVF-weighted mean errors by
\begin{equation}
\tilde{\xi}_{\mathcal{P}}
=
\frac{\sum_{i\in\mathcal{P}}V_i\xi_i}{\sum_{i\in\mathcal{P}}V_i},
\qquad
\tilde{\xi}_{\mathcal{D}}
=
\frac{\sum_{i\in\mathcal{D}}V_i\xi_i}{\sum_{i\in\mathcal{D}}V_i}.
\end{equation}
Then the CTVF weighting error can be written exactly as
\begin{equation}
\mathcal{E}(\Lambda^{\mathrm{CTVF}})
=
\frac{S_{\mathcal{P}}\tilde{\xi}_{\mathcal{P}}
+S_{\mathcal{D}}\tilde{\xi}_{\mathcal{D}}}
{S_{\mathcal{P}}+S_{\mathcal{D}}}.
\end{equation}

Define the total weight mass assigned to the proximal subset under the
two schemes as
\begin{equation}
\rho^{\ast}=\frac{S_{\mathcal{P}}}{S_{\mathcal{P}}+S_{\mathcal{D}}},
\qquad
\rho_0=\frac{M_{\mathcal{P}}}{M}.
\end{equation}
Accordingly,
\begin{equation}
\mathcal{E}(\Lambda^{\mathrm{uni}})
=
\rho_0\bar{\xi}_{\mathcal{P}}+(1-\rho_0)\bar{\xi}_{\mathcal{D}},
\end{equation}
and
\begin{equation}
\mathcal{E}(\Lambda^{\mathrm{CTVF}})
=
\rho^{\ast}\tilde{\xi}_{\mathcal{P}}+(1-\rho^{\ast})\tilde{\xi}_{\mathcal{D}}.
\end{equation}
Their difference admits the exact decomposition
\begin{equation}
\mathcal{E}(\Lambda^{\mathrm{uni}})-\mathcal{E}(\Lambda^{\mathrm{CTVF}})
=
\bigl[\rho_0(\bar{\xi}_{\mathcal{P}}-\tilde{\xi}_{\mathcal{P}})
+(1-\rho_0)(\bar{\xi}_{\mathcal{D}}-\tilde{\xi}_{\mathcal{D}})\bigr]
+(\rho^{\ast}-\rho_0)(\tilde{\xi}_{\mathcal{D}}-\tilde{\xi}_{\mathcal{P}}).
\end{equation}
This identity separates the error gap into two components. The first
term measures the within-subset reweighting effect, namely whether
CTVF concentrates more mass on lower-error evidence inside each
subset. The second term measures the between-subset reallocation
effect, namely whether CTVF assigns more total mass to the proximal
subset while that subset remains the lower-error one after
within-subset weighting.

Therefore, whenever the CTVF-induced reweighting relations satisfy
\begin{equation}
\tilde{\xi}_{\mathcal{P}}\le \bar{\xi}_{\mathcal{P}},
\qquad
\tilde{\xi}_{\mathcal{D}}\le \bar{\xi}_{\mathcal{D}},
\end{equation}
and
\begin{equation}
\rho^{\ast}>\rho_0,
\qquad
\tilde{\xi}_{\mathcal{P}}<\tilde{\xi}_{\mathcal{D}},
\end{equation}
the first bracket is non-negative and the second term is strictly
positive. Consequently,
\begin{equation}
\mathcal{E}(\Lambda^{\mathrm{CTVF}})<\mathcal{E}(\Lambda^{\mathrm{uni}}).
\end{equation}
Therefore, whenever the above CTVF-induced reweighting relations
hold, CTVF reduces weighting error along two complementary channels:
it concentrates weight on lower-error evidence inside each causal
subset through validity differences, and reallocates higher total
mass to the causally proximal subset across subsets. Under these
causal-proximity and error-ordering conditions, the normalized CTVF
weights consequently achieve a strictly lower weighting error than
uniform weights.
\end{proof}

\subsection{Proof of Proposition 2}
\label{app:proof2}

\noindent\textbf{Proposition 2.} \textit{Direction-aware deduplication
can eliminate the same-polarity Noisy-OR probability inflation induced
by overlap-identified redundant causal chains.}
\vspace{-3mm}
\begin{proof}
Fix a question $q$, let $\Pi$ denote either same-polarity set
$\Pi_q^{+}$ or $\Pi_q^{-}$, and let $\widehat\Pi \subseteq \Pi$ be
its deduplicated counterpart, for which any two retained chains have
Jaccard overlap at most $\theta$ and every discarded chain
$\pi \in \Pi \setminus \widehat\Pi$ admits some
$\pi' \in \widehat\Pi$ with $J(\pi, \pi') > \theta$. The same-polarity
Noisy-OR aggregation $p_{\mathrm{NoisyOR}}(\Pi) = 1 -
\prod_{\pi \in \Pi}(1 - \Phi(\pi, q))$ is the closed-form union
probability under the Noisy-OR independence assumption. Treating an
overlap-identified redundant chain as an additional independent factor
therefore weakly inflates the computed same-polarity aggregate.

Splitting the Noisy-OR product over $\Pi$ between $\widehat\Pi$ and
$\Pi \setminus \widehat\Pi$:
\begin{equation}
\prod_{\pi \in \Pi}\bigl(1 - \Phi(\pi, q)\bigr)
=
\prod_{\pi \in \widehat\Pi}\bigl(1 - \Phi(\pi, q)\bigr) \cdot
\prod_{\pi \in \Pi \setminus \widehat\Pi}\bigl(1 - \Phi(\pi, q)\bigr)
\leq
\prod_{\pi \in \widehat\Pi}\bigl(1 - \Phi(\pi, q)\bigr),
\end{equation}
since $\Phi(\pi, q) \in [0, 1]$ keeps the second factor in $[0, 1]$.
Hence
\begin{equation}
p_{\mathrm{NoisyOR}}(\Pi) \geq p_{\mathrm{NoisyOR}}(\widehat\Pi),
\end{equation}
and the inequality is strict whenever
$p_{\mathrm{NoisyOR}}(\widehat\Pi)<1$ and a removed chain has
$\Phi(\pi, q)>0$.

Define the polarity-wise redundancy factors
\begin{equation}
\eta^{+} = \prod_{\pi \in \Pi_q^{+} \setminus \widehat\Pi_q^{+}}
\bigl(1 - \Phi(\pi, q)\bigr) \in [0, 1],
\qquad
\eta^{-} = \prod_{\pi \in \Pi_q^{-} \setminus \widehat\Pi_q^{-}}
\bigl(1 - \Phi(\pi, q)\bigr) \in [0, 1],
\end{equation}
and the naive pre-deduplication probabilities
$p_{\mathrm{naive}}^{\pm} = p_{\mathrm{NoisyOR}}(\Pi_q^{\pm})$. The
splitting identity directly gives
\begin{equation}
1 - p_{\mathrm{naive}}^{+} = \eta^{+}\,\bigl(1 - p^{+}(q)\bigr),
\qquad
1 - p_{\mathrm{naive}}^{-} = \eta^{-}\,\bigl(1 - p^{-}(q)\bigr).
\end{equation}
Applying Boole's inequality $\mathbb{P}(\bigcup A_\pi) \leq \sum_\pi
\mathbb{P}(A_\pi)$ to the duplicated chains, equivalently the
Weierstrass product inequality
$\prod (1 - \phi_i) \geq 1 - \sum \phi_i$ for $\phi_i \in [0, 1]$,
yields
\begin{equation}
1 - \eta^{\pm} \;\leq\; B^{\pm},
\qquad
B^{\pm} \;:=\; \sum_{\pi \in \Pi_q^{\pm} \setminus \widehat\Pi_q^{\pm}}
\Phi(\pi, q),
\end{equation}
and combining with $1 - p_{\mathrm{naive}}^{\pm}
= \eta^{\pm}(1 - p^{\pm})$ delivers the polarity-wise inflation
upper bound
\begin{equation}
p_{\mathrm{naive}}^{\pm} - p^{\pm}(q)
= (1 - p^{\pm}(q)) \cdot (1 - \eta^{\pm})
\;\leq\; (1 - p^{\pm}(q)) \cdot B^{\pm}
\;\leq\; B^{\pm},
\end{equation}
which characterizes the maximum probability-level deviation that
omitting deduplication can incur and depends only on the confidences
of the removed redundant chains. Whenever $\eta^{\pm} < 1$,
$p_{\mathrm{naive}}^{\pm} \geq p^{\pm}(q)$; when $p^{\pm}(q) = 1$,
$p_{\mathrm{naive}}^{\pm} = 1$ and the inflation vanishes.

Greedy deduplication retains at least one chain from every non-empty
polarity. Define the neutral-anchored transformation
\begin{equation}
g_\varepsilon(p)=\mathrm{logit}\!\bigl(
\min(1-\varepsilon,\max(\varepsilon,(1+p)/2))\bigr).
\end{equation}
Applying the same unified combination before deduplication gives
\begin{equation}
P_{\mathrm{naive}}(q)=\sigma\!\bigl(
g_\varepsilon(p_{\mathrm{naive}}^{+})
-g_\varepsilon(p_{\mathrm{naive}}^{-})\bigr).
\end{equation}
Because $g_\varepsilon$ and $\sigma$ are non-decreasing, redundancy
confined to the positive polarity gives
$P_{\mathrm{naive}}(q)\geq P_{\mathrm{causal}}(q)$, whereas redundancy
confined to the negative polarity gives
$P_{\mathrm{naive}}(q)\leq P_{\mathrm{causal}}(q)$; each inequality is
strict whenever a removed positive-score chain changes the affected
clipped-logit value. When both polarities contain
redundancy, both polarity-wise Noisy-OR scores are inflated, while the
direction of the final change depends on their relative increments.
The same monotonicity argument applies without separate single-channel
branches, and two empty or equally strong channels return $\tfrac12$.

Therefore, direction-aware deduplication removes the overlap-identified
factors responsible for the bounded same-polarity Noisy-OR inflation,
while the neutral-anchored mapping preserves the direction of each
polarity-specific change in the final causal probability.
\end{proof}

\subsection{Proof of Proposition 3}
\label{app:proof3}

\noindent\textbf{Proposition 3.} \textit{Adaptive fusion using causal coverage and directional balance down-weights the causal estimate when evidence is unreliable, and can yield lower squared error than fixed weighting.}
\vspace{-3mm}
\begin{proof}
Let $q$ be drawn from a distribution on $\mathcal{Q}$, and let
$P_{\mathrm{causal}}(q), P_{\mathrm{llm}}(q) \in [0, 1]$ denote two
probability estimates of the true probability $p^{\star}(q) \in [0, 1]$.
Define the corresponding estimation errors by
\begin{equation}
\epsilon_C(q)=P_{\mathrm{causal}}(q)-p^{\star}(q),
\qquad
\epsilon_L(q)=P_{\mathrm{llm}}(q)-p^{\star}(q).
\end{equation}

For any weighting function $\alpha:\mathcal{Q}\to[0,\alpha_0]$, the
fused forecast is
\begin{equation}
\hat p_\alpha(q)
=
\alpha(q)P_{\mathrm{causal}}(q)
+\bigl(1-\alpha(q)\bigr)P_{\mathrm{llm}}(q),
\end{equation}
with fused error
\begin{equation}
\hat p_\alpha(q)-p^{\star}(q)
=
\alpha(q)\epsilon_C(q)+(1-\alpha(q))\epsilon_L(q).
\end{equation}
Define the corresponding squared-error risk by
\begin{equation}
R(\alpha)
=
\mathbb{E}_q\!\left[
\bigl(\alpha(q)\epsilon_C(q)+(1-\alpha(q))\epsilon_L(q)\bigr)^2
\right].
\end{equation}

For $\alpha_b>0$, $\zeta>0$, $0<\alpha_0\leq 1$, and
$\alpha_0<2\alpha_b$, define the saturation threshold
$\Omega_{\mathrm{sat}}:=\Omega_0+\zeta^{-1}
\operatorname{logit}(\alpha_0/(2\alpha_b))$ and the partition
\begin{equation}
\mathcal{G}_{\mathrm{hi}}=\{q:k(q)>0,\ \Omega(q)\ge \Omega_{\mathrm{sat}}\},
\qquad
\mathcal{G}_{\mathrm{lo}}=\mathcal{Q}\setminus\mathcal{G}_{\mathrm{hi}},
\end{equation}
and denote
\begin{equation}
\pi_g:=\mathbb{P}(\mathcal{G}_g),
\qquad
g\in\{\mathrm{hi},\mathrm{lo}\},
\qquad
\pi_{\mathrm{hi}}+\pi_{\mathrm{lo}}=1.
\end{equation}
Assume $0<\pi_{\mathrm{hi}},\pi_{\mathrm{lo}}<1$, so that both
conditional distributions are well defined.
Conditional expectations below are taken with respect to this
partition.

Suppose the causal estimate is less reliable on the low-reliability
stratum: there exists $\Delta>0$ such that
\begin{equation}
\mathbb{E}\!\left[\epsilon_C^2(q)\mid\mathcal{G}_{\mathrm{lo}}\right]
-
\mathbb{E}\!\left[\epsilon_C^2(q)\mid\mathcal{G}_{\mathrm{hi}}\right]
\ge \Delta.
\end{equation}
Suppose further that the LLM-side error structure is
stratum-invariant: there exist constants $B$ and $C$ with
\begin{equation}
\mathbb{E}\!\left[\epsilon_L^2(q)\mid\mathcal{G}_g\right]=B,
\qquad
\mathbb{E}\!\left[\epsilon_C(q)\epsilon_L(q)\mid\mathcal{G}_g\right]=C,
\qquad
g\in\{\mathrm{hi},\mathrm{lo}\}.
\end{equation}
Suppose finally that $B>C$, expressing non-degeneracy of the LLM
contribution.

By the tower property of conditional expectation,
\begin{equation}
R(\alpha)
=
\sum_{g\in\{\mathrm{hi},\mathrm{lo}\}}
\pi_g\,
\mathbb{E}\!\left[
\bigl(\alpha(q)\epsilon_C+(1-\alpha(q))\epsilon_L\bigr)^2
\,\Big|\,\mathcal{G}_g
\right].
\end{equation}

For any weighting that takes a constant value $\alpha$ on
$\mathcal{G}_g$, the conditional risk is
\begin{equation}
R_g^{\mathrm{cst}}(\alpha)
=
\alpha^2A_g+(1-\alpha)^2B+2\alpha(1-\alpha)C,
\qquad
A_g:=\mathbb{E}[\epsilon_C^2\mid\mathcal{G}_g].
\end{equation}
The quadratic coefficient admits the identity
\begin{equation}
A_g+B-2C
=
\mathbb{E}\!\left[(\epsilon_C-\epsilon_L)^2\mid\mathcal{G}_g\right]
\ge 0.
\end{equation}
Strict positivity follows from the Cauchy-Schwarz bound
$C^2 \leq A_g B$ together with $B > C$: when $C \geq 0$,
$A_g + B - 2C \geq (B - C)^2 / B > 0$; when $C < 0$,
$A_g + B - 2C > A_g + B > 0$. The coefficient is therefore strictly
positive in both regimes, so $R_g^{\mathrm{cst}}(\alpha)$ is a
strictly convex quadratic in $\alpha$ with unique unconstrained
minimizer
\begin{equation}
\alpha_g^\ast=\frac{B-C}{A_g+B-2C},
\end{equation}
and feasible minimizer
\begin{equation}
\bar\alpha_g^\ast=\mathrm{clip}_{[0,\alpha_0]}(\alpha_g^\ast).
\end{equation}

The two stratum-wise optima satisfy
\begin{equation}
\alpha_{\mathrm{hi}}^\ast-\alpha_{\mathrm{lo}}^\ast
=
\frac{(B-C)(A_{\mathrm{lo}}-A_{\mathrm{hi}})}
{(A_{\mathrm{hi}}+B-2C)(A_{\mathrm{lo}}+B-2C)}.
\end{equation}
Since $B-C>0$ and
\begin{equation}
A_{\mathrm{lo}}-A_{\mathrm{hi}}\ge \Delta>0,
\end{equation}
it follows that
\begin{equation}
\alpha_{\mathrm{hi}}^\ast>\alpha_{\mathrm{lo}}^\ast,
\qquad
\bar\alpha_{\mathrm{hi}}^\ast\ge \bar\alpha_{\mathrm{lo}}^\ast.
\end{equation}
Hence the constrained optimal fusion rule assigns no less weight to
the causal estimate in the high-reliability stratum than in the
low-reliability stratum.

For any fixed-weight scheme $\alpha_{\mathrm{fix}}\in[0,\alpha_0]$,
the same constant is assigned on both strata, so
\begin{equation}
R(\alpha_{\mathrm{fix}})
=
\pi_{\mathrm{hi}}R_{\mathrm{hi}}^{\mathrm{cst}}(\alpha_{\mathrm{fix}})
+
\pi_{\mathrm{lo}}R_{\mathrm{lo}}^{\mathrm{cst}}(\alpha_{\mathrm{fix}}).
\end{equation}
Strict convexity gives
\begin{equation}
R_g^{\mathrm{cst}}(\alpha_{\mathrm{fix}})
\ge
R_g^{\mathrm{cst}}(\bar\alpha_g^\ast),
\end{equation}
hence
\begin{equation}
R(\alpha_{\mathrm{fix}})
-
\Bigl[
\pi_{\mathrm{hi}}R_{\mathrm{hi}}^{\mathrm{cst}}(\bar\alpha_{\mathrm{hi}}^\ast)
+
\pi_{\mathrm{lo}}R_{\mathrm{lo}}^{\mathrm{cst}}(\bar\alpha_{\mathrm{lo}}^\ast)
\Bigr]
=
\sum_g\pi_g\Bigl[
R_g^{\mathrm{cst}}(\alpha_{\mathrm{fix}})
-
R_g^{\mathrm{cst}}(\bar\alpha_g^\ast)
\Bigr]
\ge 0.
\end{equation}
If $\bar\alpha_{\mathrm{hi}}^\ast\neq \bar\alpha_{\mathrm{lo}}^\ast$,
then $\alpha_{\mathrm{fix}}$ cannot simultaneously attain both
stratum-wise optima, so at least one stratum is strictly suboptimal,
and
\begin{equation}
R(\alpha_{\mathrm{fix}})
>
\pi_{\mathrm{hi}}R_{\mathrm{hi}}^{\mathrm{cst}}(\bar\alpha_{\mathrm{hi}}^\ast)
+
\pi_{\mathrm{lo}}R_{\mathrm{lo}}^{\mathrm{cst}}(\bar\alpha_{\mathrm{lo}}^\ast).
\end{equation}

Define the adaptive weighting rule by
\begin{equation}
\alpha(q):=\mathbbm{1}[k(q)>0]\,
\min\!\bigl(2\alpha_b\sigma(\zeta(\Omega(q)-\Omega_0)),\alpha_0\bigr).
\end{equation}
Since $\zeta>0$ and $\sigma$ is strictly increasing, $q\in\mathcal{G}_{\mathrm{hi}}$
gives $k(q)>0$ and $\Omega(q)\ge \Omega_{\mathrm{sat}}$, hence
\begin{equation}
2\alpha_b\sigma(\zeta(\Omega(q)-\Omega_0))\ge \alpha_0,
\end{equation}
and the truncation gives
\begin{equation}
\alpha(q)=\alpha_0.
\end{equation}
For $q\in\mathcal{G}_{\mathrm{lo}}$, either $k(q)=0$, which gives
$\alpha(q)=0$, or $\Omega(q)<\Omega_{\mathrm{sat}}$, which gives
\begin{equation}
\alpha(q)<\alpha_0.
\end{equation}

Denote $\bar\alpha_g:=\mathbb{E}[\alpha(q)\mid\mathcal{G}_g]$. Then
\begin{equation}
\bar\alpha_{\mathrm{hi}}=\alpha_0>\bar\alpha_{\mathrm{lo}},
\end{equation}
ordered in the same direction as
$\bar\alpha_{\mathrm{hi}}^\ast\ge \bar\alpha_{\mathrm{lo}}^\ast$.

Assume additionally that, within each stratum, conditioning on
$\alpha(q)$ leaves the three conditional moments $A_g$, $B$, and $C$
defined above unchanged. Since
$\alpha(q)$ is constant on $\mathcal{G}_{\mathrm{hi}}$ but may
vary on $\mathcal{G}_{\mathrm{lo}}$, the adaptive risk admits a
within-stratum variance term. As $R_g^{\mathrm{cst}}$ is quadratic,
its exact second-order Taylor expansion at $\bar\alpha_g$ gives
\begin{equation}
\mathbb{E}[R_g^{\mathrm{cst}}(\alpha(q))\mid\mathcal{G}_g]
=
R_g^{\mathrm{cst}}(\bar\alpha_g)
+
(A_g+B-2C)\,\mathrm{Var}(\alpha(q)\mid\mathcal{G}_g),
\end{equation}
hence
\begin{equation}
R(\alpha)
=
\sum_g \pi_g
\Bigl[
R_g^{\mathrm{cst}}(\bar\alpha_g)
+
(A_g+B-2C)\,\mathrm{Var}(\alpha(q)\mid\mathcal{G}_g)
\Bigr],
\end{equation}
with
\begin{equation}
\mathrm{Var}(\alpha(q)\mid\mathcal{G}_{\mathrm{hi}})=0,
\qquad
\mathrm{Var}(\alpha(q)\mid\mathcal{G}_{\mathrm{lo}})\ge 0.
\end{equation}

Therefore, for any fixed-weight scheme
$\alpha_{\mathrm{fix}}\in[0,\alpha_0]$, if
\begin{equation}
\sum_g\pi_g\!\left[
R_g^{\mathrm{cst}}(\alpha_{\mathrm{fix}})
-R_g^{\mathrm{cst}}(\bar\alpha_g)
\right]
>
\sum_g\pi_g(A_g+B-2C)
\,\mathrm{Var}(\alpha(q)\mid\mathcal{G}_g),
\end{equation}
it follows that
\begin{equation}
R(\alpha)<R(\alpha_{\mathrm{fix}})
\end{equation}
The directional-balance factor $\widetilde\beta(q)$ enters $\alpha(q)$
through the coverage score $\Omega(q)$: a smaller $\widetilde\beta(q)$
cannot increase $\Omega(q)$ and therefore cannot increase $\alpha(q)$.

Hence the adaptive fusion weight $\alpha(q)$ equals $\alpha_0$ on
$\mathcal{G}_{\mathrm{hi}}$ and is strictly smaller than $\alpha_0$
on $\mathcal{G}_{\mathrm{lo}}$. Whenever the weighted reduction in
stratum-wise mean risk exceeds the within-stratum variance penalty,
the strict inequality $R(\alpha)<R(\alpha_{\mathrm{fix}})$ shows that
adaptive fusion can attain lower squared-error risk than fixed
weighting.
\end{proof}

\section{Prompts Used in CHAIN}
\label{app:prompts}

\subsection{Hypergraph and Causal Extraction Prompt}
\label{app:prompt_a1}

\begin{figure}[t]
\centering
\includegraphics[width=1\linewidth]{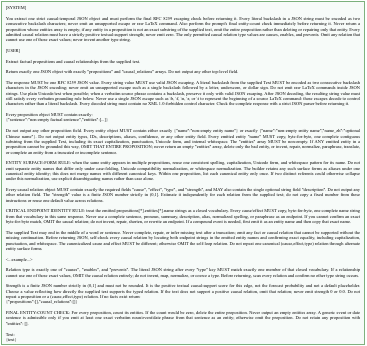}
\vspace{-6mm}
\caption{Hypergraph and Causal Extraction Prompt.}
\label{fig:prompt_a1}
\end{figure}

This prompt drives the construction of the causal-temporal
hypergraph $\mathcal{G} = (V, E_H, \mathcal{C})$ that underlies
\textbf{\textsc{Chain}}. Invoked once per factual chunk, it
implements the LLM extractor $\phi_{\mathrm{LLM}}$ that jointly
produces hyperedge-level factual propositions (forming $E_H$) and
typed directed causal edges with strength scores (forming
$\mathcal{C}$). The prompt is shown in Figure~\ref{fig:prompt_a1}.

\vspace{-8mm}
\begin{figure}[H]
\centering
\includegraphics[width=1\linewidth]{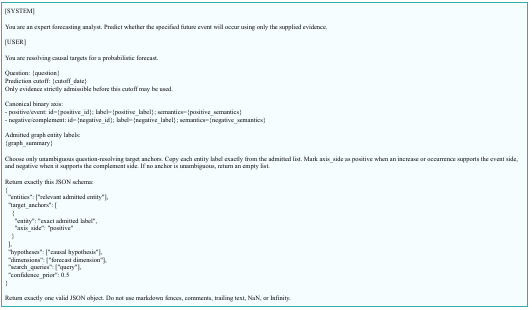}
\vspace{-6mm}
\caption{Reasoning Direction Generation Prompt.}
\label{fig:prompt_a2}
\end{figure}

\begin{figure}[H]
\centering
\includegraphics[width=1\linewidth]{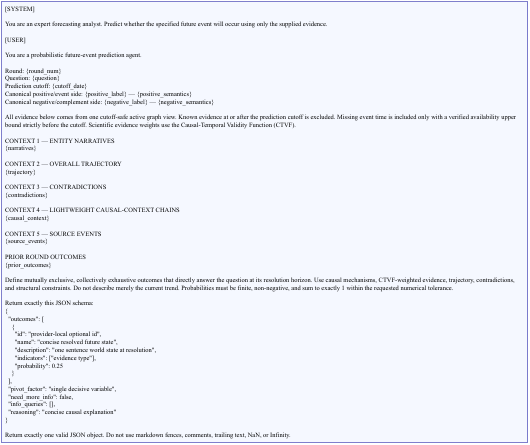}
\vspace{-6mm}
\caption{Multi-Outcome Probabilistic Reasoning Prompt.}
\label{fig:prompt_a3}
\end{figure}

\begin{figure}[H]
\centering
\includegraphics[width=1\linewidth]{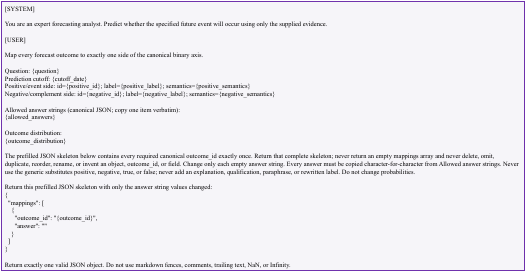}
\vspace{-6mm}
\caption{Outcome-to-Answer Mapping Prompt.}
\label{fig:prompt_a4}
\end{figure}

\subsection{Reasoning Direction Generation Prompt}
\label{app:prompt_a2}

This prompt is invoked once per query to record an LLM-side semantic
prior with question-conditioned candidate entities and hypotheses.
Candidate targets are selected from the admitted graph entities, and the
same call assigns each retained target to the positive or negative side of
the canonical binary axis; targets without an unambiguous axis assignment
are excluded. Causal distances are then computed independently by reverse BFS. The prompt is shown in
Figure~\ref{fig:prompt_a2}.

\subsection{Multi-Outcome Probabilistic Reasoning Prompt}
\label{app:prompt_a3}

This prompt implements the LLM-side estimator $f_{\mathrm{LLM}}$.
Each round consumes $q$ and graph context weighted by the prediction
cutoff (entity narratives,
contradictions, lightweight prompt-context causal chains, and
previous-round outcomes), and
emits a probability distribution over mutually exclusive
resolution outcomes; the distribution is subsequently projected
onto the binary probability $P_{\mathrm{llm}}$ by the
outcome-to-answer mapping step. These prompt-context chains precede
$P_{\mathrm{llm}}$ and are distinct from the subsequent CTVF-scored
pool $\Pi_q$ and its deduplicated subsets used to compute
$P_{\mathrm{causal}}(q)$. The prompt is shown in
Figure~\ref{fig:prompt_a3}.

\subsection{Outcome-to-Answer Mapping Prompt}
\label{app:prompt_a4}

This prompt finalizes the LLM-side estimate by aggregating the
probability mass consistent with the binary resolution of the question,
without modifying the underlying probabilities. The resulting scalar
$P_{\mathrm{llm}} \in [0, 1]$ is subsequently combined with
$P_{\mathrm{causal}}(q)$ via the adaptive fusion weight $\alpha(q)$.
The prompt is shown in Figure~\ref{fig:prompt_a4}.

\section{CHAIN Algorithm Details}
\label{app:algo}

Algorithm~\ref{alg:chain} presents the core calibration pipeline of \textbf{\textsc{Chain}}.
The three phases of the algorithm map one-to-one to the
three calibration bias sources targeted by our method, namely
evidence weighting, chain aggregation, and source combination.
\textbf{(1) Causal-Temporal Hypergraph Construction} builds the
corpus-level hypergraph $\mathcal{G} = (V, E_H, \mathcal{C})$ once.
For every factual chunk $x$, one extraction request to the extractor
$\phi_{\mathrm{LLM}}$ returns proposition/entity content and typed
directed causal edges (with retries only on failure); the graph builder attaches the chunk-derived
timestamp and optional stored similarity before
populating $\mathcal{G}$ that all queries subsequently share.
\textbf{(2) Per-Query Causal Estimation}
addresses evidence-weighting bias via the CTVF and chain-aggregation
bias via Direction-aware Noisy-OR. For each query, a cutoff-safe active
view $\mathcal{G}_q$ retains known pre-cutoff evidence and evidence with missing
or unparseable event time only when a verified availability upper bound precedes
$t_c$. Admitted occurrences
sharing the same cause, effect, and relation type are aggregated into a logical
causal edge whose strongest occurrence supplies its weight.
The reasoning-direction call receives the cutoff, canonical binary axis,
and a bounded set of question-relevant entity labels from $\mathcal{G}_q$;
keyword-filtered dense reranking then resolves $\mathcal{T}(q)$ with
positive/negative axis-side orientations $a_q$, providing the BFS roots
for the causal topological distance $d(\cdot, q)$. The distance drives the
proximity exponent $\rho$, the temporal recency $V_{\mathrm{rec}}$,
the recurrence-salience $V_{\mathrm{sal}}$, and the edge weight
$\omega(c)$. A lightweight prompt context $\mathcal{H}_{\mathrm{ctx}}(q)$,
comprising entity narratives, overall trajectory, contradictions, lightweight
causal-context chains, source events, and prior-round outcomes, is supplied to
$f_{\mathrm{LLM}}$ to obtain $P_{\mathrm{llm}}$.
These context chains are distinct from the formal CTVF chain pool $\Pi_q$
constructed below. Forward breadth-first search enumerates node-simple paths
from graph entities along outgoing causal edges; only paths that terminate at
an oriented target are retained, and target nodes are not expanded. Chains are
scored by $\Phi(\pi,q)$, partitioned by question-conditioned polarity $\delta_q(\pi)$,
deduplicated by Jaccard-$\theta$ within each polarity, and merged through the
neutral-anchored log-odds rule into $P_{\mathrm{causal}}(q)$.
\textbf{(3) Adaptive Source Fusion} addresses source-combination
bias. The unique-chain count $k$, the mean confidence
$\bar{\Phi}(q)$, the saturated reliability $\kappa(k)$, and the
floored balance $\widetilde{\beta}(q)$, computed on the
deduplicated chains, are passed through the coverage map
$\Omega(q)$ and a sigmoid gate to produce the question-specific
weight $\alpha(q)$, which combines $P_{\mathrm{causal}}(q)$ with
the LLM-side estimate
$P_{\mathrm{llm}} = f_{\mathrm{LLM}}(q,
\mathcal{H}_{\mathrm{ctx}}(q))$ into
the final forecast $\hat{p}$.

\begin{algorithm}[H]
\caption{\textbf{\textsc{Chain}}: Causal-Temporal Hypergraph Construction and
         Per-Query Calibration}
\label{alg:chain}
\small
\begin{algorithmic}[1]
\Require corpus of factual chunks $\{x\}$; query $q$ with prediction
         cutoff $t_c$ and canonical binary axis $\mathcal{A}_q$; extractor $\phi_{\mathrm{LLM}}$; estimator
         $f_{\mathrm{LLM}}$; encoder $\mathrm{Enc}(\cdot)$;
         hyperparameters $d_{\max}, B_\pi, F_{\max}, \tau_\Phi, \theta, \eta, k_{\mathrm{sat}},
         \beta_0, Z, \alpha_b, \alpha_0, \Omega_0, \zeta$; fixed $v_{\mathrm{fb}}$ and $\varepsilon=10^{-6}$
\Ensure  forecast $\hat{p} \in [0, 1]$
\Statex \textit{// 1: Causal-Temporal Hypergraph Construction}
\State Initialize $V \gets \emptyset$, $E_H \gets \emptyset$,
       $\mathcal{C} \gets \emptyset$
\For{each chunk $x$}
  \State $(\mathcal{P}_x, \mathcal{C}_x) \gets \phi_{\mathrm{LLM}}(x)$
  \State Attach to each $e\in\mathcal{P}_x$ the timestamp parsed from $x$
         and optional stored similarity $s_e$
  \State $E_H \gets E_H \cup \mathcal{P}_x$;\ \
         $\mathcal{C} \gets \mathcal{C} \cup \mathcal{C}_x$;\ \
         $V \gets V \cup \!\!\!\bigcup\limits_{(V_e, t, s) \in \mathcal{P}_x}\!\!\! V_e$
\EndFor
\State Encode $\mathrm{Enc}(v)$ for $v \in V$ and $\mathrm{Enc}(e)$
       for $e \in E_H$
\Statex \textit{// 2: Per-Query Causal Estimation}
\State $\mathcal G_q=(V_q,E_{H,q},\mathcal C_q) \gets
       \mathrm{CutoffSafeActiveView}(\mathcal G,q,t_c)$ using verified pre-cutoff
       admissibility and typed-edge aggregation
\State $\mathcal{L}_q \gets$ bounded question-relevant entity labels from $\mathcal G_q$
\State $\mathcal{D}(q) \gets \mathrm{ReasoningDirection}(q,t_c,\mathcal A_q,\mathcal L_q)$
\State $(\mathcal{T}(q),a_q) \gets \mathrm{ResolveAnchors}(\mathcal{D}(q),\mathcal G_q)$ by
       keyword-filtered dense reranking; discard weak or conflicting matches
\State $d(v, q) \gets$ exact reverse-BFS distance from $\mathcal{T}(q)$ on
       $\mathcal{C}_q$ within depth $d_{\max}$ for $v\in V_q$; set $d(v,q)=\infty$ otherwise
\State $d(e,q)\gets\min_{v\in V_e}d(v,q)$ for each $e=(V_e,t,s)\in E_{H,q}$
\State $\rho(n, q) \gets \min(d(n, q) / d_{\max},\, 1)$ for
       $n \in V_q \cup E_{H,q}$
\For{each $e = (V_e, t, s) \in E_{H,q}$}
  \State Set $N(e)$ to the number of distinct admitted records sharing its canonicalized proposition key
  \If{$t$ is parseable and $t<t_c$}
    \State $V_{\mathrm{rec}}(e, q) \gets
         \exp\!\bigl(-\tfrac{1}{2}\,\rho(e, q)\,
         \ln\max\{1,(t_c-t)_{\mathrm{days}}\}\bigr)$;\ \
         $V_{\mathrm{sal}}(e) \gets \mu(N(e))\,\sqrt{N(e)}\,M_r(e)$
    \State $V_{\mathrm{CTVF}}(e, q) \gets
           \min(V_{\mathrm{rec}}(e, q) + V_{\mathrm{sal}}(e),\, 1)$
  \Else
    \State $V_{\mathrm{CTVF}}(e,q)\gets v_{\mathrm{fb}}$ (only with a verified availability bound before $t_c$)
  \EndIf
\EndFor
\State $V_{\mathrm{CTVF}}(v, q) \gets$ mean of fixed incident-hyperedge scores
       $V_{\mathrm{CTVF}}(e, q)$ for $v \in V_q$ (else neutral $0.5$)
\State $\mathcal{H}_{\mathrm{ctx}}(q) \gets \mathrm{PromptContext}(\mathcal{G}_q, q, t_c)$;\quad
       $P_{\mathrm{llm}} \gets f_{\mathrm{LLM}}(q, \mathcal{H}_{\mathrm{ctx}}(q))$
\For{each $c = (u, v, \mathrm{type}, w) \in \mathcal{C}_q$}
  \State $\rho_c \gets \min\!\bigl((d(u, q) + d(v, q)) / (2\,d_{\max}),\;1\bigr)$;\ \
         $\omega(c) \gets
         \bigl(\tfrac{1}{2}(V_{\mathrm{CTVF}}(u, q)
         + V_{\mathrm{CTVF}}(v, q))\bigr)^{\rho_c}$
\EndFor
\State Forward breadth-first expand node-simple paths from each entity
       to depth $d_{\max}$ using at most the $F_{\max}$ strongest outgoing edges;
       set $\widetilde\Phi(\pi,q)\gets\prod_{\ell}w_\ell\omega(c_\ell)$
       and prune when $\widetilde\Phi(\pi,q)<\tau_\Phi$
\State Stop enumeration after $H_\pi\gets\max(5B_\pi,1000)$
       score-qualified target-terminating paths have been collected
\State Let $\Pi_q$ be the top $B_\pi$ target-terminating paths and split by
       $\delta_q(\pi)=a_q(v_L)(-1)^{P(\pi)}$ into
       $\Pi_q^{+},\, \Pi_q^{-}$
\State Greedy Jaccard-$\theta$ deduplication within each polarity
       $\to \widehat{\Pi}_q^{+},\, \widehat{\Pi}_q^{-}$
\State $p^{+}(q) \gets 1 - \prod_{\pi \in \widehat{\Pi}_q^{+}}
       (1 - \Phi(\pi, q))$;\ \
       $p^{-}(q) \gets 1 - \prod_{\pi \in \widehat{\Pi}_q^{-}}
       (1 - \Phi(\pi, q))$
\State $P_{\mathrm{causal}}(q) \gets \sigma\!\bigl(
       g_\varepsilon(p^{+}(q))-g_\varepsilon(p^{-}(q))\bigr)$, where
       $g_\varepsilon(p)=\mathrm{logit}(\min(1-\varepsilon,
       \max(\varepsilon,(1+p)/2)))$
\Statex \textit{// 3: Adaptive Source Fusion}
\State $k \gets |\widehat{\Pi}_q^{+}| + |\widehat{\Pi}_q^{-}|$;\ \
       $\bar{\Phi}(q) \gets \tfrac{1}{\max(k,1)}\!\!\!
       \sum\limits_{\pi \in \widehat{\Pi}_q^{+} \cup \widehat{\Pi}_q^{-}}\!\!\! \Phi(\pi, q)$
\State $\kappa(k) \gets \min(\ln(1+k) / \ln(1+k_{\mathrm{sat}}),\, 1)$
\State $\beta(q) \gets \min(|\widehat{\Pi}_q^{+}|, |\widehat{\Pi}_q^{-}|)
       / \max(|\widehat{\Pi}_q^{+}|, |\widehat{\Pi}_q^{-}|, 1)$;\ \
       $\widetilde{\beta}(q) \gets \beta_0 + (1 - \beta_0)\,\beta(q)$
\State $\Omega(q) \gets \min(k\,\bar{\Phi}(q)\,\kappa(k)\,
       \widetilde{\beta}(q) / Z,\, 1)$
\State $\alpha(q) \gets \mathbbm{1}[k > 0] \cdot
       \min(2\alpha_b\,\sigma(\zeta\,(\Omega(q) - \Omega_0)),\, \alpha_0)$
\State \Return $\hat{p} = \alpha(q)\,P_{\mathrm{causal}}(q)
       + (1 - \alpha(q))\,P_{\mathrm{llm}}$
\end{algorithmic}
\end{algorithm}

\textbf{Computational Complexity.} The computational cost of
\textbf{\textsc{Chain}} can be decomposed into three stages.
\textit{Hypergraph construction} issues one extraction request per factual
chunk, with retries only on failure, and scales linearly with
the corpus size; this cost is incurred only once per corpus and is
amortized over all downstream queries. Per-query construction of $\mathcal G_q$
adds one graph scan plus deterministic grouping of admitted occurrences.
\textit{Per-query causal estimation} performs a reverse breadth-first search over
$\mathcal{C}_q$ for each target in $\mathcal{T}(q)$, which runs in
$\mathcal{O}(|\mathcal{T}(q)|(|V_q| + |\mathcal{C}_q|))$ in the worst case and is bounded by the depth
$d_{\max}$, followed by breadth-first chain enumeration with uncapped
worst-case complexity $\mathcal{O}(|V_q|\,b^{d_{\max}})$ for average fan-out $b$.
The controls $F_{\max}$ and $\tau_\Phi$ restrict expansion,
$H_\pi=\max(5B_\pi,1000)$ caps score-qualified enumeration, $B_\pi$ bounds
$|\Pi_q|$, and Jaccard-$\theta$ deduplication determines $k$. Thus, the polarity-wise
Noisy-OR aggregation and the subsequent neutral-anchored combination admit
$\mathcal{O}(k)$ complexity. \textit{Adaptive fusion} runs in
$\mathcal{O}(1)$ given the chain statistics $k$, $\bar{\Phi}(q)$, and
the polarity counts. In practice, the per-query runtime is
dominated by the language-model invocations rather than by graph
traversal, encompassing the reasoning-direction generation, the
multi-outcome estimation performed by $f_{\mathrm{LLM}}$, and the
outcome-to-answer mapping.

\section{Dataset Details}
\label{app:datasets}

We conduct experiments on eight public forecasting datasets,
covering both in-distribution and out-of-distribution sources, and
spanning forecasting tasks across diverse domains.

\textbf{AI-Futures}~\citep{turtel2026future,aifutures2025}: A
forecasting dataset of AI-domain questions on AI labs, model
releases, agentic deployments, and AI-industry developments,
generated under the future-as-label paradigm so that ground-truth
labels arise from real-world events post-dating the model knowledge
cutoff.

\textbf{Metaculus}~\citep{chandak2026scalingopenendedreasoningpredict}:
A forecasting dataset of community-resolved questions extracted
from the Metaculus forecasting platform, with explicit closing and
resolution times and forecaster-count metadata.

\textbf{Polymarket}~\citep{polymarket}: A
market-driven forecasting dataset of prediction-market questions
sourced from Polymarket, where each question is backed by a
tradable contract whose outcome is settled by a real-world event.

\textbf{Future-as-Label}~\citep{turtel2026future}: A forecasting
dataset of mixed-domain questions whose ground-truth labels are
determined by future real-world outcomes occurring after the model
knowledge cutoff, ensuring that the task cannot be solved by
memorization.

\textbf{Clinical Trial Outcomes}~\citep{gao2024automatically}: A
large-scale benchmark of clinical trial outcomes for drug
development, providing trial-level success/failure labels and
introducing a biomedical out-of-distribution shift relative to the
in-distribution forecasting sources.

\textbf{FOReCAst}~\citep{yuan2025forecast}: A future-outcome
reasoning and confidence-assessment benchmark spanning Boolean,
timeframe, and quantity question types; we use its Boolean subset,
in which each question requires the model to predict an outcome
together with a confidence estimate.

\textbf{Golf-Forecasting}~\citep{turtel2026future,golfforecasting}:
A forecasting dataset of questions about professional golf
tournaments, generated under the future-as-label paradigm with news
articles as the underlying source, providing a fast-resolving
sports out-of-distribution setting.

\textbf{KalshiBench}~\citep{nel2025largelanguagemodelsknow}:
A forecasting benchmark sourced from the Kalshi regulated
event-contract exchange, covering sports, politics, macroeconomics,
and other event categories, designed to evaluate epistemic
calibration on prediction-market questions.

We designate the first four datasets as in-distribution sources and
the remaining four as held-out-source OOD datasets spanning diverse
forecasting domains, and choose three backbone LLMs whose knowledge
cutoffs predate the resolution dates
of nearly all sampled questions, so that calibration performance
reflects forward-only forecasting rather than memorization.
FOReCAst contains some potentially overlapping cases whose resolution
dates precede the model knowledge cutoffs, but this leakage risk applies
uniformly to all baselines and does not favor \textbf{\textsc{Chain}}.

To balance the cost of hypergraph construction against evaluation
coverage, we construct a separate causal-temporal hypergraph for
each of the four in-distribution datasets from 2{,}048 instances
sampled from that dataset. We also randomly sample 128 test instances
from each of the eight datasets; these test instances are disjoint from
the instances used for hypergraph construction. For the main OOD evaluation reported in
Table~2, the four out-of-distribution datasets do not use
dataset-specific retrieval corpora. Instead, we take the
union of the source documents of the four in-distribution
datasets as a single unified ID corpus, and each retrieval-based
method (NaiveRAG, GraphRAG, LightRAG, HippoRAG, HyperGraphRAG,
and \textbf{\textsc{Chain}}) builds its own retrieval structure
on top of this unified corpus using its own native construction
pipeline. This setup keeps the upstream data identical across
all methods while letting each method retain its own structural
representation, so observed OOD calibration differences reflect
algorithmic rather than corpus differences. Although the OOD datasets
are held out at the data-source level, their topical domains can overlap
with those of the in-distribution datasets. This setup therefore evaluates
cross-source generalisation under a unified ID retrieval corpus rather than
strictly domain-disjoint transfer; the degree of topical mismatch varies by
target dataset.

\section{Baseline Details}
\label{app:baselines}

We compare CHAIN with 13 baselines spanning direct probability elicitation,
step-by-step reasoning, retrieval-augmented generation, and post-hoc
probability calibration. The following descriptions summarize each method's
input protocol, structural representation, or calibration mapping.

\textbf{LLM Direct}~\citep{ouyang2022training} prompts the LLM with the raw question and asks it to output a probability and answer in a single shot.

\textbf{LLM CoT}~\citep{wei2022chain} prepends a chain-of-thought
prompt so that the LLM produces step-by-step reasoning before its
final probability and answer.

\textbf{AutoCast-style Prompting}~\citep{zou2022forecasting}
uses the forecasting setup of AutoCast, providing the LLM with the
question together with associated background news context as
supporting information.

\textbf{NaiveRAG}~\citep{lewis2020retrieval} follows the standard
retrieval-augmented generation paradigm, performing chunk-level
dense retrieval over the corpus and providing the top-$k$ chunks
to the LLM as supporting context.

\textbf{GraphRAG}~\citep{edge2024local} extracts an entity-level
knowledge graph from the corpus, applies hierarchical community
detection, and retrieves pre-generated community summaries to
support query-focused generation.

\textbf{LightRAG}~\citep{guo2024lightrag} builds a compact
entity-relation graph and performs dual-level retrieval that
combines a low-level entity-specific view and a high-level
theme-level view of the graph.

\textbf{HippoRAG}~\citep{gutierrez2024hipporag} adopts a
neurobiologically inspired memory index, applying personalized
PageRank over an OpenIE-style entity-and-triple graph to surface
multi-hop evidence.

\textbf{HyperGraphRAG}~\citep{luo2025hypergraphrag} represents
knowledge as $n$-ary relational hyperedges and performs
hypergraph-structured retrieval that fuses entity-level and
hyperedge-level evidence.

\textbf{Temperature Scaling}~\citep{guo2017calibration} learns
a single scalar temperature on a validation set to rescale logits
before the softmax.

\textbf{Platt Scaling}~\citep{platt1999probabilistic} fits a
sigmoid mapping from uncalibrated scores to class probabilities via
maximum likelihood.

\textbf{Isotonic Regression}~\citep{zadrozny2002transforming}
fits a non-parametric monotone mapping from predicted confidences to
empirical class probabilities.

\textbf{Histogram Binning}~\citep{zadrozny2001obtaining}
partitions the confidence axis into bins and replaces each bin's
prediction with the empirical class frequency observed in that bin
on the validation set.

\textbf{Conformal-Score Probability Adjustment}~\citep{angelopoulos2023conformal}
is a conformal-inspired scalar baseline. From the calibration predictions,
we compute nonconformity scores $r_i=|p_i-y_i|$ and their $0.9$ quantile $q$;
each evaluation probability is then mapped to
$\widetilde p=0.5+(1-q)(p-0.5)$. It outputs adjusted scalar probabilities
rather than prediction sets and makes no conformal coverage claim.

\section{Evaluation Metrics}
\label{app:metrics}

We evaluate calibration quality and predictive accuracy across
seven metrics spanning complementary dimensions of probabilistic
forecasting. ECE, ACE, MCE, and Reliability quantify calibration
error; NLL measures the logarithmic loss assigned to observed
outcomes; Brier score is a proper scoring rule capturing both
calibration and sharpness; and Accuracy measures the correctness
of binary predictions. Let $\{(p_i, y_i)\}_{i=1}^{N}$ denote $N$
predictions, where $p_i \in [0, 1]$ is the predicted probability
of the positive event and $y_i \in \{0, 1\}$ is the ground-truth
label. For each binning scheme, $B_m \subseteq \{1,\ldots,N\}$ denotes
the indices assigned to bin $m$, with $M=10$, mean predicted event
probability $\bar{p}_m=|B_m|^{-1}\sum_{i\in B_m}p_i$, and empirical
event frequency $\bar{y}_m=|B_m|^{-1}\sum_{i\in B_m}y_i$. Empty bins
are omitted from all bin-based sums and maxima.

\textbf{ECE (Expected Calibration Error)}~\citep{guo2017calibration}.
We compute the bin-weighted gap between the mean predicted event
probability and empirical event frequency under equal-width binning
of $[0, 1]$:
\begin{equation}
\mathrm{ECE}
=
\sum_{m=1}^{M} \frac{|B_m|}{N}
\bigl|\bar{p}_m - \bar{y}_m\bigr|.
\end{equation}
Lower values indicate better global calibration.

\textbf{ACE (Adaptive Calibration Error)}~\citep{nixon2019measuring}.
We use an adaptive-binning calibration error implemented with
approximately equal-mass bins:
\begin{equation}
\mathrm{ACE}
=
\sum_{m=1}^{M} \frac{|B_m|}{N}
\bigl|\bar{p}_m - \bar{y}_m\bigr|.
\end{equation}
Lower values indicate better calibration when predictions are
unevenly distributed across the predicted-probability axis.

\textbf{MCE (Maximum Calibration Error)}~\citep{guo2017calibration}.
Using the same equal-width bins as ECE, we take the worst bin-wise
calibration gap, restricting attention to bins containing at least
five samples for stability:
\begin{equation}
\mathrm{MCE}
=
\max_{m:\,|B_m| \ge 5}
\bigl|\bar{p}_m - \bar{y}_m\bigr|.
\end{equation}
Lower values indicate that no single predicted-probability region exhibits
severely miscalibrated predictions.

\textbf{Rel (Reliability)}~\citep{murphy1973new}. Using the same
equal-width bins as ECE, we report the binned estimate of the reliability
component of the Brier-score decomposition:
\begin{equation}
\mathrm{Rel}
=
\sum_{m=1}^{M} \frac{|B_m|}{N}
\bigl(\bar{p}_m - \bar{y}_m\bigr)^2.
\end{equation}
Lower values indicate that bin-wise mean event probabilities are closer to
the corresponding empirical frequencies.

\textbf{NLL (Negative Log-Likelihood)}. We compute the mean binary
logarithmic loss assigned to the observed outcomes:
\begin{equation}
\mathrm{NLL}
=
-\frac{1}{N}\sum_{i=1}^{N}
\left[
y_i\log \widetilde{p}_i
+(1-y_i)\log(1-\widetilde{p}_i)
\right],
\end{equation}
where $\widetilde{p}_i=\min(\max(p_i,\epsilon),1-\epsilon)$ and
$\epsilon=10^{-7}$ prevents numerical singularities. Lower values
indicate better probabilistic predictions.

\textbf{Brier (Brier Score)}~\citep{glenn1950verification}. We
compute the mean squared deviation between the predicted
probability and the binary outcome:
\begin{equation}
\mathrm{Brier}
=
\frac{1}{N} \sum_{i=1}^{N} (p_i - y_i)^{2}.
\end{equation}
Lower values indicate predictions that are jointly accurate and
well-calibrated.

\textbf{Acc (Accuracy)}. We report the fraction of correctly
predicted labels under a $0.5$ decision threshold:
\begin{equation}
\mathrm{Acc}
=
\frac{1}{N} \sum_{i=1}^{N}
\mathbbm{1}\!\left[\hat{y}_i = y_i\right],
\qquad
\hat{y}_i = \mathbbm{1}[p_i \ge 0.5].
\end{equation}
Higher values indicate more accurate binary predictions.

ECE, ACE, MCE, Rel, and Brier are reported in percent and are
lower-better; Accuracy is reported in percent and is higher-better.
NLL is reported on its dimensionless natural-log scale and is
lower-better.

\begin{table}[t]
\caption{Comparison of parameter settings across the three baseline categories and \textbf{\textsc{Chain}}.}
\label{tab:method_compare}
\centering
\fontsize{7.3pt}{8pt}\selectfont
\setlength{\tabcolsep}{2.56mm}
\begin{tabular*}{\textwidth}{@{}l@{\extracolsep{\fill}}ccccc@{}}
\toprule
\textbf{Group} & \textbf{Parameter}
& \textbf{Direct} & \textbf{RAG} & \textbf{Post-hoc} & \textbf{\textsc{Chain}} \\
\midrule
\multirow{4}{*}{Models}
& Backbone LLMs       & 3 LLMs                & 3 LLMs                  & 3 LLMs                & 3 LLMs \\
& Construction LLM    & --                    & GPT-4o-mini             & --                    & GPT-4o-mini \\
& Encoder             & --                    & text-embedding-3-small  & --                    & text-embedding-3-small \\
& Embedding dim       & --                    & 1536                    & --                    & 1536 \\
\midrule
\multirow{3}{*}{Inference}
& Temperature         & 0.3                   & 0.3                     & 0.3                   & 0.3 \\
& Max tokens          & 8192                  & 8192                    & 8192                  & 8192 \\
& Inference API       & official              & official                & official              & official \\
\midrule
\multirow{4}{*}{Retrieval}
& Chunk size          & --                    & 512 tokens              & --                    & 512 tokens \\
& Chunk overlap       & --                    & 64 tokens               & --                    & 64 tokens \\
& RAG top-$k$         & --                    & 10                      & --                    & -- \\
& Similarity          & --                    & cosine                  & --                    & cosine \\
\bottomrule
\end{tabular*}
\end{table}

\begin{table}[t]
\caption{Core \textbf{\textsc{Chain}}-specific hyperparameters and
default values used unless otherwise specified.}
\label{tab:chain_hparams}
\centering
\fontsize{7.3pt}{8pt}\selectfont
\setlength{\tabcolsep}{20pt}
\begin{tabular*}{\textwidth}{@{}l@{\extracolsep{\fill}}ccc@{}}
\toprule
\textbf{Symbol} & \textbf{Meaning} & \textbf{Value} & \textbf{Sensitivity} \\
\midrule
$d_{\max}$ & reachability/search depth cap                    & $4$    & swept $\{2,3,4,5\}$ \\
$\theta$   & Jaccard threshold for same-polarity dedup         & $0.5$  & swept $\{0.3,0.5,0.7\}$ \\
$\eta$     & recurrence saturation temperature in $V_{\mathrm{sal}}$ & $5.0$ & --- \\
$M_r(e)$   & default when stored similarity is unavailable            & $0.5$ & --- \\
$v_{\mathrm{fb}}$ & cutoff-safe missing-time CTVF fallback          & $0.25$ & --- \\
$k_{\mathrm{sat}}$ & reliability saturation anchor             & $10$   & --- \\
$\beta_0$  & balance floor                                     & $0.3$  & --- \\
$Z$        & coverage normalizer                               & $5.0$  & --- \\
$\alpha_b$ & mapping scale                                     & $0.5$  & --- \\
$\alpha_0$ & fusion upper bound $(1.2\alpha_b)$                & $0.6$  & --- \\
$\Omega_0$ & coverage activation threshold                     & $0.3$  & --- \\
$\zeta$    & sigmoid sharpness                                 & $3.0$  & --- \\
\bottomrule
\end{tabular*}
\end{table}

\begin{table}[t]
\caption{\textbf{\textsc{Chain}}'s gain over the strongest
baseline by backbone LLM. For each backbone and each metric, the
strongest baseline is identified by its mean over the four
in-distribution datasets, and the reported value is the gap
between \textbf{\textsc{Chain}}'s mean and this baseline's mean.
The hypergraph is extracted with GPT-4o-mini in all rows.}
\label{tab:extractor_robustness}
\centering
\fontsize{7.3pt}{8pt}\selectfont
\setlength{\tabcolsep}{26pt}
\begin{tabular*}{\textwidth}{@{}l@{\extracolsep{\fill}}ccc@{}}
\toprule
\textbf{Backbone} & $\boldsymbol{\Delta\text{ECE}\downarrow}$ & $\boldsymbol{\Delta\text{Brier}\downarrow}$ & $\boldsymbol{\Delta\text{Acc}\uparrow}$ \\
\midrule
\textbf{GPT-4o-mini}     & $\mathbf{-4.77}$ & $\mathbf{-5.57}$ & $\mathbf{+9.57}$ \\
Gemini-2.0-Flash         & $-4.15$          & $-4.34$          & $+8.20$ \\
DeepSeek-V3              & $-3.07$          & $-3.30$          & $+7.42$ \\
\bottomrule
\end{tabular*}
\end{table}

\section{Implementation Details}
\label{app:implementation}

We construct the causal-temporal hypergraph using
GPT-4o-mini~\citep{hurst2024gpt} as the extractor and
\texttt{text-embedding-3-small} as the shared encoder, and evaluate
\textbf{\textsc{Chain}} on three LLM backbones:
DeepSeek-V3~\citep{deepseekai2025deepseekv3technicalreport},
GPT-4o-mini~\citep{hurst2024gpt},
Gemini-2.0-Flash~\citep{kavukcuoglu2025gemini2}. All experiments
are conducted via the official APIs. For \textbf{\textsc{Chain}} and
all baselines, the reported comparison results are averaged over three
independent runs. To ensure a controlled comparison, every
method is run under the unified experimental setup summarized in
Table~\ref{tab:method_compare}, including matched decoding
parameters, the same evaluation set for each dataset, and the same
backbone LLMs. For all baselines, we adopt
the prompts from their official implementations. The
\textbf{\textsc{Chain}}-specific
hyperparameters that govern causal proximity, deduplication,
evidence aggregation, and source fusion are listed in
Table~\ref{tab:chain_hparams}. We use the listed defaults across
all eight datasets and three backbone LLMs unless
otherwise specified in the sensitivity analyses.

\noindent\textbf{Prompt Matching and Baseline Fairness.}
To ensure comparability across forecasting methods, Direct, CoT,
NaiveRAG, GraphRAG, HyperGraphRAG, and \textbf{\textsc{Chain}} follow
a unified input and evaluation protocol. All methods receive the same
question statement, prediction cutoff, and admissible corpus boundary.
For questions with multiple candidate outcomes, we apply the same
outcome-decomposition rule to map the task into a common binary-event
prediction space. We also use consistent option representations,
probability-output formats, and result-parsing rules across methods,
thereby controlling differences in prompt structure, output constraints,
and post-processing. For retrieval-augmented methods, the same cutoff
and corpus boundary apply, and all records directly associated with
test questions are excluded from the retrieval corpus. The resulting evaluation compares
the reasoning, retrieval structure, causal-temporal hypergraph
construction, and evidence aggregation used by each method under the
same protocol.

Although the hypergraph is extracted with GPT-4o-mini for all
datasets, \textbf{\textsc{Chain}}'s gain over the strongest
baseline transfers stably across backbone LLMs.
For each backbone and each evaluation metric, we identify
the strongest baseline as the one with the best mean across the
four in-distribution datasets, and report the gap between
\textbf{\textsc{Chain}}'s mean and this baseline's mean.
Across GPT-4o-mini, Gemini-2.0-Flash, and DeepSeek-V3, respectively,
\textbf{\textsc{Chain}} reduces ECE by $4.77$, $4.15$, and $3.07$
points, reduces Brier score by $5.57$, $4.34$, and $3.30$ points,
and improves Accuracy by $9.57$, $8.20$, and $7.42$ points;
per-backbone gains are reported in
Table~\ref{tab:extractor_robustness}. The evaluated search-control
configurations and the defaults adopted by \textbf{\textsc{Chain}}
are reported in Table~\ref{tab:search_configuration}.

\section{Results of DeepSeek-V3}
\label{app:deepseek_ood_results}

Table~\ref{tab:deepseek_ood_complete} reports the complete results of
DeepSeek-V3 on Clinical Trial Outcomes, FOReCAst, Golf-Forecasting, and
KalshiBench under the same unified in-distribution retrieval-corpus protocol
used for the main OOD evaluation. \textbf{\textsc{Chain}} achieves the lowest
ECE and Brier score and the highest Accuracy on each dataset and in the
four-dataset macro average. On the macro average, \textbf{\textsc{Chain}}
reduces ECE from $6.63$ to $5.75$ and Brier score from $22.82$ to $21.87$,
while increasing Accuracy from $62.70$ to $64.45$, relative to Temperature
Scaling, the strongest post-hoc baseline in this comparison. These results
show that the calibration and predictive-performance gains persist with
DeepSeek-V3 across all four OOD datasets.

\begin{table*}[t]
\caption{Performance comparison of \textbf{\textsc{Chain}} and the baselines
on 4 OOD datasets with DeepSeek-V3.}
\centering
\fontsize{7.3pt}{8pt}\selectfont
\setlength{\tabcolsep}{0.68mm}{
\begin{tabular}{lccc|ccc|ccc|ccc|ccc}
\toprule
\multirow{2.5}{*}{\textbf{Method}} & \multicolumn{3}{c}{\shortstack{\textbf{CTO}}} & \multicolumn{3}{c}{\textbf{FOReCAst}} & \multicolumn{3}{c}{\textbf{Golf-Forecasting}} & \multicolumn{3}{c}{\textbf{KalshiBench}} & \multicolumn{3}{c}{\textbf{Average}} \\
\cmidrule(lr){2-4} \cmidrule(lr){5-7} \cmidrule(lr){8-10} \cmidrule(lr){11-13} \cmidrule(lr){14-16}
& \textbf{ECE$\downarrow$} & \textbf{Brier$\downarrow$} & \textbf{Acc$\uparrow$} & \textbf{ECE$\downarrow$} & \textbf{Brier$\downarrow$} & \textbf{Acc$\uparrow$} & \textbf{ECE$\downarrow$} & \textbf{Brier$\downarrow$} & \textbf{Acc$\uparrow$} & \textbf{ECE$\downarrow$} & \textbf{Brier$\downarrow$} & \textbf{Acc$\uparrow$} & \textbf{ECE$\downarrow$} & \textbf{Brier$\downarrow$} & \textbf{Acc$\uparrow$} \\
\midrule
\rowcolor{grpDirect} LLM Direct & 15.47 & 25.80 & 60.16 & 9.72 & 20.03 & 72.66 & 14.77 & 26.38 & 57.03 & 16.09 & 25.30 & 60.94 & 14.01 & 24.38 & 62.70 \\
\rowcolor{grpDirect} LLM CoT & 16.52 & 25.01 & 57.81 & 10.80 & 20.28 & 71.09 & 17.76 & 29.48 & 54.69 & 20.16 & 25.54 & 58.59 & 16.31 & 25.08 & 60.55 \\
\rowcolor{grpDirect} AutoCast & 12.50 & 24.27 & 61.72 & 16.33 & 25.61 & 57.81 & 13.32 & 27.53 & 56.25 & 13.24 & 26.28 & 53.91 & 13.85 & 25.92 & 57.42 \\
\rowcolor{grpRetrieval} NaiveRAG & 19.84 & 25.79 & 60.16 & 19.56 & 27.31 & 60.16 & 11.13 & 26.63 & 49.22 & 15.59 & 28.68 & 46.88 & 16.53 & 27.10 & 54.10 \\
\rowcolor{grpRetrieval} GraphRAG & 13.28 & 25.29 & 59.38 & 8.55 & 20.77 & 71.09 & 14.22 & 26.61 & 53.91 & 18.74 & 26.46 & 53.12 & 13.70 & 24.78 & 59.38 \\
\rowcolor{grpRetrieval} LightRAG & 17.15 & 26.16 & 53.12 & 11.79 & 21.58 & 71.88 & 12.77 & 26.98 & 51.56 & 20.39 & 25.48 & 56.25 & 15.53 & 25.05 & 58.20 \\
\rowcolor{grpRetrieval} HippoRAG & 20.04 & 27.48 & 50.00 & 13.23 & 21.04 & 71.09 & 20.47 & 28.17 & 56.25 & 28.40 & 30.99 & 46.09 & 20.54 & 26.92 & 55.86 \\
\rowcolor{grpRetrieval} HyperGraphRAG & 21.48 & 26.54 & 50.00 & 13.24 & 21.04 & 71.09 & 13.63 & 28.56 & 48.44 & 23.95 & 26.26 & 53.12 & 18.08 & 25.60 & 55.66 \\
\rowcolor{grpPosthoc} TempScaling & 4.05 & 23.71 & 60.16 & 9.72 & 20.19 & 72.66 & 5.23 & 24.35 & 57.03 & 7.52 & 23.04 & 60.94 & 6.63 & 22.82 & 62.70 \\
\rowcolor{grpPosthoc} PlattScaling & 30.08 & 32.95 & 42.19 & 27.97 & 29.96 & 46.88 & 32.01 & 34.34 & 44.53 & 34.93 & 35.11 & 39.84 & 31.25 & 33.09 & 43.36 \\
\rowcolor{grpPosthoc} IsotonicReg & 28.57 & 31.87 & 48.44 & 25.02 & 26.47 & 67.19 & 29.91 & 33.01 & 50.78 & 30.69 & 32.26 & 57.03 & 28.55 & 30.90 & 55.86 \\
\rowcolor{grpPosthoc} HistogramBin & 36.45 & 44.86 & 44.53 & 27.30 & 32.08 & 62.50 & 31.03 & 35.59 & 57.81 & 33.19 & 38.78 & 53.91 & 31.99 & 37.83 & 54.69 \\
\rowcolor{grpPosthoc} ConformalAdj & 3.91 & 23.97 & 60.16 & 15.34 & 22.02 & 72.66 & 5.16 & 24.35 & 57.03 & 8.12 & 23.58 & 60.94 & 8.13 & 23.48 & 62.70 \\
\rowcolor{grpChain} \textbf{CHAIN (ours)} & \textbf{3.61} & \textbf{22.73} & \textbf{62.50} & \textbf{6.98} & \textbf{17.90} & \textbf{74.22} & \textbf{5.15} & \textbf{23.91} & \textbf{58.59} & \textbf{7.25} & \textbf{22.95} & \textbf{62.50} & \textbf{5.75} & \textbf{21.87} & \textbf{64.45} \\
\bottomrule
\end{tabular}}
\label{tab:deepseek_ood_complete}
\end{table*}

\begin{table}[H]
\centering
\fontsize{7.3pt}{8pt}\selectfont
\setlength{\tabcolsep}{0pt}
\caption{Evaluated search-budget, depth, fan-out, and pruning-threshold
settings, including the defaults used in the main experiments.}
\label{tab:search_configuration}
\begin{tabular*}{\textwidth}{@{}l@{\extracolsep{\fill}}cc@{}}
\toprule
\textbf{Parameter} & \textbf{Evaluated settings} & \textbf{Default} \\
\midrule
Chain budget $B_\pi$ & 50, 100, 200, 400 & 200 \\
Maximum search depth $d_{\max}$ & 2, 3, 4, 5 & 4 \\
Fan-out cap $F_{\max}$ & 2, 4, 8, 16, unlimited & unlimited \\
Pruning threshold $\tau_\Phi$ & 0, 0.001, 0.005, 0.02, 0.05 & 0.005 \\
\bottomrule
\end{tabular*}
\end{table}

\section{Search Configuration and Graph-Scale Sensitivity}
\label{app:search_graph_sensitivity}

\subsection{Search Configuration}
\label{app:search_configuration}

We examine the effects of the chain budget $B_\pi$, maximum search depth
$d_{\max}$, per-node outgoing-edge cap $F_{\max}$, and low-score prefix
pruning threshold $\tau_\Phi$. We use DeepSeek-V3 and report macro-averaged
results over AI-Futures, Future-as-Label, Clinical Trial Outcomes,
and Golf-Forecasting. All other settings remain fixed at their defaults.
Table~\ref{tab:search_configuration} summarizes the evaluated settings and
the defaults used in the main experiments. Table~\ref{tab:search_performance}
reports the corresponding predictive performance, where $\dagger$ denotes
the default adopted by \textbf{\textsc{Chain}}.

\begin{table}[t]
\centering
\fontsize{7.3pt}{8pt}\selectfont
\setlength{\tabcolsep}{3.2pt}
\caption{Predictive performance under different search configurations.}
\label{tab:search_performance}
\begin{tabular*}{\textwidth}{@{}l@{\extracolsep{\fill}}rccc@{}}
\toprule
\textbf{Parameter} & \textbf{Setting} & \textbf{ECE $\downarrow$} &
\textbf{Brier $\downarrow$} & \textbf{Accuracy $\uparrow$} \\
\midrule
Chain budget $B_\pi$ & 50 & 6.43 & 20.13 & 67.58 \\
 & 100 & 6.62 & 20.67 & 67.58 \\
 & $200^\dagger$ & 6.85 & 20.70 & 67.58 \\
 & 400 & 6.68 & 20.70 & 67.58 \\
\addlinespace[1pt]
Maximum search depth $d_{\max}$ & 2 & 6.53 & 20.14 & 67.97 \\
 & 3 & 5.90 & 20.47 & 67.58 \\
 & $4^\dagger$ & 6.85 & 20.70 & 67.58 \\
 & 5 & 6.60 & 20.81 & 67.38 \\
\addlinespace[1pt]
Outgoing-edge cap $F_{\max}$ & 2 & 7.04 & 20.37 & 68.16 \\
 & 4 & 6.32 & 20.23 & 68.36 \\
 & 8 & 6.01 & 20.40 & 67.77 \\
 & 16 & 6.49 & 20.21 & 67.97 \\
 & Unlimited$^\dagger$ & 6.85 & 20.70 & 67.58 \\
\addlinespace[1pt]
Pruning threshold $\tau_\Phi$ & 0 & 6.85 & 20.70 & 67.58 \\
 & 0.001 & 6.85 & 20.70 & 67.58 \\
 & $0.005^\dagger$ & 6.85 & 20.70 & 67.58 \\
 & 0.02 & 6.83 & 20.70 & 67.58 \\
 & 0.05 & 6.88 & 20.72 & 67.77 \\
\bottomrule
\end{tabular*}
\end{table}

Accuracy remains 67.58 across all chain budgets, while ECE and Brier
vary only slightly, showing stable predictions across changes in the
candidate-chain pool size. Performance also remains similar as the maximum
search depth and outgoing-edge cap change. Some smaller depth and edge-cap
settings improve all three metrics relative to their defaults, but the best
ECE, Brier, and Accuracy are attained by different configurations. Thus,
\textbf{\textsc{Chain}} maintains stable predictive quality across different
search spaces without depending on a single search configuration.

The pruning results show similar stability. Disabling pruning yields an ECE
of 6.85, a Brier score of 20.70, and an Accuracy of 67.58, identical to the
default threshold $\tau_\Phi=0.005$. Across the evaluated thresholds, ECE is
6.83--6.88, Brier is 20.70--20.72, and Accuracy is 67.58--67.77. Changing the
threshold therefore does not cause a material loss in predictive performance.

Table~\ref{tab:pruning_efficiency} further reports the numbers of candidate
chains before and after Jaccard deduplication, together with the mean runtime
of offline path search and prediction aggregation. Relative runtime is
normalized by the no-pruning setting $\tau_\Phi=0$. These measurements use
fixed graphs and fixed model outputs and exclude graph construction and
language-model calls.

\begin{table}[t]
\centering
\fontsize{7.3pt}{8pt}\selectfont
\setlength{\tabcolsep}{3.0pt}
\caption{Candidate-chain counts and computational cost under different
pruning thresholds. Relative runtime is normalized by $\tau_\Phi=0$.}
\label{tab:pruning_efficiency}
\begin{tabular*}{\textwidth}{@{}r@{\extracolsep{\fill}}rrrr@{}}
\toprule
$\boldsymbol{\tau_\Phi}$ & \textbf{Pre-dedup.} & \textbf{Post-dedup.} &
\textbf{Mean time (s)} & \textbf{Relative time $\downarrow$} \\
\midrule
0 & 30.76 & 13.31 & 10.48 & 1.00 \\
$0.005^\dagger$ & 30.76 & 13.31 & 2.47 & 0.24 \\
0.02 & 29.68 & 13.15 & 1.70 & 0.16 \\
0.05 & 24.16 & 12.32 & 1.25 & 0.12 \\
\bottomrule
\end{tabular*}
\end{table}

The default threshold $\tau_\Phi=0.005$ produces the same predictions as
disabling pruning while reducing the mean search-and-aggregation time from
10.48 to 2.47 seconds. Increasing the threshold to 0.02 and 0.05 further
reduces both candidate-chain counts and runtime while preserving predictive
quality. Low-score prefix pruning terminates subsequent expansion of weak
paths, reducing path-enumeration and aggregation costs while retaining the
principal causal-temporal evidence.

Overall, the predictive results remain stable as the chain budget, maximum
search depth, outgoing-edge cap, and pruning threshold change. Low-score
prefix pruning additionally reduces the cost of path search and prediction
aggregation.

\subsection{Graph Scale}
\label{app:graph_scale}

We next examine how the coverage of the causal-temporal graph affects
predictive performance. For each corpus source, we use a fixed random order
to construct nested 25\%, 50\%, 75\%, and 100\% source-record subsets. Each
larger subset contains the records retained by the smaller subsets, together
with all text chunks, nodes, and causal relations associated with the selected
records. Subset construction does not use prediction labels or model outputs.

\begin{table}[t]
\centering
\fontsize{7.3pt}{8pt}\selectfont
\setlength{\tabcolsep}{4.2pt}
\caption{Predictive performance under different causal-temporal graph scales.
Graph scale denotes the configured source-record sampling fraction; results
are macro-averaged over four datasets.}
\label{tab:graph_scale}
\begin{tabular*}{\textwidth}{@{}l@{\extracolsep{\fill}}ccc@{}}
\toprule
\textbf{Source-record fraction} & \textbf{ECE $\downarrow$} &
\textbf{Brier $\downarrow$} & \textbf{Accuracy $\uparrow$} \\
\midrule
25\%  & 8.62 & 22.24 & 63.87 \\
50\%  & 9.24 & 21.87 & 64.45 \\
75\%  & 8.86 & 21.58 & 65.23 \\
100\% & \textbf{6.85} & \textbf{20.70} & \textbf{67.58} \\
\bottomrule
\end{tabular*}
\end{table}

As the source-record fraction increases from 25\% to 100\%, the Brier score
decreases monotonically from 22.24 to 20.70, while Accuracy increases
monotonically from 63.87 to 67.58. ECE is not strictly monotonic at the
intermediate scales, but the complete graph still achieves the lowest ECE
and the best Brier score and Accuracy simultaneously.

The graph variants are nested, label-independent source-record subsets, with
the forecasting model, inference procedure, and evaluation protocol held
fixed. Expanding causal-temporal evidence coverage supplies
\textbf{\textsc{Chain}} with more complete events, entities, and causal
relations and improves probabilistic prediction quality. The complete graph
achieves the best result on all three metrics, demonstrating that
\textbf{\textsc{Chain}} can use the additional causal-temporal information
to produce more accurate and better-calibrated predictions.

Together, the search results show stable predictive quality under different
path-search configurations and lower search cost from low-score prefix
pruning, while the graph-scale results show improved performance with broader
causal-temporal evidence coverage. \textbf{\textsc{Chain}} therefore controls
path-search cost while effectively using a more complete evidence graph to
improve predictive quality.

\section{Robustness to Hypergraph Extractor Choice}
\label{app:extractor_robustness}

To assess the robustness of \textbf{\textsc{Chain}} to the choice of
hypergraph extractor, we separately construct the causal-temporal hypergraph
with GPT-4o-mini and GPT-4.1-nano while keeping the DeepSeek-V3 forecasting
backbone, input data, inference pipeline, and evaluation protocol fixed.
This evaluation covers all eight forecasting datasets: AI-Futures,
Metaculus, Polymarket, Future-as-Label, Clinical Trial Outcomes, FOReCAst,
Golf-Forecasting, and KalshiBench. We denote the graph-free and retrieval-free
forecasting control evaluated under the same protocol as Direct Multi-Outcome
forecasting (Direct-MO). For each question, it decomposes the resolution space into
mutually exclusive and collectively exhaustive outcomes, assigns a
probability to each outcome, projects the resulting distribution onto the
canonical binary axis, and sums the probability mass consistent with the
target event to obtain the final event probability.

\begin{table}[t]
\centering
\fontsize{7.3pt}{8pt}\selectfont
\setlength{\tabcolsep}{3.2pt}
\caption{Performance of \textbf{\textsc{Chain}} with different hypergraph
extractors across eight datasets.}
\label{tab:nano_extractor_datasets}
\begin{tabular*}{\textwidth}{@{}l@{\extracolsep{\fill}}lccc@{}}
\toprule
\textbf{Dataset} & \textbf{Hypergraph extractor} & \textbf{ECE$\downarrow$} &
\textbf{Brier$\downarrow$} & \textbf{Accuracy$\uparrow$} \\
\midrule
AI-Futures & GPT-4o-mini & 8.54 & 17.34 & 76.56 \\
AI-Futures & GPT-4.1-nano & 10.38 & 17.94 & 73.44 \\
Metaculus & GPT-4o-mini & 9.72 & 18.05 & 73.44 \\
Metaculus & GPT-4.1-nano & 12.26 & 19.40 & 74.22 \\
Polymarket & GPT-4o-mini & 13.74 & 17.41 & 79.69 \\
Polymarket & GPT-4.1-nano & 11.21 & 21.49 & 67.97 \\
Future-as-Label & GPT-4o-mini & 10.09 & 18.82 & 72.66 \\
Future-as-Label & GPT-4.1-nano & 6.11 & 20.38 & 68.75 \\
Clinical Trial Outcomes & GPT-4o-mini & 3.61 & 22.73 & 62.50 \\
Clinical Trial Outcomes & GPT-4.1-nano & 5.24 & 22.97 & 61.72 \\
FOReCAst & GPT-4o-mini & 6.98 & 17.90 & 74.22 \\
FOReCAst & GPT-4.1-nano & 11.62 & 18.96 & 71.88 \\
Golf-Forecasting & GPT-4o-mini & 5.15 & 23.91 & 58.59 \\
Golf-Forecasting & GPT-4.1-nano & 13.21 & 24.49 & 62.50 \\
KalshiBench & GPT-4o-mini & 7.25 & 22.95 & 62.50 \\
KalshiBench & GPT-4.1-nano & 9.07 & 23.14 & 59.38 \\
\bottomrule
\end{tabular*}
\end{table}

\begin{table}[t]
\centering
\fontsize{7.3pt}{8pt}\selectfont
\setlength{\tabcolsep}{4.2pt}
\caption{Macro-averaged performance across eight datasets under different
hypergraph-extractor settings.}
\label{tab:nano_extractor_macro}
\begin{tabularx}{\textwidth}{llCCC}
\toprule
\textbf{Method} & \textbf{Extractor} & \textbf{ECE$\downarrow$} &
\textbf{Brier$\downarrow$} & \textbf{Accuracy$\uparrow$} \\
\midrule
Direct-MO & None & 15.21 & 23.47 & 60.25 \\
\textbf{\textsc{Chain}} & GPT-4.1-nano & 9.89 & 21.10 & 67.48 \\
\rowcolor{grpChain} \textbf{\textsc{Chain}} & \textbf{GPT-4o-mini} & \textbf{8.14} &
\textbf{19.89} & \textbf{70.02} \\
\bottomrule
\end{tabularx}
\end{table}

Per-dataset results under the two extractors are reported in
Table~\ref{tab:nano_extractor_datasets}, and the eight-dataset macro-average
comparison is reported in Table~\ref{tab:nano_extractor_macro}. With
GPT-4.1-nano extraction, \textbf{\textsc{Chain}} reduces macro-average ECE
and Brier by 5.32 and 2.37 percentage points relative to Direct-MO and
improves Accuracy by 7.23 percentage points, showing that its calibration
and predictive advantages persist with a smaller hypergraph extractor.

With the default GPT-4o-mini extractor, macro-average ECE and Brier further
decrease to 8.14 and 19.89, while Accuracy increases to 70.02\%. Relative to
GPT-4.1-nano, GPT-4o-mini lowers ECE and Brier by 1.75 and 1.21 percentage
points and improves Accuracy by 2.54 percentage points.

\textbf{\textsc{Chain}} outperforms Direct-MO in macro-average ECE, Brier,
and Accuracy under both extractors, while the default GPT-4o-mini extractor
achieves the best result on all three metrics. Therefore,
\textbf{\textsc{Chain}} maintains its calibration and predictive advantages
across the evaluated hypergraph extractors and achieves higher overall
predictive quality with its default extractor.

\section{Comparison with Modern Post-Hoc Calibration Baselines}
\label{app:modern_calibration}

To compare \textbf{\textsc{Chain}} with modern post-hoc calibration
methods, we apply Temperature Scaling, Platt Scaling, Isotonic Regression,
Beta Calibration, Venn--Abers, and Binary Dirichlet calibration to the
corresponding LLM Direct predictions. All post-hoc methods use a
leakage-safe fitting protocol. For in-distribution data, we use five-fold
cross-fitting, so the calibrated prediction for each evaluated instance is
produced by a calibrator fitted without that instance's label. For
out-of-distribution data, each calibrator is fitted only on the pooled
in-distribution data for the corresponding backbone and then applied directly
to the out-of-distribution predictions. In contrast,
\textbf{\textsc{Chain}} performs calibration at inference time using
structured causal evidence. The table reports macro averages over the 24
dataset--backbone configurations formed by three backbones and eight datasets.

Because all outcomes in this evaluation are binary, our Binary Dirichlet
implementation uses the same unconstrained log-probability feature model as
Beta Calibration, with the same fitting and regularization protocol. The two
methods are therefore mathematically equivalent under this binary setting,
and their identical values are expected.

\begin{table}[t]
\centering
\fontsize{7.3pt}{8pt}\selectfont
\setlength{\tabcolsep}{4.2pt}
\caption{Comparison with modern post-hoc calibration methods under a
leakage-safe fitting protocol.}
\label{tab:modern_calibration}
\begin{tabularx}{\textwidth}{lCCCC}
\toprule
\textbf{Method} & \textbf{ECE$\downarrow$} & \textbf{Brier$\downarrow$} &
\textbf{NLL$\downarrow$} & \textbf{Acc$\uparrow$} \\
\midrule
Temperature Scaling & 12.81 & 23.72 & 0.6714 & 61.49 \\
Platt Scaling & 20.21 & 26.69 & 0.7405 & 54.43 \\
Isotonic Regression & 20.27 & 26.73 & 0.7860 & 52.90 \\
Beta Calibration & 20.61 & 26.64 & 0.7389 & 53.12 \\
Venn--Abers & 20.04 & 26.60 & 0.7384 & 52.86 \\
Binary Dirichlet & 20.61 & 26.64 & 0.7389 & 53.12 \\
\rowcolor{grpChain} \textbf{\textsc{Chain}} & \textbf{8.53} & \textbf{20.18} &
\textbf{0.5904} & \textbf{69.47} \\
\bottomrule
\end{tabularx}
\end{table}

The results in Table~\ref{tab:modern_calibration} show that
\textbf{\textsc{Chain}} achieves the lowest ECE, Brier score, and NLL
and the highest Accuracy relative to all listed post-hoc calibration baselines.
Temperature Scaling performs best among the listed post-hoc methods on all four
metrics. Relative to Temperature Scaling,
\textbf{\textsc{Chain}} lowers ECE and Brier score by 4.28 and 3.55
percentage points, respectively, lowers NLL by 0.0810, and improves Accuracy
by 7.98 percentage points. The experimental results show that inference-time
calibration based on structured causal evidence outperforms the evaluated
post-hoc calibration baselines in both probability quality and predictive
accuracy.

\section{Controlling for LLM Forecasting Calls with Multi-Sample Direct-MO}
\label{app:forecast_call_control}

To test whether increasing the number of LLM forecasts can attain the
performance gains of \textbf{\textsc{Chain}}, we compare a single
Direct-MO forecast, the probability average of five independent
Direct-MO forecasts, and the complete \textbf{\textsc{Chain}} method
on all eight datasets with DeepSeek-V3. Table~\ref{tab:forecast_call_control}
reports the average number of LLM calls and tokens per evaluation
instance, together with macro-averaged ECE, Brier score, NLL, and
Accuracy across the eight datasets. All methods use the same test set
for each dataset and the same metric definitions as in the main text;
LLM calls and token consumption are computed from the corresponding
inference runs.

\begin{table}[t]
\centering
\fontsize{7.3pt}{8pt}\selectfont
\setlength{\tabcolsep}{4.2pt}
\caption{LLM forecasting-call control and multi-sample Direct-MO
comparison with DeepSeek-V3 across eight datasets. ECE, Brier, and
Accuracy are in \%.}
\label{tab:forecast_call_control}
\begin{tabularx}{\textwidth}{lCCCCCC}
\toprule
\textbf{Method} & \textbf{LLM calls/case} & \textbf{Tokens/case} &
\textbf{ECE$\downarrow$} & \textbf{Brier$\downarrow$} &
\textbf{NLL$\downarrow$} & \textbf{Acc$\uparrow$} \\
\midrule
Direct-MO & 1.00 & 550 & 15.21 & 23.47 & 0.6627 & 60.25 \\
Five-sample Direct-MO & 5.00 & 2,751 & 14.91 & 23.07 & 0.6514 & 60.84 \\
\rowcolor{grpChain} \textbf{\textsc{Chain}} & 5.00 & 13,371 &
\textbf{8.14} & \textbf{19.89} & \textbf{0.5835} & \textbf{70.02} \\
\bottomrule
\end{tabularx}
\end{table}

Increasing Direct-MO from one to five independent forecasts reduces
ECE from 15.21 to 14.91, Brier score from 23.47 to 23.07, and NLL from
0.6627 to 0.6514, while increasing Accuracy from 60.25 to 60.84.
Independent repeated forecasting and probability averaging therefore
produce only small improvements across all four metrics, and their
overall performance remains below that of \textbf{\textsc{Chain}}.

Relative to five-sample Direct-MO, \textbf{\textsc{Chain}} reduces ECE
and Brier score by 6.77 and 3.18 percentage points, respectively,
reduces NLL by 0.0678, and improves Accuracy by 9.18 percentage points.
Both methods use approximately 5.00 LLM calls per evaluation instance,
while \textbf{\textsc{Chain}} performs better on all four predictive
metrics. These results show that increasing the number of independent
forecasts and averaging their probabilities does not attain the gains
of \textbf{\textsc{Chain}}; with essentially the same number of LLM
calls, structured causal-temporal inference more effectively improves
probability calibration and predictive accuracy.


\section{Robustness Across Calibration Error Estimators}
\label{app:calibration_estimator_robustness}

To test whether CHAIN's calibration advantage depends on a particular
error estimator, we evaluate the same predictive outputs with four
estimators that use different binning and finite-sample bias treatments:
Adaptive ECE-10, debiased RMSCE, Plugin RMSCE, and Jackknife
bias-corrected ECE. All metrics are computed from the same verified set
of test-set predictions. The evaluation comprises 24 evaluated
dataset--backbone combinations spanning 8 datasets and 3 backbone
language models, and Table~\ref{tab:calibration_estimator_robustness}
reports the macro-average over these combinations.

\begin{table}[t]
\centering
\fontsize{7.3pt}{8pt}\selectfont
\setlength{\tabcolsep}{4.2pt}
\caption{Macro-averaged calibration errors under four estimators across
24 evaluated dataset--backbone combinations spanning 8 datasets and 3
backbone language models.}
\label{tab:calibration_estimator_robustness}
\begin{tabularx}{\textwidth}{lCCCC}
\toprule
\textbf{Method} & \textbf{Adaptive ECE-10$\downarrow$} &
\textbf{Debiased RMSCE$\downarrow$} & \textbf{Plugin RMSCE$\downarrow$} &
\textbf{Jackknife ECE$\downarrow$} \\
\midrule
LLM Direct & 19.144 & 16.654 & 20.639 & 15.606 \\
\rowcolor{grpChain} \textbf{\textsc{Chain}} & \textbf{11.742} & \textbf{4.037} &
\textbf{11.474} & \textbf{4.312} \\
\bottomrule
\end{tabularx}
\end{table}

Across all four calibration error estimators, \textbf{\textsc{Chain}}
obtains lower error than LLM Direct. Specifically, its Adaptive ECE is
7.402 percentage points lower, its debiased RMSCE is 12.617 percentage
points lower, its Plugin RMSCE is 9.165 percentage points lower, and its
Jackknife bias-corrected ECE is 11.295 percentage points lower.

The four estimators characterize calibration from complementary
perspectives. Adaptive ECE-10 uses equal-frequency bins to reduce
sensitivity to uneven predicted-probability density; debiased RMSCE corrects the
finite-sample bias in squared calibration error; Plugin RMSCE directly
estimates root mean squared calibration error from empirical bin
statistics; and Jackknife bias-corrected ECE corrects finite-sample bias
through leave-one-out resampling. The consistently lower errors of
\textbf{\textsc{Chain}} show that its calibration improvement is not an
artifact of a particular binning scheme or bias-correction procedure.

\section{Extended Calibration Diagnostics and Probability Resolution}
\label{app:extended_calibration_diagnostics}

We further assess probabilistic prediction quality through complementary
measures of maximum bin-wise calibration error, reliability, probability
resolution, strictly proper scoring rules, and classification accuracy.
All metrics are computed from the same verified test-set predictions used
in the preceding section. The analysis covers 24 dataset--backbone
combinations formed by 3 backbone language models and 8 datasets, with
Table~\ref{tab:extended_calibration_diagnostics} reporting the macro-average
over these combinations. MCE-10 is the
largest absolute gap between mean predicted event probability and empirical
event frequency among ten probability bins. Reliability-10 measures the
weighted squared calibration gap across bins, whereas Resolution-10
measures between-bin variation in empirical outcome frequencies. We
also report Brier score, negative log-likelihood (NLL), and Accuracy to
jointly evaluate probabilistic predictions and classification decisions.

\begin{table}[t]
\centering
\fontsize{7.3pt}{8pt}\selectfont
\setlength{\tabcolsep}{4.2pt}
\caption{Extended calibration diagnostics and predictive performance,
macro-averaged over 24 dataset--backbone combinations. All metrics except
NLL are reported as their original values multiplied by 100.}
\label{tab:extended_calibration_diagnostics}
\begin{tabularx}{\textwidth}{lCCCCCC}
\toprule
\textbf{Method} & \textbf{MCE-10$\downarrow$} &
\textbf{Reliability-10$\downarrow$} & \textbf{Resolution-10$\uparrow$} &
\textbf{Brier$\downarrow$} & \textbf{NLL$\downarrow$} &
\textbf{Acc.$\uparrow$} \\
\midrule
LLM Direct & 39.121 & 4.530 & 2.805 & 25.314 & 0.721 & 61.491 \\
\rowcolor{grpChain} \textbf{\textsc{Chain}} & \textbf{22.701} & \textbf{1.428} &
\textbf{4.716} & \textbf{20.178} & \textbf{0.590} & \textbf{69.466} \\
\bottomrule
\end{tabularx}
\end{table}

As shown in Table~\ref{tab:extended_calibration_diagnostics},
\textbf{\textsc{Chain}} outperforms LLM Direct on all six complementary
metrics. It reduces MCE-10 from 39.121 to 22.701 and Reliability-10 from
4.530 to 1.428, corresponding to reductions of 16.420 and 3.102 on the
reported $\times 100$ scale, respectively. These results show that
\textbf{\textsc{Chain}} reduces both the maximum bin-wise calibration
error and the aggregate reliability error. Resolution-10
increases from 2.805 to 4.716, indicating that its predicted
probabilities more effectively distinguish samples with different
empirical outcome frequencies.

For overall probabilistic prediction quality, \textbf{\textsc{Chain}}
reduces Brier score from 25.314 to 20.178 and NLL from 0.721 to 0.590,
while increasing Accuracy from 61.491 to 69.466. The concurrent
improvements in calibration error, probability resolution, strictly proper
scoring rules, and classification accuracy demonstrate that
\textbf{\textsc{Chain}} produces probabilities that better align with
empirical outcomes while providing stronger discrimination and
predictive performance.

\section{Causal-Coverage Response of the Adaptive Fusion Weight}
\label{app:adaptive_fusion_coverage}

To characterize how \textbf{\textsc{Chain}} adjusts the contributions
of its prediction sources according to question-specific causal
evidence, we analyze the causal coverage $\Omega(q)$ and final fusion
weight $\alpha(q)$ for DeepSeek-V3 on the evaluation sets of all
eight datasets. Causal coverage summarizes the number of retained
causal chains, their mean confidence, evidence reliability, and
directional balance. The fusion weight $\alpha(q)$ determines the
contribution of the causal estimate $P_{\mathrm{causal}}(q)$ to the
final forecast. The recorded weights use mapping scale $\alpha_b=0.5$,
fusion upper bound $\alpha_0=0.6$, coverage activation threshold
$\Omega_0=0.3$, and sigmoid-transition sharpness $\zeta=3$; when the
retained-chain count is $k=0$, the zero-chain gate sets $\alpha(q)=0$.
Table~\ref{tab:adaptive_fusion_coverage} stratifies the verified prediction
records by causal coverage and reports the share of records and the mean and
median fusion weights within each stratum.

\begin{table}[t]
\centering
\fontsize{7.3pt}{8pt}\selectfont
\setlength{\tabcolsep}{4.2pt}
\caption{Adaptive fusion weights stratified by causal coverage on
DeepSeek-V3 across the evaluation sets of all eight datasets.}
\label{tab:adaptive_fusion_coverage}
\begin{tabular*}{\textwidth}{@{}l@{\extracolsep{\fill}}ccc@{}}
\toprule
\textbf{Causal coverage $\Omega(q)$} & \textbf{Record share (\%)} &
\textbf{Mean $\alpha(q)$} & \textbf{Median $\alpha(q)$} \\
\midrule
$\Omega(q)=0$                    & 49.22 & 0.000 & 0.000 \\
$0<\Omega(q)<0.10$               & 22.46 & 0.309 & 0.305 \\
$0.10\leq\Omega(q)<0.25$         &  7.42 & 0.399 & 0.404 \\
$0.25\leq\Omega(q)<0.50$         &  3.22 & 0.541 & 0.536 \\
$0.50\leq\Omega(q)\leq1.00$     & 17.68 & 0.600 & 0.600 \\
\bottomrule
\end{tabular*}
\end{table}

As shown in Table~\ref{tab:adaptive_fusion_coverage}, the fusion weight
increases with causal coverage. In the prediction records, the
$\Omega(q)=0$ stratum corresponds to $k=0$; the zero-chain gate therefore
sets the recorded fusion weight to zero, and the final forecast is given
by $P_{\mathrm{llm}}(q)$. For nonzero but low causal
coverage, the mean fusion weights are 0.309 and 0.399 in the first two
positive-coverage strata. The mean weight increases to 0.541 for
$0.25\leq\Omega(q)<0.50$. When causal coverage reaches 0.50 or above,
all corresponding prediction records attain the evaluation-time upper bound
of 0.600. Across all prediction records, the Spearman rank correlation
between $\Omega(q)$ and $\alpha(q)$ is 0.998, summarizing
their monotonic association in the recorded forecasts.

These results verify the expected response of $\alpha(q)$ to
question-specific causal coverage: the causal estimate does not
contribute to the final forecast when no usable causal evidence is
available, its contribution increases as evidence coverage grows, and
it reaches the prescribed upper bound in the high-coverage regime.
Thus, \textbf{\textsc{Chain}} does not apply a single fixed fusion
weight, but adjusts the contribution of the causal estimate to the
final probability within a bounded range according to question-level
causal coverage.

\section{Causal-Distance-Stratified Evidence-Node Audit Protocol}
\label{app:causal_distance_audit}

To characterize how \textbf{\textsc{Chain}} assigns validity across
causal distances, we define a node-level audit over the non-target
evidence nodes appearing in retained causal chains. Every retained chain
terminates at a target entity, so its terminal anchor has distance zero by
construction and is excluded from the audit. For each remaining node
$n\notin\mathcal{T}(q)$, we record its causal distance $d(n,q)$ and its
assigned validity $V_{\mathrm{CTVF}}(n,q)$, and stratify the observations
by $d(n,q)\in\{1,\ldots,d_{\max}\}$. Table~\ref{tab:causal_distance_audit}
formalizes the two complementary aggregation schemes.

\begin{table}[t]
\centering
\fontsize{7.3pt}{8pt}\selectfont
\setlength{\tabcolsep}{3.2pt}
\caption{Definition of the node-level causal-distance evidence audit.}
\label{tab:causal_distance_audit}
\begin{tabular*}{\textwidth}{@{}p{0.20\textwidth}@{\extracolsep{\fill}}p{0.35\textwidth}p{0.35\textwidth}@{}}
\toprule
\textbf{Component} & \textbf{Evidence-occurrence weighted} &
\textbf{Case-balanced} \\
\midrule
Analysis unit & Each occurrence of a non-target evidence node in a retained chain
& Each case--distance stratum \\
Distance strata & $d(n,q)\in\{1,\ldots,d_{\max}\}$
& $d(n,q)\in\{1,\ldots,d_{\max}\}$ \\
Aggregation & Mean $V_{\mathrm{CTVF}}(n,q)$ across node occurrences at each distance
& Mean within each case and distance, followed by an average across cases \\
Target handling & Exclude $n\in\mathcal{T}(q)$
& Exclude $n\in\mathcal{T}(q)$ \\
\bottomrule
\end{tabular*}
\end{table}

The evidence-occurrence-weighted estimate measures the validity assigned
to all recorded non-target node occurrences at a given distance. The
case-balanced estimate first averages within each case and distance
stratum and then averages across cases, preventing cases with more
retained chains from dominating the comparison. This formulation is
consistent with the target-terminated chain pool and isolates the
distance--validity relationship without conflating it with the
mechanically zero distance of the terminal target anchor.

\section{Controlled Perturbations of Causal-Edge Direction and Strength Ordering}
\label{app:causal_edge_perturbations}

To isolate the effects of causal-edge direction and relative strength
ordering, we conduct a controlled perturbation study with DeepSeek-V3
on all eight datasets. We hold fixed the forecasting questions,
target entities, node relevance scores, language-model probabilities,
adaptive fusion rule, and search settings, modifying only causal-edge
direction or the relative ordering of edge strengths. Causal distances,
retrieval chains, and final predictions are recomputed under every setting.
The three perturbations randomly permute strengths among causal edges,
apply the complementary transformation $w\leftarrow 1-w$, or reverse all
causal-edge directions.

Table~\ref{tab:causal_edge_perturbations} reports macro-average metric
changes relative to the unperturbed configuration, defined as
$\Delta m=m_{\mathrm{perturbed}}-m_{\mathrm{original}}$. Positive changes
in ECE, Brier score, and NLL indicate degradation, whereas a negative
change in Accuracy indicates degradation. Changes in ECE, Brier score,
and Accuracy are measured in percentage points.

\begin{table}[t]
\centering
\fontsize{7.3pt}{8pt}\selectfont
\setlength{\tabcolsep}{4.2pt}
\caption{Macro-average performance changes across eight datasets under
causal-edge direction and strength-order perturbations with DeepSeek-V3,
relative to the unperturbed configuration. ECE, Brier, and Accuracy
changes are in percentage points.}
\label{tab:causal_edge_perturbations}
\begin{tabularx}{\textwidth}{@{}>{\raggedright\arraybackslash}X
*{4}{>{\centering\arraybackslash}X}@{}}
\toprule
\textbf{Perturbation} & \textbf{$\Delta$ECE$\downarrow$} &
\textbf{$\Delta$Brier$\downarrow$} & \textbf{$\Delta$Accuracy$\uparrow$} &
\textbf{$\Delta$NLL$\downarrow$} \\
\midrule
Unperturbed \textsc{Chain} & 0.000 & 0.000 & 0.000 & 0.00000 \\
Shuffle edge strengths & $-0.034$ & $+0.024$ & $-0.098$ & $+0.00033$ \\
Complement strengths ($w\leftarrow1-w$) & $+0.829$ & $+0.015$ & $-0.195$ & $-0.00003$ \\
Reverse causal-edge directions & $+1.689$ & $+1.996$ & $-5.371$ & $+0.06869$ \\
\bottomrule
\end{tabularx}
\end{table}

As shown in Table~\ref{tab:causal_edge_perturbations}, reversing
causal-edge directions increases ECE and Brier score by 1.689 and 1.996
percentage points, decreases Accuracy by 5.371 percentage points, and
increases NLL by 0.06869. All four metrics change in the direction of
degradation, and their absolute changes are the largest among the three
perturbations. Direction reversal preserves the nodes, undirected
adjacency, and edge strengths while changing only edge orientation.
The concurrent deterioration in calibration error, probabilistic loss,
classification accuracy, and negative log likelihood therefore shows that the
evidence-propagation order and search reachability induced by causal-edge
direction provide important structural information for organizing
\textbf{\textsc{Chain}}'s inference paths.

Randomly shuffling edge strengths preserves the original multiset and
overall distribution of strengths but changes their association with
individual causal edges. This perturbation changes ECE by $-0.034$,
Brier score by $+0.024$, Accuracy by $-0.098$, and NLL by $+0.00033$.
The small magnitudes show that local reassignment of edge strengths does
not substantially alter overall predictive performance.

The complementary transformation further reverses the relative ordering
of the original strengths. It increases ECE by 0.829 percentage points
and decreases Accuracy by 0.195 percentage points, while changing Brier
score and NLL by only $+0.015$ and $-0.00003$, respectively. Compared with
random shuffling, this systematic reversal has a larger effect on
calibration and classification decisions, but every change remains
smaller than that induced by reversing causal-edge directions. Thus,
strength ordering modulates the relative contribution of inference paths,
while \textbf{\textsc{Chain}} remains comparatively stable across the two
strength-order perturbations.

Overall, direction and strength-order perturbations produce distinct
performance responses. Strength shuffling and complementation yield
limited changes with mixed signs, whereas direction reversal causes a
consistent and largest degradation across all four metrics. Under this
controlled evaluation, causal-edge direction has a more consistent effect
on predictions than the particular ordering of edge strengths, providing
empirical support for direction-sensitive causal retrieval while
demonstrating robustness to changes in strength ordering.

\section{Domain-Specific Retrieval-Corpus Control}
\label{app:ood_boundary_audit}

\subsection{Corpus Construction and Boundary Audit}

We conduct a domain-specific retrieval-corpus control to test whether
\textbf{\textsc{Chain}}'s relative performance depends on a mismatch
between the retrieval corpus and the target data source.

For Clinical Trial Outcomes, FOReCAst, Golf-Forecasting, and
KalshiBench, we construct target-domain-relevant corpora with
dataset-level temporal cutoffs, retaining only documents available on
or before the corresponding cutoff. Documents are collected from
prespecified domain categories rather than through question-directed
retrieval; evaluation-question records and fields containing answers,
labels, or resolution information are excluded. Within each dataset,
NaiveRAG, GraphRAG, HyperGraphRAG, and \textbf{\textsc{Chain}} share the
same audited source-document set, cutoff, and exclusion rule while
constructing their native retrieval structures. Table~\ref{tab:ood_corpus_audit}
summarizes the corpus boundaries.

\begin{table}[t]
\centering
\fontsize{7.3pt}{8pt}\selectfont
\setlength{\tabcolsep}{4.2pt}
\caption{Corpus-boundary audit for the domain-specific retrieval-corpus control.}
\label{tab:ood_corpus_audit}
\begin{tabular*}{\textwidth}{@{}l@{\extracolsep{\fill}}ccccc@{}}
\toprule
\textbf{Dataset} & \textbf{Common cutoff} & \textbf{Documents} &
\textbf{Chunks} & \textbf{Test-record exclusion} & \textbf{Shared boundary} \\
\midrule
Clinical Trial Outcomes & 2024-12-20 & 11 & 24 & Yes & Identical \\
FOReCAst & 2023-01-01 & 9 & 23 & Yes & Identical \\
Golf-Forecasting & 2025-10-12 & 9 & 27 & Yes & Identical \\
KalshiBench & 2025-08-19 & 12 & 39 & Yes & Identical \\
\bottomrule
\end{tabular*}
\end{table}

All four corpus audits return \texttt{status=passed} with empty error
lists. Retained documents have parseable dates, satisfy the corresponding
cutoff, and exclude evaluation-question records. Within each dataset,
all methods are bound to the same source-document set, holding corpus scope,
temporal boundaries, and exclusion conditions fixed across methods.

\subsection{Performance under Domain-Specific Retrieval Corpora}

Using DeepSeek-V3, we evaluate NaiveRAG, GraphRAG, HyperGraphRAG, and
\textbf{\textsc{Chain}} on the same test questions and metric definitions.
Table~\ref{tab:ood_domain_performance} reports equal-weight macro averages
over the four OOD datasets.

\begin{table}[t]
\centering
\fontsize{7.3pt}{8pt}\selectfont
\setlength{\tabcolsep}{4.2pt}
\caption{Macro-averaged performance under the domain-specific
retrieval-corpus control over four OOD datasets with DeepSeek-V3.}
\label{tab:ood_domain_performance}
\begin{tabularx}{\textwidth}{lCCCC}
\toprule
\textbf{Method} & \textbf{ECE$\downarrow$} & \textbf{Brier$\downarrow$} &
\textbf{NLL$\downarrow$} & \textbf{Acc$\uparrow$} \\
\midrule
NaiveRAG & 12.71 & 23.45 & 0.663 & 60.94 \\
GraphRAG & 17.57 & 24.15 & 0.705 & 59.57 \\
HyperGraphRAG & 12.19 & 22.95 & 0.650 & 62.70 \\
\rowcolor{grpChain} \textbf{\textsc{Chain}} & \textbf{8.22} &
\textbf{22.57} & \textbf{0.640} & \textbf{63.09} \\
\bottomrule
\end{tabularx}
\end{table}

Table~\ref{tab:ood_domain_performance} shows that
\textbf{\textsc{Chain}} achieves the best macro average on all four
metrics. Relative to the metric-specific strongest retrieval baseline,
it lowers ECE and Brier by 3.964 and 0.381 percentage points, respectively,
lowers NLL by 0.010, and increases Accuracy by 0.391 percentage points.
These macro-average results show that its overall calibration and predictive
advantages persist when all methods share the same corpus scope, temporal
boundaries, and exclusion conditions.

\begin{figure}[t]
\centering
\includegraphics[width=1\linewidth]{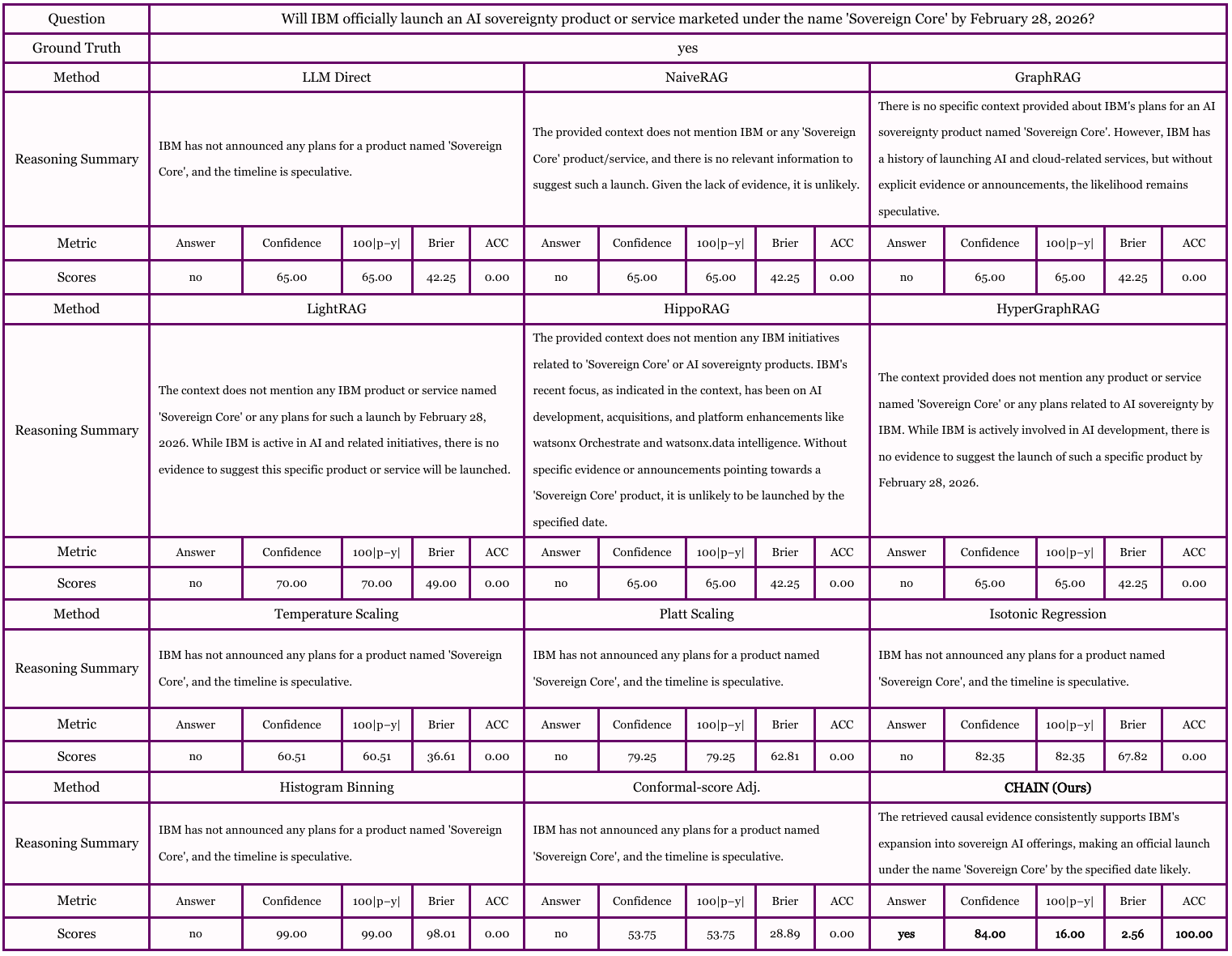}
\vspace{-3.6mm}
\caption{Single-question forecast comparison for a resolved forecasting
question, comparing LLM Direct, five RAG methods (NaiveRAG,
GraphRAG, LightRAG, HippoRAG, HyperGraphRAG), and five post-hoc calibrators (Temperature Scaling, Platt Scaling, Isotonic
Regression, Histogram Binning, Conformal-Score Probability Adjustment) based on
DeepSeek-V3, with \textbf{\textsc{Chain}} (Ours). The displayed $100|p-y|$
is the absolute probability error for this instance, not dataset-level ECE.}
\label{fig:case_study}
\end{figure}

\section{Case Study}
\label{app:case_study}

As shown in Figure~\ref{fig:case_study}, for the question of whether
IBM would officially launch an AI-sovereignty product or service
marketed under the name ``Sovereign Core'' by February 28, 2026,
all eleven baselines shown in the figure (LLM Direct, five RAG methods, and five post-hoc
calibrators) produce a \emph{no} prediction at confidences ranging
from $53.75\%$ to $99.00\%$, whereas the ground truth is \emph{yes}.
Among the methods shown in Figure~\ref{fig:case_study}, only
\textbf{\textsc{Chain}} produces the correct \emph{yes} prediction.
Its Brier score is $2.56\%$, compared with
$28.89\%$--$98.01\%$ for the displayed baselines, and is therefore approximately
$1/11.3$--$1/38.3$ of the corresponding baseline scores. Its absolute
probability error $100|p-y|$ is $16.00$, compared with
$53.75$--$99.00$ for the displayed baselines, or approximately
$1/3.4$--$1/6.2$ of their errors. Table~\ref{tab:calibration_trace}
further reports $P(\mathrm{yes})$ at each stage inside
\textbf{\textsc{Chain}}, showing how directionally consistent causal
evidence strengthens the probability support for an existing positive
prediction.

The displayed baselines exhibit a consistent prior-dominated failure. LLM Direct
rejects the event because there is no direct evidence that IBM had
announced a product named ``Sovereign Core.'' Although the retrieval
baselines recover information related to IBM's AI and cloud services,
the retrieved context does not explicitly mention ``Sovereign Core,''
and the absence of a direct reference is still interpreted as evidence
against the event. The five post-hoc calibrators transform the
probability produced by LLM Direct, but none corrects the prediction
direction in this case.

\begin{table}[t]
\centering
\caption{External reference prediction and internal probability
components of \textbf{\textsc{Chain}} in the case study (ground truth:
\emph{yes}, $P^{\star}=1$). The first row provides the standalone
LLM Direct reference; the remaining rows report \textsc{Chain}'s
LLM-side estimate, causal estimate, and final adaptive-fusion output.}
\label{tab:calibration_trace}
\fontsize{7.3pt}{8pt}\selectfont
\setlength{\tabcolsep}{32pt}
\begin{tabular*}{\textwidth}{@{}l@{\extracolsep{\fill}}cc@{}}
\toprule
\textbf{Prediction source} & $\boldsymbol{P(\mathrm{yes})}$ & $\boldsymbol{|P-P^{\star}|}$ \\
\midrule
LLM Direct (external reference)                         & $0.3500$          & $0.6500$          \\
$P_{\mathrm{llm}}$ (LLM inside \textsc{Chain})          & $0.6000$          & $0.4000$          \\
$P_{\mathrm{causal}}$ (Noisy-OR over 39 chains)         & $0.99999$         & $0.00001$         \\
$\hat p$ (Adaptive Fusion, $\alpha=0.60$)               & $\mathbf{0.8400}$ & $\mathbf{0.1600}$ \\
\bottomrule
\end{tabular*}
\end{table}

\textbf{\textsc{Chain}} obtains the correct, lower-error forecast
through two coupled structural stages. Anchored at the IBM entity in
the causal-temporal hypergraph, direction-aware evidence aggregation
retains 39 valid causal chains, of which 34 support and 5 oppose the
event, with causal-evidence coverage of $1.00$. Direction-aware
Noisy-OR aggregates these chains into
$P_{\mathrm{causal}}=0.99999$, providing stronger causal probability
support for the existing positive LLM-side forecast. Causal-Coverage-Balanced Adaptive
Fusion then combines $P_{\mathrm{llm}}=0.6000$ and
$P_{\mathrm{causal}}=0.99999$ with $\alpha=0.60$, producing
$\hat p=0.8400$, which is the $84.00\%$ confidence reported in
Figure~\ref{fig:case_study}.

As Table~\ref{tab:calibration_trace} shows, standalone LLM Direct
assigns only $0.3500$ probability to the true outcome, whereas
\textsc{Chain}'s internal LLM-side estimate is
$P_{\mathrm{llm}}=0.6000$. Within \textsc{Chain}, direction-aware
causal aggregation produces $P_{\mathrm{causal}}=0.99999$ from the
39 valid causal chains, and adaptive fusion combines it with
$P_{\mathrm{llm}}$ to yield $\hat p=0.8400$. The final probability
retains the uncertainty represented by the LLM-side estimate while
using directionally consistent causal evidence to increase confidence
in the positive prediction, resulting in lower Brier score and absolute
probability error than all eleven baselines shown in
Figure~\ref{fig:case_study}.

\section{Configuration-Level Comparative Analysis}
\label{app:significance}

To summarize \textbf{\textsc{Chain}}'s empirical improvements over the
metric-specific strongest baseline, we conduct a paired analysis on the
full matrix of $24$ (dataset, backbone) configurations
spanning three backbone large language models and eight forecasting
datasets, treating each configuration as one paired observation. For
each configuration and each evaluation metric, the strongest baseline
is identified separately by the best per-metric value among the thirteen
baselines under comparison: lowest ECE for ECE, lowest Brier score for
Brier, and highest Accuracy for Accuracy.

We report three complementary one-sided tests on the resulting paired
differences: the non-parametric Wilcoxon signed-rank test, the parametric
paired $t$-test, and the binomial test on the number of configurations in
which \textbf{\textsc{Chain}} outperforms the metric-specific strongest
baseline. The one-sided alternatives favor lower ECE and Brier score or
higher Accuracy for \textbf{\textsc{Chain}}.

\begin{table}[t]
\caption{Configuration-level paired comparisons of \textbf{\textsc{Chain}}
with the metric-specific strongest baseline across
$24$ (dataset, backbone) configurations spanning three backbone LLMs
and eight forecasting datasets. All tests are one-sided with the
alternative hypothesis ``\textbf{\textsc{Chain}} better''
($\Delta\text{ECE},\Delta\text{Brier}<0$;
$\Delta\text{Acc}>0$); $n=24$. Median differences are in percentage
points.}
\label{tab:sig_tests}
\centering
\fontsize{7.3pt}{8pt}\selectfont
\setlength{\tabcolsep}{2.86mm}{
\begin{tabular*}{\textwidth}{@{}l@{\extracolsep{\fill}}ccccc@{}}
\toprule
\textbf{Metric} & \textbf{Median $\Delta$}
& \textbf{Wilcoxon $p$} & \textbf{Paired-$t$ $p$}
& \textbf{Wins} & \textbf{Binomial $p$} \\
\midrule
$\Delta\text{ECE}\!\downarrow$   & $-0.581$ & $\mathbf{5.96\!\times\!10^{-8}}$ & $\mathbf{5.46\!\times\!10^{-4}}$ & $24/24$ & $\mathbf{5.96\!\times\!10^{-8}}$ \\
$\Delta\text{Brier}\!\downarrow$ & $-1.765$ & $\mathbf{5.96\!\times\!10^{-8}}$ & $\mathbf{7.37\!\times\!10^{-7}}$ & $24/24$ & $\mathbf{5.96\!\times\!10^{-8}}$ \\
$\Delta\text{Acc}\!\uparrow$     & $+2.734$ & $\mathbf{5.96\!\times\!10^{-8}}$ & $\mathbf{4.34\!\times\!10^{-7}}$ & $24/24$ & $\mathbf{5.96\!\times\!10^{-8}}$ \\
\bottomrule
\end{tabular*}}
\end{table}

As shown in Table~\ref{tab:sig_tests}, the selected metric-specific
strongest baseline is outperformed by \textbf{\textsc{Chain}} in all
$24$ configurations for each metric. For ECE, Brier score, and Accuracy,
\textbf{\textsc{Chain}} outperforms the metric-specific strongest
baseline in all $24$ configurations. The median paired differences
are $-0.581$ percentage points for ECE, $-1.765$ percentage points for
Brier score, and $+2.734$ percentage points for Accuracy. The agreement
among the Wilcoxon signed-rank test, paired $t$-test, and
configuration-level win-count test is consistent with these favorable
paired differences. Because the comparison baseline is selected from the
same evaluation results, the reported tests are descriptive rather than
confirmatory significance tests.

\section{Question-Level Paired Tests}
\label{app:question_level_statistics}

To assess differences at the question level, we pair
\textbf{\textsc{Chain}} with NaiveRAG, GraphRAG, LightRAG, HippoRAG,
and HyperGraphRAG on identical test questions and labels for
GPT-4o-mini, Gemini-2.0-Flash, and DeepSeek-V3 across all eight
datasets, yielding 120 paired comparisons. Each comparison corresponds
to one (backbone, dataset, baseline) combination.

For ECE, we use a paired prediction-swap randomization test with 10,000
randomizations; for Brier score, we use a paired sign-flip permutation
test with 10,000 randomizations. For Accuracy, we use the exact McNemar
test. We additionally construct 95\% paired bootstrap intervals for all
three metrics using 10,000 resamples. All tests use prespecified
one-sided alternatives that favor lower ECE and Brier score or higher
Accuracy for \textbf{\textsc{Chain}}. Holm correction is applied
separately to the 120 tests for each metric to control the family-wise
error rate. Table~\ref{tab:question_level_wins} reports directional
consistency, median effect size, bootstrap support, and significance
counts before and after multiple-comparison correction.

\begin{table}[t]
\centering
\fontsize{7.3pt}{8pt}\selectfont
\setlength{\tabcolsep}{2.4pt}
\caption{Question-level paired comparisons between
\textbf{\textsc{Chain}} and five retrieval-augmented baselines across
three backbone LLMs and eight datasets. $\Delta$ denotes
\textbf{\textsc{Chain}} minus the baseline in percentage points; the CI
column counts paired comparisons whose 95\% bootstrap interval lies
entirely in the favorable direction.}
\label{tab:question_level_wins}
\begin{tabular*}{\textwidth}{@{}l@{\extracolsep{\fill}}ccccc@{}}
\toprule
\textbf{Metric} & \shortstack{\textbf{Favorable}\\\textbf{comparisons}}
& \shortstack{\textbf{Median $\Delta$}\\\textbf{(pp)}}
& \shortstack{\textbf{Favorable}\\\textbf{95\% CI}}
& \shortstack{\textbf{Raw}\\$\boldsymbol{p<.05}$}
& \shortstack{\textbf{Holm}\\$\boldsymbol{p<.05}$} \\
\midrule
ECE $\downarrow$      & \textbf{120/120} & $-9.2591$  & 49/120 & 89/120 & 44/120 \\
Brier $\downarrow$    & \textbf{120/120} & $-5.7738$  & 75/120 & 84/120 & 27/120 \\
Accuracy $\uparrow$   & \textbf{120/120} & $+12.1094$ & 58/120 & 67/120 & 20/120 \\
\bottomrule
\end{tabular*}
\end{table}

As shown in Table~\ref{tab:question_level_wins},
\textbf{\textsc{Chain}} has the favorable direction in all 120 paired
comparisons for each metric: it yields lower ECE and Brier score and
higher Accuracy for every (backbone, dataset, baseline) combination.
This directional consistency spans all three backbone LLMs, all eight
datasets, and all five retrieval-augmented baselines. The median paired
differences also favor \textbf{\textsc{Chain}}: ECE and Brier score
decrease by 9.2591 and 5.7738 percentage points, respectively, while
Accuracy increases by 12.1094 percentage points.

Before multiple-comparison correction, 89 ECE, 84 Brier, and 67
Accuracy comparisons attain $p<.05$. After Holm correction, 44, 27,
and 20 comparisons, respectively, remain significant, with minimum
adjusted $p$-values of 0.0120, 0.0120, and
$4.08\!\times\!10^{-8}$. The 95\% bootstrap interval lies entirely in
the favorable direction in 49 ECE, 75 Brier, and 58 Accuracy
comparisons. The unanimous direction of the paired effects, their
median magnitudes, the bootstrap intervals, and the Holm-adjusted tests
provide mutually reinforcing evidence that \textbf{\textsc{Chain}}
consistently improves calibration, probabilistic prediction quality,
and classification accuracy across backbone models, datasets, and
retrieval-augmented baselines.

\section{Application Analysis}
\label{app:application_analysis}

Our evaluation covers financial events and event-contract markets
(Polymarket and KalshiBench), technology and industry (AI-Futures),
clinical research and development (Clinical Trial Outcomes), sports
forecasting (Golf-Forecasting), weather and climate events, and general
open-domain event forecasting (Metaculus, Future-as-Label, and
FOReCAst). Weather- and climate-related questions occur in Metaculus,
Future-as-Label, FOReCAst, and KalshiBench. In these domains, the
practical value of probabilistic forecasting depends not only on
directional accuracy, but also on whether predicted probabilities
retain a consistent statistical meaning across questions and
information environments. Systematic miscalibration causes risk
thresholds, event rankings, and resource-allocation decisions to be
based on biased confidence. \textbf{\textsc{Chain}} maps heterogeneous,
dynamic, and potentially dependent evidence into calibrated
probabilities through causal-temporal evidence weighting,
direction-aware aggregation, and causal-coverage-adaptive fusion. Its
consistent advantages across domains and under distribution shift
provide methodological and empirical support for the applications
discussed below.

\noindent\textbf{Financial Events and Event-Contract Markets.}
Event-contract and macro-risk analysis require probabilities that are
directly comparable across events and can support exposure ranking,
threshold management, and scenario monitoring. These information
environments often contain reports that are source-correlated or
semantically duplicated; treating them as independent evidence can
systematically inflate confidence. \textbf{\textsc{Chain}}'s
direction-aware deduplication and evidence aggregation control repeated
accumulation while retaining the structure of evidence both for and
against an event. Causal-coverage-adaptive fusion further regulates the
contribution of the structured estimate according to effective causal
coverage. The resulting probabilities can support event-level risk
assessment, contract screening, and macro-event monitoring.

\noindent\textbf{Technology Development and Public-Event Analysis.}
Technology milestones, industrial evolution, and public events are
commonly driven by multiple interacting factors whose evidence differs
in temporal validity and causal proximity. \textbf{\textsc{Chain}}
assigns evidence weights according to event-time differences and causal
distance, increasing the contribution of recent evidence that is more
directly connected to the forecast target while reducing the influence
of stale background information and distant associations. This
structure can support technology-roadmap monitoring, industrial
scenario comparison, critical-event alerts, and probability updates at
decision points.

\noindent\textbf{Clinical Trials and Drug Development.}
Clinical-development decisions require the integration of trial design,
drug mechanisms, disease context, interim outcomes, and external
research evidence to continuously assess project-level success
probabilities. Such evidence often exhibits explicit causal
dependencies and temporal ordering that are not preserved by simply
concatenating retrieved passages. \textbf{\textsc{Chain}} organizes
relevant evidence into traceable causal chains and derives the final
probability from chain reliability, directional balance, and causal
coverage. These event-level probabilities can support trial-risk
assessment, development-pipeline screening, candidate prioritization,
and resource allocation on a common probabilistic scale.

\noindent\textbf{Weather and Climate Event Forecasting.}
Weather and climate event forecasting requires the integration of
historical climate conditions, recent observations, seasonal variation,
and continuously updated event information into probabilities with a
clear statistical meaning. Such evidence is strongly time-dependent and
may be repeated across multiple sources. \textbf{\textsc{Chain}}'s
temporal-validity weighting reduces the influence of stale information,
direction-aware deduplication limits repeated accumulation of duplicated
reports, and causal-distance weighting together with
causal-coverage-adaptive fusion regulates the final probability according
to the relevance of the evidence to the forecast target. These mechanisms
can support probabilistic forecasts, risk monitoring, and warning
decisions for weather and climate events such as rainfall thresholds,
storm and hurricane occurrence, extreme temperatures, snowfall, and
drought.

\noindent\textbf{Sports Forecasting.}
Sports events have short resolution horizons, frequent information
updates, and a strong dependence on recent conditions. A forecasting
system must incorporate new evidence promptly while preventing
historical information or repeated reporting from exerting
disproportionate influence. The temporal decay in
\textbf{\textsc{Chain}} reduces the weight of less timely evidence,
causal-distance weighting prioritizes information more closely
connected to the target outcome, and direction-aware deduplication
limits confidence shifts caused by repeated descriptions of the same
event. The resulting estimates can support event-state monitoring,
outcome-probability updating, and critical-event alerts.

\noindent\textbf{Open-Domain Event Forecasting.}
Open-domain forecasting spans heterogeneous topics, evidence sources,
and resolution horizons, requiring probability quality to remain stable
as the domain and information distribution change.
\textbf{\textsc{Chain}} does not depend on a fixed set of domain-specific
features; instead, it constructs each forecast from question-conditioned
causal structure, temporal validity, and evidence coverage. Its
cross-domain and distribution-shift results support its use as a general
probabilistic inference layer for open-domain forecasting platforms,
risk-monitoring systems, and analyst workflows, providing outputs with
a consistent statistical meaning for question ranking, scenario
tracking, threshold-based decisions, and human review.

\section{Limitations}
\label{app:limitations}

While \textbf{\textsc{Chain}} shows consistent calibration advantages in our experiments, five limitations remain. First, the causal-temporal hypergraph is constructed by an LLM extractor, so errors in entities, relation directions, or strengths can propagate to downstream aggregation and fusion; deduplication and directional-balance correction mitigate but do not eliminate this uncertainty. Second, bounded reachability and finite chain-enumeration budgets trade runtime efficiency against long-range coverage and may omit relevant evidence. Third, the current output is an endpoint-event probability rather than an explicit model of intermediate states, transition probabilities, event timing, or alternative trajectories. Fourth, per-query inference requires multiple LLM calls and a non-trivial prompt budget, which may constrain large-scale or low-latency deployment. Fifth, temporal admissibility depends on corpus boundaries and metadata, and empirical validation remains concentrated on English event-forecasting benchmarks and selected knowledge-intensive domains. These limitations concern robustness, computational scaling, task scope, and external validity rather than the core calibration mechanism under the evaluated forecasting protocol.

\section{Future Work}
\label{app:future_work}

Future work will address these limitations in five directions. First, multi-sample extraction, cross-chunk consistency constraints, and outcome-supervised relation estimation can improve the hypergraph and propagate extraction uncertainty into aggregation and fusion. Second, reachability and chain-enumeration budgets can become query-adaptive, allocating deeper search when evidence is distant and stopping when sufficient directional evidence has accumulated. Third, \textbf{\textsc{Chain}} can extend endpoint prediction to process-level event trajectories covering intermediate states, transitions, timing, and alternative paths. Fourth, prompt compression, auxiliary-model distillation, subgraph caching, and reuse of stable relation judgments can reduce inference cost. Fifth, joint temporal-boundary validation and evaluation on multilingual, scientific, legal, and low-resource forecasting can broaden evidence for calibration, temporal validity, and interpretable decision support. These extensions will test whether the same mechanisms remain reliable under broader evidence, languages, temporal boundaries, and deployment constraints and broader operating conditions.

\end{document}